%% file: main.tex
\documentclass[11pt,letterpaper]{article}
\usepackage[T1]{fontenc}
\usepackage{lmodern}
\usepackage[margin=1in]{geometry}
\usepackage{amsmath,amssymb,amsthm,mathtools}
\usepackage{graphicx,booktabs,microtype,xcolor}
\usepackage{algorithm,algpseudocode}
\usepackage[authoryear,round]{natbib}
\usepackage[font=small,labelfont=bf]{caption}
\usepackage{url}
\usepackage[bookmarksnumbered]{hyperref}
\hypersetup{
  hidelinks,
  pdftitle={Information Limits of Replay-Free Recurrent State Migration},
  pdfauthor={Khizar Qureshi},
  pdfsubject={Recurrent state migration and causal memory},
  pdfkeywords={recurrent models, state migration, causal memory, state-space models}
}
\setcitestyle{citesep={;},aysep={,},yysep={;}}

\newtheorem{theorem}{Theorem}
\newtheorem{proposition}[theorem]{Proposition}
\newtheorem{corollary}[theorem]{Corollary}
\newtheorem{lemma}[theorem]{Lemma}
\theoremstyle{definition}

\newcommand{\R}{\mathbb R}
\newcommand{\B}{\mathbb B}
\newcommand{\norm}[1]{\left\lVert#1\right\rVert}
\newcommand{\rank}{\operatorname{rank}}
\newcommand{\im}{\operatorname{im}}
\newcommand{\spn}{\operatorname{span}}
\newcommand{\KL}{\operatorname{KL}}
\newcommand{\diag}{\operatorname{diag}}

\newcommand{\calI}{\mathcal I}
\newcommand{\calV}{\mathcal V}

\input{generated/results}
\input{generated/discussion_results}
\title{\LARGE\bfseries Information Limits of Replay-Free\\Recurrent State Migration}
\author{Khizar Qureshi\\[0.3em]MIT\\[0.2em]\href{mailto:kqureshi@mit.edu}{\texttt{kqureshi@mit.edu}}}
\date{}
\begin{document}
\maketitle
\begin{abstract}
\input{sections/abstract}
\end{abstract}
\input{sections/main}
\clearpage
\input{sections/statements}
\bibliographystyle{plainnat}
\bibliography{references}
\clearpage
\appendix
\input{sections/proofs}
\input{sections/new_proofs}
\input{sections/implementation}
\input{sections/experiments}
\input{sections/revision_experiments}

\input{sections/round2_experiments}

\end{document}

%% file: generated/results.tex
\newcommand{\SpectralDimensions}{0, 1, 1, 3, 5}
\newcommand{\SpectralBits}{0.00000,\allowbreak 1.90465,\allowbreak 5.22657,\allowbreak 20.78339,\allowbreak 49.74601}

\newcommand{\LanguageCubicMs}{1.163}
\newcommand{\LanguageCubicIQR}{[1.152, 1.201]}
\newcommand{\LanguageDegreeThreeMs}{1.123}
\newcommand{\LanguageDegreeThreeIQR}{[1.098, 1.209]}
\newcommand{\CubicWinningSeeds}{3}
\newcommand{\LanguageCubicError}{5.30\times10^{-4}}
\newcommand{\LanguageDegreeThreeError}{7.49\times10^{-4}}
\newcommand{\DominantSeedShare}{95.7}
\newcommand{\TaskDegreeThreeNmse}{2.15\times10^{-5}}
\newcommand{\TaskStaleNmse}{0.375}
\newcommand{\TaskAnchorNmse}{0.134}
\newcommand{\TaskRecoveredBenefitMean}{99.99045}
\newcommand{\TaskRecoveredBenefitMinimum}{99.98134}
\newcommand{\HistoryEightStatePercent}{10.7}
\newcommand{\HistoryEightOutputMaximum}{3.07\times10^{-7}}
\newcommand{\SplitEightOutputMaximum}{6.56\times10^{-2}}
\newcommand{\SplitSixtyFourOutputMaximum}{1.86\times10^{-7}}
\newcommand{\RepeatedEightTaskNmse}{2.50\times10^{-9}}
\newcommand{\SplitAnchorNmse}{1.40\times10^{-2}}
\newcommand{\SplitStaleNmse}{2.90\times10^{-2}}
\newcommand{\SplitReplayNmse}{3.08\times10^{-10}}
\newcommand{\SplitUnrestrictedNmse}{2.66\times10^{-10}}
\newcommand{\SplitDegreeThreeNmse}{3.54\times10^{-9}}
\newcommand{\ReachableFloatError}{4.79\times10^{-8}}
\newcommand{\ReachableDoubleError}{2.49\times10^{-16}}

%% file: generated/discussion_results.tex
\newcommand{\PairedLanguageP}{0.475}
\newcommand{\PairedLanguageFlipP}{0.75}
\newcommand{\PairedLanguageDifference}{-2.19\times10^{-4}}
\newcommand{\PairedLanguageSd}{6.23\times10^{-4}}
\newcommand{\PairedLanguageCI}{[-9.92\times10^{-4},5.54\times10^{-4}]}
\newcommand{\SplitStitchNmse}{1.59\times10^{-6}}
\newcommand{\SplitFirstNmse}{7.22\times10^{-5}}
\newcommand{\SplitStitchOutputMax}{2.04\times10^{-3}}

%% file: sections/abstract.tex
When a recurrent model is replaced, its old state may not determine what the new model would compute from the same history. We characterize the information that must be retained before the replacement is known. For a time-invariant affine linear family, the minimum continuous causal memory for order-$k$ local migration equals the reachable dimension of a lifted coefficient system minus the information in the anchor state. A controllable $d$-state model with scalar input needs at most $kd$ additional real coordinates, even when arbitrary matrix replacement directions are revealed only after streaming. A single rank-one direction can already require all $kd$. Conditional singular values and auxiliary-bit bounds distinguish this local information law from recovery at a fixed tolerance. We also give invariant-subspace and future-output constructions. Experiments compare coefficient, interpolation, and history-basis representations at matched memory budgets and prescribed output tolerances. In a coupled identification task with repeated slow poles, eight history coordinates meet normalized output error $10^{-4}$ across five fitted seeds. Splitting the poles raises the smallest tested successful width to 64. These results separate necessary historical information from the cost and accuracy of a chosen representation.

%% file: sections/main.tex
\section{Introduction}
A recurrent model compresses a history under its current parameters. Replacing those parameters can change which historical distinctions matter. An expressive migration map cannot recover distinctions that the old state has discarded. We ask what information a processor must retain \emph{before the replacement is known} to reconstruct the state that the new model would compute from the same inputs.

The target is retrospective. A query at time $t$ requests the state obtained by applying one replacement model to the complete prefix $x_{1:t}$, not the trajectory generated by historical weight changes. This question arises when a running filter is recalibrated or an existing recurrent context is reinterpreted. We study linear recurrent models with fixed input injections. Arbitrary changes to a multilayer network are outside this setting.

One relevant case is transition-only adaptation of a linear recurrent block, such as an LRU, with its input projection and upstream features frozen \citep{orvieto2023}. The general causal characterization permits vector inputs. The sharper $kd$ bound additionally requires a controllable scalar-input anchor.

Our main result identifies the minimum continuous causal memory for local approximation of order $k$. It equals the reachable dimension of a lifted coefficient system minus the information already present in the anchor state. For a controllable $d$-state model with scalar input, the sharp worst-case bound is $kd$ additional real coordinates. The encoder can be fixed before \emph{any matrix replacement direction} is specified, while one rank-one direction can already require all $kd$ coordinates. Unknown arbitrary directions and one suitably chosen direction therefore have the same worst-case local information cost. The result counts independent historical information, which can be smaller than the coefficient array that represents it.

Exact local order, finite-tolerance accuracy, and practical storage are distinct. We give conditional singular-value and auxiliary-bit bounds for a fixed finite query set, then construct causal coefficient memories under an invariant-subspace condition. An output quotient separates preservation of future predictions from preservation of every state coordinate. The experiments compare coefficient, interpolation, and history-basis representations at matched memory budgets and prescribed prediction tolerances. The memory-width study relates the required storage to the requested output and the anchor's pole structure, with shared setup and decoder costs accounted for separately.

\paragraph{Relation to prior work.}
Recurrent-state staleness is studied by R2D2 \citep{kapturowski2019}. RTRL propagates exact first derivatives \citep{williams1989}, UORO approximates them online \citep{tallec2018}, and structured recurrent learning reduces their maintenance cost \citep{farr2026}. Expansion, invariant projection, rational realization, and residual estimation are established tools of parametric model reduction \citep{benner2015,haasdonk2011}. Continuous nonlinear widths supply the fixed-horizon language \citep{devore1989,siegel2024}. Laguerre models and HiPPO retain historical signals in prescribed bases \citep{wahlberg1991,gu2020}, whereas stitching transforms a representation using model-pair calibration \citep{bansal2021,mai2026}. Context composition within an SSM \citep{liu2025} and state compression during training \citep{chahine2026} concern different targets. The contribution is a causal information characterization for a replacement revealed after streaming. The differentiation rules, orthogonal filter families, and generic residual bounds are established ingredients. Appendix~\ref{app:positioning} gives result-level comparisons.

\section{Causal information before a replacement}\label{sec:lower}
Let $h_t(a)$ be the state produced from zero by processing $x_{1:t}$ with the same coordinate $a$. The anchor is $h_t(0)$. An auxiliary summary updates from its previous value, the anchor, and the current input without knowing future queries. A query returns an estimate of $h_t(a)$ but does not change the underlying summary. Queries at $a$ and then $a'$ each concern the full prefix. They do not describe a trajectory that first used $a$ and then $a'$. After a deployment change, maintaining the anchor adds storage and computation to those of the active model.

Two histories already show why auxiliary information is necessary. The scalar old state is $h_2(0)=\lambda x_1+x_2$. Histories $(x_1,x_2)$ and $(x_1+v,x_2-\lambda v)$ collide in that state, but replacing $\lambda$ by $\lambda+a$ separates their targets by $av$. Retaining $x_1$ before the query resolves the ambiguity. Whether this distinction matters depends on the tolerance and readout.

\begin{table}[t]
\caption{Four different memory questions. Local results count continuous real coordinates, not bits. All targets reprocess the same historical inputs under one replacement model.}
\label{tab:scope}
\centering\small
\begin{tabular}{p{.22\linewidth}p{.36\linewidth}p{.32\linewidth}}
\toprule
Question & Result & Essential scope \\
\midrule
Causal local order $k$ & $\dim\mathcal K_k-\dim\pi_0\mathcal K_k$ & Fixed time-invariant affine family, real inputs \\
Directions revealed later & Sharp worst-case $kd$ extra coordinates & Controllable scalar input, exact arithmetic \\
Fixed-horizon tolerance & Conditional singular values and a bit lower bound & Unit-energy histories, finite queries, exact old state \\
Implementation cost & Coefficients, interpolation, or history banks & Maintenance, conditioning, and predictive tolerance measured separately \\
\bottomrule
\end{tabular}
\end{table}

\subsection{The reachable coefficient space}
Consider the time-invariant family
\begin{equation}
h_t(a)=\left(A_0+\sum_{i=1}^p a_iA_i\right)h_{t-1}(a)+Bx_t,
\qquad h_0(a)=0.
\label{eq:local-family-main}
\end{equation}
An auxiliary summary supports order-$k$ local migration when, for every finite history,
\begin{equation}
\norm{h_t(a)-D_a(h_t(0),S_t)}=O(\norm a^{k+1})\quad(a\to0).
\label{eq:local}
\end{equation}
The implied constant may depend on the history. The parameter set contains a neighborhood of zero. This definition concerns an asymptotic order, not a fixed-radius numerical error.

For two update directions, the first-order state and sensitivity coefficients obey
\begin{equation}
\begin{bmatrix}h_t^0\\z_{t,1}\\z_{t,2}\end{bmatrix}
=
\begin{bmatrix}A_0&0&0\\A_1&A_0&0\\A_2&0&A_0\end{bmatrix}
\begin{bmatrix}h_{t-1}^0\\z_{t-1,1}\\z_{t-1,2}\end{bmatrix}
+\begin{bmatrix}Bx_t\\0\\0\end{bmatrix}.
\label{eq:first-lift}
\end{equation}
The three blocks can contain redundant historical information. With $d=2$ and scalar input, a controllable example has only four reachable coordinates in this six-coordinate array. Its anchor accounts for two, leaving two auxiliary coordinates rather than four. Appendix~\ref{app:round2-reachable} gives explicit noncommuting matrices, exact ranks through degree three, and a numerical realization.

Stack all coefficients of degrees zero through $k$ into $j_t$. Its exact recurrence is $j_t=\mathsf T_kj_{t-1}+\mathsf B_kx_t$, with $A_0$ on every diagonal block and $A_i$ from block $\alpha-e_i$ to block $\alpha$. The input acts only on the constant block. Let $n_k=d\binom{p+k}{k}$, let $\pi_0$ extract that block, and define the reachable coefficient space
\begin{equation}
\mathcal K_k=\sum_{s=0}^{n_k-1}\im(\mathsf T_k^s\mathsf B_k).
\label{eq:reachable-main}
\end{equation}
\begin{theorem}[Minimum causal local memory]\label{thm:causal-minimal}
For real-vector inputs, the minimum number of continuous causal auxiliary coordinates beyond the anchor that supports~\eqref{eq:local} at every finite horizon is
\begin{equation}
\boxed{\nu_k=\dim\mathcal K_k-\dim(\pi_0\mathcal K_k).}
\label{eq:causal-dimension}
\end{equation}
A linear causal encoder attains this minimum, even when the competing decoders are unrestricted.
\end{theorem}
Histories that share a summary must agree in every coefficient through degree $k$ to satisfy local accuracy. The antipodal argument gives a lower bound once the coefficient system is fully reached. For the upper bound, choose a linear complement to $\pi_0$ on $\mathcal K_k$, retain its coordinates, and update them before any query. Appendix~\ref{app:causal-minimal} proves both directions. The minimum concerns exact local order, not numerical conditioning or update cost.

\begin{theorem}[A rank-one direction can be worst case]\label{thm:rankone}
For every $d,k\geq1$, a uniformly contractive scalar-input family
\begin{equation}
h_s(a)=(A+a vv^\top)h_{s-1}(a)+bu_s
\label{eq:rankone}
\end{equation}
requires $kd$ additional continuous coordinates for~\eqref{eq:local} at horizon $d(k+1)$. Retaining its first $k$ coefficient vectors attains this count.
\end{theorem}
Choose $A$ diagonal with distinct eigenvalues in $(0,1)$ and all entries of $v,b$ nonzero. Its coefficient lift has $A$ on each diagonal block and $vv^\top$ immediately below. Controllability makes its $d(k+1)$ coordinates independent functions of a sufficiently long history. The anchor retains only $d$. Appendix~\ref{app:rankone} proves this by the PBH criterion. This is an existence result, not a claim about every rank-one family.

The same $kd$ bound suffices for a scalar-input model permitting \emph{arbitrary} transition replacements.
\begin{corollary}[Sharp direction-independent scalar-input memory]\label{cor:universal}
If $(A_0,b)$ is controllable and $x_t$ is scalar, $kd$ auxiliary real coordinates suffice for order-$k$ local migration under every matrix replacement $A_0+\Delta$. The causal encoder can be chosen from $(A_0,b,k)$ before $\Delta$ or any update directions are known. No smaller worst-case count is possible. In particular, for every finite affine family, $\nu_k\leq kd$, independently of $p$.
\end{corollary}
Let $P$ be the characteristic polynomial of $A_0$. The matrix $P(\mathsf T_k)$ strictly increases coefficient degree, so $P(\mathsf T_k)^{k+1}=0$. A scalar input can therefore reach at most $d(k+1)$ lifted dimensions, regardless of the number of monomials. Subtracting the controllable anchor gives $kd$.

To construct a direction-independent encoder, put $p(z)=\det(I-zA_0)$ and $J(z)=\operatorname{adj}(I-zA_0)$. The order-$k$ response to $\Delta$ has transfer function
\begin{equation}
H_k(z,\Delta)=\frac{\sum_{\ell=0}^k[J(z)z\Delta]^\ell J(z)b\,p(z)^{k-\ell}}{p(z)^{k+1}}.
\label{eq:universal-main}
\end{equation}
Its numerator degree is less than $d(k+1)$ and its denominator is independent of $\Delta$. One scalar filter therefore supports all replacement directions. A coordinate transformation incorporates the ordinary anchor and leaves $kd$ auxiliary coordinates. Appendix~\ref{app:universal} proves the construction and truncation bound and discusses conditioning. The information result assumes exact arithmetic. It does not justify using a high-degree companion filter in a large model; a smaller declared family can reduce implementation cost.

\section{Fixed-horizon tolerance and precision}\label{sec:information}
Throughout, vector and operator norms are Euclidean. The continuous lower bounds apply to real histories and need not extend to finite-alphabet tokens. For a fixed horizon, write the old state as $R_0u$, with $u\in\B_2^n$, and stack the target maps into $Ru$. Let $P_0=I-R_0^\dagger R_0$. The auxiliary encoder $S:\B_2^n\to\R^m$ must be continuous. This excludes discontinuous infinite-precision encodings of an entire history into one real. The decoder need not be continuous.

\begin{theorem}[State side information]\label{thm:width}
With singular-value lists extended by zeros,
\begin{equation}
\inf_{S\ \mathrm{continuous},D}\ \sup_{\norm u\leq1}
\norm{Ru-D(R_0u,S(u))}=\sigma_{m+1}(RP_0).
\label{eq:width}
\end{equation}
In particular, exact recovery requires and is attained by
\begin{equation}
\rank\begin{bmatrix}R_0\\R\end{bmatrix}-\rank R_0
\label{eq:rank}
\end{equation}
additional coordinates at that horizon.
\end{theorem}
The old state supplies $R(I-P_0)u$. Leading right-singular coordinates recover the remaining component optimally. For the lower bound, Borsuk--Ulam on a sphere in $\ker R_0$ gives indistinguishable antipodal histories with separated targets. This is a specialization of classical width arguments, not a new topological method. Appendix~\ref{app:width} gives the full proof. The attaining sketch need not admit an inexpensive causal update.

Normalize a finite query set by $R_Q=J^{-1/2}[R_1^\top,\ldots,R_J^\top]^\top$. Its Euclidean error is RMS error over the $J$ queries. The next statement turns exact rank into an accuracy requirement.
\begin{theorem}[Tolerance and auxiliary bits]\label{thm:bits}
For $\epsilon>0$, the minimum continuous auxiliary dimension for worst-history RMS error at most $\epsilon$ is
\begin{equation}
 m_\epsilon=\#\{i:\sigma_i(R_QP_0)>\epsilon\}.
\label{eq:effective-dimension}
\end{equation}
If instead the auxiliary message has at most $2^b$ values, even a discontinuous encoder requires
\begin{equation}
 b\geq\sum_i\log_2^+\!\left(\frac{\sigma_i(R_QP_0)}{\epsilon}\right).
\label{eq:bits}
\end{equation}
The old state is supplied exactly in both statements.
\end{theorem}
The bit bound follows by covering the conditional target ellipsoid with $2^b$ radius-$\epsilon$ balls. Appendix~\ref{app:finite-tolerance} gives the proof and an explicit quantized upper bound. Per-query error $\epsilon$ implies the RMS condition, but the converse need not hold. Neither statement identifies a finite-bit, causal optimum over a continuum of queries.

Exact recovery can be much more demanding. A contractive scalar recurrence $h_s(a)=(\lambda+a)h_{s-1}(a)+u_s$ can require $t-1$ extra coordinates to recover every allowed target exactly. Evaluations at $t$ distinct parameters recover the entire history through a Vandermonde system (Proposition~\ref{prop:scalar}). Small conditional singular values can nevertheless make far fewer coordinates sufficient at a nonzero tolerance.

\section{Structured memory and output preservation}\label{sec:method}
For time-invariant matrices the smallest $A_0$-invariant space containing the update images is
\begin{equation}
\calV_* =\spn\{A_0^j\im(A_i):0\leq j<d,\ 1\leq i\leq p\}.
\label{eq:closure}
\end{equation}
Its dimension can be $d$ for a rank-one update. Conversely, a low-dimensional space permits coefficient propagation without realizing the entire lift. This construction is sufficient even when the retained coefficients contain redundant information.

\subsection{A fixed-anchor coefficient memory}
Consider the possibly input-dependent affine recurrence
\begin{equation}
h_t(a)=\left(A_{0,t}+\sum_i a_iUD_{i,t}\right)h_{t-1}(a)+b_t,
\quad \norm a_1\leq R,
\label{eq:family}
\end{equation}
where $U\in\R^{d\times r}$ is orthonormal and fixed before the stream. The matrices and injections may depend causally on inputs, but \emph{not} on $a$. Write
\begin{equation}
T_t=U^\top A_{0,t}U,\quad G_{i,t}=D_{i,t}U,\quad E_t=A_{0,t}U-UT_t.
\label{eq:reduced}
\end{equation}
For $\calI_k=\{\alpha:1\leq|\alpha|\leq k\}$, maintain an anchor and reduced polynomial coefficients, initialized to zero,
\begin{align}
h_t^0&=A_{0,t}h_{t-1}^0+b_t,\label{eq:anchor}\\
z_{t,\alpha}&=T_tz_{t-1,\alpha}+\sum_{i:\alpha_i>0}
\begin{cases}D_{i,t}h_{t-1}^0,&\alpha=e_i,\\G_{i,t}z_{t-1,\alpha-e_i},&|\alpha|>1.
\end{cases}\label{eq:jets}
\end{align}
Every right-hand side uses quantities from the previous time step. The resulting update-closed memory (UCM) answers a query with
\begin{equation}
\widehat h_t^{(k)}(a)=h_t^0+U\sum_{\alpha\in\calI_k}a^\alpha z_{t,\alpha}.
\label{eq:query}
\end{equation}
Under exact closure, $E_t=0$, these are the exact coefficients of $a^\alpha$. The array uses $rN_{p,k}$ scalars, where $N_{p,k}=\binom{p+k}{k}-1$, and degree $k\geq t-1$ recovers the state exactly (Proposition~\ref{prop:coeff}).
These are derivatives divided by $\alpha!$, and mixed coefficients include every ordering of noncommuting matrices. The array can exceed the minimum $\nu_k$. For a scalar anchor block and scalar input, different noncommuting update words can share the same history moments, as shown in Appendix~\ref{app:shared-moments}.

Beyond an independently maintained active $d$-state model, generic auxiliary storage is $d+rN_{p,k}$ scalars. The first $d$ store the anchor and incur additional storage after a deployment change. Exact closure allows sharing that reduces extra storage to $r+rN_{p,k}$. A scalar certificate adds one coordinate. For fixed dense matrices, maintenance beyond the anchor costs $O(prd+pr^2N_{p,k}+rN_{p,k})$ operations and a query costs $O(rN_{p,k}+dr+pN_{p,k})$. Preprocessing, input-dependent reductions, and shared calibration are charged separately in Appendix~\ref{app:algorithm}.

\subsection{Finite-radius error and a streaming certificate}
Assume
\begin{equation}
\norm{A_{0,t}}\leq\rho<1,\quad\max_i\norm{D_{i,t}}\leq L,\quad
\norm{b_t}\leq B,\quad\norm{E_t}\leq\delta,
\label{eq:assumptions}
\end{equation}
and set $\kappa=\rho+LR<1$ and $q=LR/(1-\rho)$. These norm assumptions are stronger than eigenvalue stability \citep{miller2019}.
\begin{theorem}[Truncation and leakage]\label{thm:uniform}
Uniformly in the horizon and in $\norm a_1\leq R$,
\begin{equation}
\norm{h_t(a)-\widehat h_t^{(k)}(a)}
\leq\frac{B}{1-\kappa}q^{k+1}
+\frac{\delta LRB}{(1-\rho)(1-\kappa)^2}.
\label{eq:uniform}
\end{equation}
\end{theorem}
The first term is a geometric coefficient tail. The second bounds the error of the untruncated projected response. Increasing degree leaves the leakage term unchanged. This upper bound does not imply an unavoidable positive error floor.

For a path-dependent bound that can be tighter, define
\begin{equation}
M_{t,\ell}=\max_{|\alpha|=\ell}\frac{\norm{z_{t,\alpha}}}{\binom{\ell}{\alpha}}.
\label{eq:degree-mass}
\end{equation}
Choose upper bounds $\kappa_t$ on the family transition norm, $\delta_t$ on $\norm{E_t}$, and $\ell_t$ on $\max_i\norm{G_{i,t}}$.
\begin{theorem}[Radius-uniform residual certificate]\label{thm:certificate}
With $c_0=0$, the causal update
\begin{equation}
c_t=\kappa_t c_{t-1}+\delta_t\sum_{j=1}^kR^jM_{t-1,j}
+\ell_tR^{k+1}M_{t-1,k}
\label{eq:certificate}
\end{equation}
bounds $\sup_{\norm a_1\leq R}\norm{h_t(a)-\widehat h_t^{(k)}(a)}$ at every finite horizon.
\end{theorem}
The one-step defect consists of retained-response leakage and an omitted degree-$(k+1)$ term. Propagating it gives an exact-arithmetic guarantee, not validated floating-point certification. Each later query uses the original memory and its current bound, so earlier migration errors do not accumulate in that memory. Changing the anchor, injection, or update family is a different problem.

Contraction also helps replay. If $H=B/(1-\kappa)$, replaying a retained suffix of length $L_{\rm buf}$ from zero has error at most $H\kappa^{L_{\rm buf}}$. Appendix~\ref{app:decision} compares this sufficient buffer budget with~\eqref{eq:uniform}. Degree must be provisioned before the stream. Selecting a lower-degree answer from a previously maintained larger array does not retrospectively save memory or maintenance.

\subsection{Preserving future outputs}\label{sec:outputs}
Current-output equality does not imply future-output equality. For a fixed $A,C$, the kernel of $\mathcal O_{A,C}=[C^\top,(CA)^\top,\ldots,(CA^{d-1})^\top]^\top$ identifies states with equal future outputs under common inputs. Theorem~\ref{thm:width} applies to the corresponding stacked output maps.

A causal reduction valid under later parameter switches requires a \emph{common} quotient. If $Q$ has orthonormal rows, $QA_i=\bar A_iQ$ for every $i=0,\ldots,p$, and $C=\bar C Q$, then $Qh_t$ is an exact recurrent state for those outputs. Under update closure $U$, only $\rank(QU)N_{p,k}$ coefficient scalars are needed (Corollary~\ref{cor:output-memory}). Row-space closure constructs $Q$ before streaming. This does not cover arbitrary nonlinear heads. Appendix~\ref{app:outputs} also bounds cumulative predictive KL under common future inputs, not ground-truth loss or closed-loop generated trajectories.

\section{Experiments}\label{sec:experiments}
The experiments separate exact local dimension, finite-tolerance fidelity, predictive usefulness, and complete cost. Each reported mean and sample SD aggregates within a fitted-model seed first. Streams and updates are paired observations within a seed, not independent training runs. The source, retained measurements, and new diagnostics are included with the experimental materials.

\subsection{Dimension is not a tolerance budget}
The four-state rank-one example has the $kd=8$ local requirement for $k=2$. At horizon $12$ and query radius $0.2$, however, it needs $\SpectralDimensions$ continuous coordinates at RMS tolerances $10^{-2},10^{-3},10^{-4},10^{-6},10^{-8}$. In particular, one coordinate suffices at $10^{-4}$. The spectrum, precision bounds, and normalization are given in Appendix~\ref{app:revision-spectra}. Independent null-space and projector calculations agree within $1.1\times10^{-16}$.

A separate dense noncommuting study has $d=24,r=4,p=2$, five model seeds, and $21{,}600$ queries spanning degree, radius, leakage, and history length. Observed small-radius slopes are $2.00,2.99,3.99,4.99$ for degrees one through four. There are no sampled certificate violations with the declared $10^{-10}$ tolerance, but leakage can prevent certification at every retained degree. Appendix~\ref{app:revision-certificates} reports this failure rate and certificate conservatism. These finite numerical checks do not establish validated floating-point certification.

\subsection{Equal-memory language migration}
Five trained 64-state byte models undergo eight learned replacements on each of eight held-out streams. The family modifies eight separable coordinates through two fixed directions. Calibration, training, tuning, and evaluation histories are disjoint. A direct input-to-logit path weakens dependence on the migrated state. This experiment tests state reconstruction, not a general need for migration in language modeling (Appendix~\ref{app:language}).
\begin{table}[t]
\caption{Language migration over 320 events per method. Mean $\pm$ sample SD across five models. Extra bytes exclude the common active state. Predictive KL averages the first 32 post-update positions.}
\label{tab:language}
\centering\small\input{tables/revision_language_main}
\end{table}

The cubic Chebyshev bank is fixed before queries and matches degree-three coefficients' 256-byte budget. Mean state NRMSE is $\LanguageCubicError$, versus $\LanguageDegreeThreeError$ for degree three, with lower error in $\CubicWinningSeeds$ of five seeds. Neither method changes predictive loss meaningfully. Recorded state-only stream medians are $\LanguageCubicMs$ ms for the bank and $\LanguageDegreeThreeMs$ ms for coefficients, with IQRs $\LanguageCubicIQR$ and $\LanguageDegreeThreeIQR$ ms. The overlapping IQRs do not establish a reliable speed ordering. Appendix~\ref{app:round2-paired} shows the paired seed outcomes.

Using fitted seeds as the paired units gives a two-sided $t$-test $p=\PairedLanguageP$ and a sign-flip $p=\PairedLanguageFlipP$. These five models do not establish an interpolation advantage.

\subsection{Useful adaptation and competing history representations}
The coupled study fits a 128-state scalar-input model with a known injection and fixed known readout, without a direct input skip. Full-state transitions identify an anchor. Six independent pilot domains determine two update directions and an eight-dimensional image through a fixed singular-value energy rule. Four new domains then provide independent calibration data for fitting replacements. Matrices are noncommuting and mixed coefficients are nonzero. The image is estimated, but the synthetic true family deliberately occupies a slow invariant block and the readout lies in that block. The task is retrospective identification with observed calibration states, not latent-state fine-tuning.

Each of five seeds has eight held-out streams of 4096 inputs, with replacements every 512 inputs. The resulting 280 events per method are evaluated over 32 subsequent positions. Task NMSE measures squared prediction error against each generating model's clean conditional mean over the full prefix. Restricted exact replay reaches $2.50\times10^{-9}$ NMSE, close to unrestricted fits at $2.01\times10^{-9}$ using the same calibration data. The permitted family therefore retains the useful adaptation available to the unrestricted fit in this setting.
\begin{table}[t]
\caption{Coupled task. Prediction NMSE is mean $\pm$ SD across five fitted seeds. Extra storage excludes the common 1024-byte active state but includes a retained anchor when required. Times include one complete stream, seven migrations, initialization, and readouts. Shared setup is separate. A dash is untimed, not free.}
\label{tab:task}
\centering\small\input{tables/revision_task_main}
\end{table}

Stale reuse has mean NMSE $\TaskStaleNmse$, compared with $\TaskAnchorNmse$ without adaptation. Degree three reaches $\TaskDegreeThreeNmse$. Seed 4 contributes $\DominantSeedShare\%$ of the summed degree-three seed NMSE, making the paired outcomes important for interpreting the mean. The analytic Laguerre history bank \citep{wahlberg1991} matches replay's prediction error at the same 1600-byte budget and is faster on the measured stream. Its shared decoder depends on the replacement dynamics and injection, but not on the readout or prediction labels. It requires setup but no historical replay. Appendix~\ref{app:revision-task} gives the basis, exact-arithmetic residual formula, and complete calibration protocol.

\subsection{How much of the history bank is needed?}
We evaluate the prespecified widths $1,2,4,8,16,32,64,128,200,400,800$ on the same five fitted systems. Each bank processes the stream before targets are revealed, with pole fixed at $0.995$. Eight float64 coordinates, or 64 extra bytes, suffice for output NRMSE below $10^{-4}$ in every seed, while mean \emph{full-state} NRMSE remains $\HistoryEightStatePercent\%$. Output fidelity here is relative to the fitted replacement's exact replay, rather than the clean target used for task NMSE. Table~\ref{tab:width} uses pooled event-window error within each seed and requires the tolerance in all five seeds, not at every individual event.

To test the repeated-pole advantage, a paired structural variant replaces the true slow block $0.995I_8$ by eight poles equally spaced from $0.955$ to $0.995$. All models and update families are refitted from independent calibration data using the same procedure and seeds. Adaptation remains useful, but the eight-coordinate bank's maximum seed output NRMSE rises to $\SplitEightOutputMaximum$. Among tested widths, 64 coordinates are now needed at $10^{-4}$. This deterioration concerns the prescribed history basis, not every migration method. Degree-three coefficients have mean task NMSE $\SplitDegreeThreeNmse$ in the variant. The full protocol and all widths appear in Appendix~\ref{app:round2-width}.

At output NRMSE tolerance $10^{-3}$, degree-three memory and a 32-coordinate bank both qualify, using 1600 and 256 extra bytes per stream. At $10^{-4}$, degree three fails the all-five-seed criterion and a 64-coordinate bank succeeds. Appendix~\ref{app:round2-split} reports the stitching comparison and shared-storage costs.
\begin{table}[t]
\caption{Smallest tested history-bank width meeting each tolerance in every fitted seed. ``Repeated'' and ``split'' refer to the true slow poles. Each coordinate costs eight per-stream bytes. These are empirical widths for one prescribed basis, not minimum-memory theorems.}
\label{tab:width}
\centering\small\input{tables/round2_tolerances}
\end{table}

\begin{figure}[t]
\centering\includegraphics[width=\linewidth]{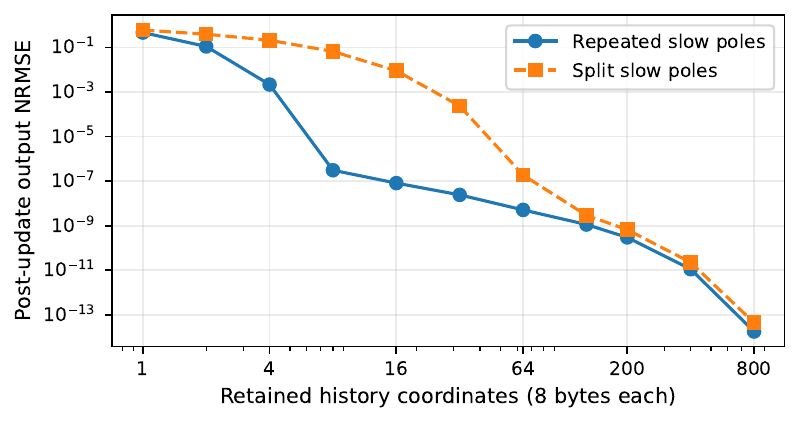}
\caption{Maximum seed-level output NRMSE across the five fitted systems at each prescribed bank width. Inputs, replacement schedules, and calibration seeds are paired between structural variants. Each ordinate is the maximum of five pooled within-seed window errors, not a worst-query or population bound.}
\label{fig:width-output}
\end{figure}

\subsection{Maintenance, regimes, and numerical realization}
Whole-stream times include every input, retained-memory maintenance, migrations, initialization, and readouts. In the coupled instance, 21 timing repetitions give medians of $24.67$ ms for degree three, $30.50$ ms for replay, and $8.78$ ms for the 200-coordinate analytic bank. Its four decoders occupy 819,200 shared bytes and take $4.83$ ms to construct, versus $0.57$ ms for coefficient reductions. The timings use one fitted model and one stream, not five timing seeds. Separately measured medians cannot be added into a joint latency quantile. The new width sweep makes no new runtime claim.

Appendix~\ref{app:revision-costs} gives total cost as a function of stream count and query frequency, including shared setup and decoder storage.

Appendix~\ref{app:revision-sweeps} reports the contraction and update-frequency sweeps. Relative leakage, history exposure, and domain mixture also change across settings. The comparisons are therefore descriptive rather than isolated effects of either factor. On the two-state example, an orthogonal reachable-space realization uses the exact local minimum of two, four, or six extra coordinates at degrees one through three. Its maximum polynomial-response NRMSE is $\ReachableFloatError$ in float32 and $\ReachableDoubleError$ in float64. Appendix~\ref{app:round2-reachable} reports exact ranks, numerical rank thresholds, closure residuals, and conditioning. This small fixed-family check does not supply a general finite-precision theorem for the direction-independent construction.

Appendix~\ref{app:numerical-precision} bounds accumulated arithmetic error for a specified realization. The bound identifies conditioning quantities to assess before choosing a minimal rather than a larger representation.

\section{Discussion}\label{sec:discussion}
The causal coefficient space identifies what must be remembered for exact local migration. For controllable scalar-input models, unknown arbitrary matrix directions and a suitable rank-one direction have the same sharp worst-case $kd$ cost. Neither coefficient-array size nor update rank alone determines the number of independent historical distinctions.

Representation choice remains separate. Interpolation, coefficients, and orthogonal history banks have different maintenance and approximation costs. The experiments show how the readout and pole structure affect that comparison.  The bit bound conditions on an exact old state; residual certificates concern exact arithmetic; general upstream updates can invalidate the injection model. The language task shows negligible predictive benefit, and both coupled variants retain full-state calibration and planted structure. A memory-width study on a naturally trained linear recurrent model with transition-only rank-one adaptation remains future work. The information theorem remains an exact local result; practical migration must also account for these restrictions.

%% file: tables/revision_language_main.tex
\begin{tabular}{lrrr}
\toprule
Method & Extra bytes & State NRMSE & Predictive KL \\
\midrule
Stale state & 0 & $(2.95\pm0.79)\times10^{-2}$ & $(2.93\pm0.98)\times10^{-6}$ \\
Checkpoint replay, 256 & 320 & $(6.76\pm5.24)\times10^{-3}$ & $(6.72\pm7.12)\times10^{-8}$ \\
Linear stitching & 0 & $(2.33\pm0.54)\times10^{-3}$ & $(4.91\pm0.76)\times10^{-9}$ \\
First-order sensitivity & 128 & $(1.40\pm0.41)\times10^{-2}$ & $(3.64\pm0.74)\times10^{-7}$ \\
Degree-2 memory & 192 & $(3.01\pm1.39)\times10^{-3}$ & $(1.32\pm1.29)\times10^{-8}$ \\
Degree-3 memory & 256 & $(7.49\pm5.50)\times10^{-4}$ & $(1.10\pm1.83)\times10^{-9}$ \\
Cubic interpolation & 256 & $(5.30\pm3.46)\times10^{-4}$ & $(2.78\pm2.27)\times10^{-10}$ \\
\bottomrule
\end{tabular}

%% file: tables/revision_task_main.tex
\begin{tabular}{lrrr}
\toprule
Method & Extra bytes & Prediction NMSE & Stream ms \\
\midrule
No adaptation & 0 & $(1.34\pm2.34)\times10^{-1}$ & \textemdash \\
Unrestricted fit + replay & $4t$ & $(2.01\pm1.62)\times10^{-9}$ & \textemdash \\
Full replay & $4t$ & $(2.50\pm2.95)\times10^{-9}$ & $30.50$ \\
Stale state & 0 & $(3.75\pm6.65)\times10^{-1}$ & $6.95$ \\
Checkpoint replay, 256 & 2048 & $(2.26\pm3.47)\times10^{-2}$ & $9.78$ \\
Linear stitching & 0 & $(1.98\pm4.29)\times10^{-4}$ & $7.06$ \\
First-order sensitivity & 1152 & $(8.38\pm17.50)\times10^{-3}$ & $18.18$ \\
Degree-3 memory & 1600 & $(2.15\pm4.54)\times10^{-5}$ & $24.67$ \\
Cubic interpolation & 1600 & $(1.73\pm3.33)\times10^{-4}$ & $30.41$ \\
Analytic history bank & 1600 & $(2.50\pm2.95)\times10^{-9}$ & $8.78$ \\
\bottomrule
\end{tabular}

%% file: tables/round2_tolerances.tex
\begin{tabular}{rrrrr}
\toprule
NRMSE tolerance & \multicolumn{2}{c}{Output fidelity} & \multicolumn{2}{c}{Full-state fidelity} \\
 & Repeated & Split & Repeated & Split \\
\midrule
$10^{-2}$ & 4 & 16 & 128 & 200 \\
$10^{-3}$ & 8 & 32 & 400 & 400 \\
$10^{-4}$ & 8 & 64 & 800 & 800 \\
\bottomrule
\end{tabular}

%% file: sections/statements.tex
\section*{Ethics statement}
This study concerns preservation of a model's computation, not a guarantee that its outputs are accurate, fair, or safe. A replay-free summary is not necessarily privacy-preserving and should not be treated as a substitute for deletion or access-control policies. The experiments use synthetic numerical streams and an existing small news-text corpus, with no newly recruited human participants. The underlying news articles are not redistributed. No deployment in a consequential decision system is evaluated.

\section*{AI use statement}
Generative AI was used for grammatical checks and proofreading. It was also used for reference/literature search and verification. The author assumes full responsibility for the content.

\section*{Reproducibility statement}
The supplement contains the implementations, trained models, numerical results, and calibration data. Complete proofs and experimental protocols are given in the appendices.

%% file: sections/proofs.tex
\section{Information limits with an existing state}\label{app:width}
The appendices give proofs of every theoretical statement in the main text. The encoder continuity assumption is imposed on the auxiliary history summary. Unless explicitly stated otherwise, no regularity is imposed on the decoder. The arguments therefore also apply to continuous online memories with arbitrary nonlinear migration rules. A fixed-horizon bound does not by itself imply that its attaining summary can be maintained using bounded computation at earlier times.

\subsection{The topological ingredient}
We use the following standard form of the Borsuk--Ulam theorem \citep{matousek2003}.

\begin{lemma}[Borsuk--Ulam]\label{lem:bu}
For every $m\geq0$ and every continuous map $f:S^m\to\R^m$, there is $v\in S^m$ such that $f(v)=f(-v)$.
\end{lemma}

For $m=0$, the codomain is the singleton $\R^0$, so the assertion holds for the two points of $S^0$. We use this classical theorem without reproving it and give all remaining steps below.

\subsection{Proof of Theorem~\ref{thm:width}}
Let $r_0=\rank(R_0)$ and $M=RP_0$. The Moore--Penrose identities imply that $R_0^\dagger R_0$ is the orthogonal projector onto $(\ker R_0)^\perp$, so $P_0$ is the orthogonal projector onto $\ker R_0$. Every history decomposes as
\begin{equation}
u=R_0^\dagger R_0u+P_0u,
\qquad Ru=RR_0^\dagger R_0u+Mu.
\label{eq:side-decompose}
\end{equation}
The first target component is therefore exactly recoverable from the old state without an auxiliary sketch.

Take a singular value decomposition $M=\sum_{j=1}^{r_M}\sigma_j w_jv_j^\top$, where $r_M=\rank(M)$ and $\sigma_1\geq\cdots\geq\sigma_{r_M}>0$. Put $s=\min(m,r_M)$ and choose
\begin{equation}
S(u)=(v_1^\top u,\ldots,v_s^\top u,0,\ldots,0)\in\R^m,
\qquad
D(y,z)=RR_0^\dagger y+\sum_{j=1}^s\sigma_j w_jz_j.
\end{equation}
This sketch and decoder are linear. Their error is
\begin{equation}
Ru-D(R_0u,S(u))=\sum_{j>s}\sigma_j w_jv_j^\top u.
\end{equation}
Its supremum on the unit ball is $\sigma_{m+1}$ under the zero-padding convention. This proves attainability, including the case $m\geq r_M$.

For the lower bound, it suffices to consider $\sigma_{m+1}>0$. A positive right-singular vector of $M$ lies in $\ker R_0$, because
$M^\top M=P_0R^\top RP_0$. Define the $(m+1)$-dimensional subspace
$V=\spn(v_1,\ldots,v_{m+1})\subseteq\ker R_0$. For an arbitrary continuous sketch, its restriction to the unit sphere of $V$ is a continuous map to $\R^m$. Lemma~\ref{lem:bu} provides $u\in V$ with $\norm u=1$ and $S(u)=S(-u)$. Both histories have old state zero. Thus, if $y=D(0,S(u))$, the triangle inequality gives
\begin{equation}
\max\{\norm{Ru-y},\norm{-Ru-y}\}
\geq \frac12\norm{(Ru-y)-(-Ru-y)}
=\norm{Ru}
\geq\sigma_{m+1}(M).
\end{equation}
The last inequality follows from $u\in V$ and $P_0u=u$. This holds for every decoder, whether or not it is continuous or measurable, and proves~\eqref{eq:width}.

Exact recovery is possible precisely when $m\geq\rank(M)$. To identify this rank, restrict $R$ to $\ker R_0$. Rank--nullity on that subspace gives
\begin{align}
\rank(RP_0)
&=\dim\ker R_0-\dim(\ker R_0\cap\ker R)\\
&=(n-r_0)-\left(n-\rank\begin{bmatrix}R_0\\R\end{bmatrix}\right),
\end{align}
which is~\eqref{eq:rank}. This completes the proof. \hfill$\square$

\paragraph{Scope of the information bound.}
The lower bound is deterministic and worst-case. If a random sketch must satisfy the guarantee for every realization, the same argument applies after conditioning on that realization. The theorem does not establish the same formula for expected error under randomized encoders, distribution-dependent loss, or discontinuous finite-bit encodings. For a target stack $R=(R_1^\top,\ldots,R_K^\top)^\top$, the error is the square root of the sum of squared target errors. Replacing this by a maximum across targets changes the optimization problem and is not covered by the singular-value formula.

The no-memory case has a simple exact interpretation. Setting $m=0$ in~\eqref{eq:rank}, exact recovery of $Ru$ from $R_0u$ alone is possible if and only if $\ker R_0\subseteq\ker R$. Equivalently, $R=LR_0$ for some linear $L$. Thus allowing a nonlinear decoder does not improve exact recovery over unrestricted linear histories. On a restricted data distribution, a learned nonlinear decoder can still improve average error, as the experiments illustrate.

\subsection{Proof of Proposition~\ref{prop:scalar}}\label{app:scalar}
\begin{proposition}[Exact recovery can retain the entire past]\label{prop:scalar}
The scalar recurrence $h_s(a)=(\lambda+a)h_{s-1}(a)+u_s$, starting from zero, can be contractive on an arbitrarily narrow interval around zero while requiring $t-1$ additional continuous coordinates for exact recovery of $h_t(a)$ at every permitted $a$.
\end{proposition}
Take $\lambda=1/2$ and $I=[-\eta,\eta]$ for any $0<\eta<1/2$. Then $|\lambda+a|\leq1/2+\eta<1$ for every permitted $a$, including intervals as narrow as desired. Unrolling the recurrence yields
\begin{equation}
h_t(a)=\sum_{s=1}^t(\lambda+a)^{t-s}u_s.\label{eq:scalar}
\end{equation}
Choose $t$ distinct points $a_1,\ldots,a_t\in I$, one of which is $0$. The map from $u$ to $(h_t(a_1),\ldots,h_t(a_t))$ has rows
$((\lambda+a_j)^{t-1},\ldots,\lambda+a_j,1)$. Reversing the columns gives a Vandermonde matrix. Its determinant, up to sign, is
$\prod_{i<j}(a_j-a_i)$, which is nonzero. The stacked target map therefore has rank $t$, while the old-state row has rank one. Theorem~\ref{thm:width} implies that any continuous summary supporting exact recovery for all $a\in I$ needs at least $t-1$ extra coordinates, since it must recover these $t$ particular queries.

The bound is attainable at that horizon by keeping $u_1,\ldots,u_{t-1}$ and recovering $u_t$ from the old state, whose coefficient on $u_t$ is one. The size of this summary grows with $t$. Narrowing the interval can make the Vandermonde matrix poorly conditioned but does not change its exact rank. The proposition is therefore an exact-information obstruction, not a claim that approximate migration on a small interval always needs nearly the whole history. \hfill$\square$

\section{A local lower bound for rank-one updates}\label{app:rankone}
We first prove the controllability criterion used in the dimension argument.

\begin{lemma}[A finite-dimensional controllability criterion]\label{lem:pbh}
For $T\in\R^{N\times N}$ and $b\in\R^N$, the matrix
$[b,Tb,\ldots,T^{N-1}b]$ has rank $N$ if and only if there is no nonzero complex row vector $w$ and complex scalar $\lambda$ with $wT=\lambda w$ and $wb=0$.
\end{lemma}
\begin{proof}
If such a row exists, then $wT^jb=\lambda^jwb=0$ for every $j$, so the controllability matrix cannot have full rank. Conversely, suppose its column span $\mathcal K$ is proper. By Cayley--Hamilton, $T\mathcal K\subseteq\mathcal K$. Its nonzero annihilator, after complexification, is invariant under the row-vector map $w\mapsto wT$. Every linear operator on a nonzero finite-dimensional complex vector space has an eigenvector. Choose such a row in the annihilator. It satisfies $wb=0$ and the desired eigenvector equation.
\end{proof}

\subsection{Proof of Theorem~\ref{thm:rankone}}
Fix $d,k\geq1$. Let $A=\diag(\lambda_1,\ldots,\lambda_d)$, with all $\lambda_i$ distinct and in $(0,1)$. Choose $v,b\in\R^d$ with every entry nonzero and set $D=vv^\top$. In particular, $D$ has rank one and $D_{ii}=v_i^2>0$. For any
\begin{equation}
0<\varepsilon<\frac{1-\norm A}{\norm D},
\end{equation}
all transitions $A+aD$ with $|a|\leq\varepsilon$ have norm at most $\norm A+\varepsilon\norm D<1$.

Write $H_{s,j}(u)$ for the coefficient of $a^j$ in $h_s(a;u)$, with $H_{s,0}=h_s(0;u)$. Substitution into the recurrence gives
\begin{align}
H_{s,0}&=AH_{s-1,0}+bu_s,\\
H_{s,j}&=AH_{s-1,j}+DH_{s-1,j-1},\qquad 1\leq j\leq k.
\end{align}
The lift has the explicit form
\begin{equation}
\mathsf T_k=\begin{bmatrix}
A&0&\cdots&0\\D&A&\ddots&\vdots\\0&\ddots&\ddots&0\\0&0&D&A
\end{bmatrix},\qquad \mathsf b=(b^\top,0,\ldots,0)^\top.
\label{eq:lift}
\end{equation}
Thus the stacked coefficients form a linear system with transition $\mathsf T_k$ and input $\mathsf b$ from~\eqref{eq:lift}, of dimension $N=d(k+1)$.

We prove this lifted system is controllable. Its eigenvalues are the $\lambda_i$, since it is block triangular with diagonal blocks $A$. Let $w=(w_0,\ldots,w_k)$ be a left eigenvector at $\lambda_i$, with each $w_j$ a row in $\mathbb C^d$. The block equations are
\begin{equation}
w_j(A-\lambda_iI)+w_{j+1}D=0\quad(0\leq j<k),
\qquad w_k(A-\lambda_iI)=0.
\label{eq:left-eigen}
\end{equation}
Distinctness of the eigenvalues gives $w_k=c_ke_i^\top$. Multiply the equation for $j=k-1$ on the right by $e_i$. Its first term is zero, so $c_kD_{ii}=0$ and hence $w_k=0$. The last nontrivial equation then implies $w_{k-1}=c_{k-1}e_i^\top$. The equation one block earlier forces $c_{k-1}D_{ii}=0$. Repeating gives
$w_1=\cdots=w_k=0$ and $w_0=c_0e_i^\top$. Since $w\ne0$, $c_0\ne0$, and therefore
$w\mathsf b=c_0b_i\ne0$. Lemma~\ref{lem:pbh} proves controllability.

At horizon $N$, the input-to-coefficient map is the reversed controllability matrix
\begin{equation}
(H_{N,0},\ldots,H_{N,k})
=\bigl[\mathsf T_k^{N-1}\mathsf b,\ldots,\mathsf b\bigr]u,
\end{equation}
which is invertible. In particular, the old-state map has rank $d$ and its kernel $V$ has dimension $N-d=kd$.

Suppose a continuous sketch $S:\B_2^N\to\R^m$ with $m<kd$ supports~\eqref{eq:local}. Choose any $(m+1)$-dimensional subspace of $V$. By Lemma~\ref{lem:bu}, its unit sphere contains $u$ with $S(u)=S(-u)$, and both histories have the same zero old state. Let $y(a)$ be their common decoded estimate. The two local error guarantees imply
\begin{equation}
\norm{h_N(a;u)-h_N(a;-u)}
\leq\norm{h_N(a;u)-y(a)}+\norm{h_N(a;-u)-y(a)}
=O(|a|^{k+1}).
\label{eq:local-pair}
\end{equation}
Because the recurrence is linear in the history, the left difference is $2h_N(a;u)$. Invertibility of the coefficient map implies that a nonzero $u\in V$ has some nonzero coefficient among $H_{N,1},\ldots,H_{N,k}$. Let $j\leq k$ be the smallest such order. Then
$a^{-j}h_N(a;u)\to H_{N,j}(u)\ne0$ as $a\to0$, contradicting~\eqref{eq:local-pair}. This proves $m\geq kd$ without requiring differentiability of $S$ or any regularity of $D_a$.

Finally, retaining the $k$ coefficient vectors and returning
$h_N(0;u)+\sum_{j=1}^ka^jH_{N,j}(u)$ attains order-$k$ local accuracy with $kd$ extra scalars. For fixed $N$, $h_N(a;u)$ is a finite polynomial, so its remaining terms are $O(|a|^{k+1})$. This finishes the proof. \hfill$\square$

\paragraph{Sharpness and scope.}
For each $d,k$, the theorem constructs a family attaining the bound. It does not require $kd$ additional coordinates for every rank-one family or make multivariate coefficient arrays minimal. Rank one, scalar parameterization, and uniform contraction are jointly insufficient to reduce the worst-case $kd$ local-information cost. The input is real-valued and ranges over an open ball. Discrete language histories do not inherit this continuous-domain lower bound automatically.

\section{Invariant closure and the streaming construction}\label{app:closure}
\subsection{The smallest structural invariant space}
For time-invariant matrices $A_0,A_1,\ldots,A_p$, define $\calV_*$ by~\eqref{eq:closure}. Every $\im(A_i)$ is contained in $\calV_*$. Applying $A_0$ to a generator with exponent $j<d-1$ produces another generator. Applying it at exponent $d-1$ produces $A_0^d\im(A_i)$, which is contained in $\calV_*$ by Cayley--Hamilton. Therefore $\calV_*$ is $A_0$-invariant. Conversely, any $A_0$-invariant subspace containing every $\im(A_i)$ contains $A_0^j\im(A_i)$ for all $j$, hence contains $\calV_*$. This proves the claimed minimality.

Choose an orthonormal basis $U$ of $\calV_*$. Then $A_0U=U(U^\top A_0U)$ and $A_i=U(U^\top A_i)$. Thus the family has the form~\eqref{eq:family} with $D_i=U^\top A_i$ and zero leakage. This construction gives a sufficient space for all possible anchor states. It may exceed what is necessary when injections visit only a restricted reachable set or when only certain outputs are relevant. For time-varying families, a common invariant space may instead require closure under several anchor transitions. Formula~\eqref{eq:closure} is asserted only for the time-invariant case.

\subsection{Proof of Proposition~\ref{prop:coeff}}
\begin{proposition}[Exact retained coefficients]\label{prop:coeff}
If $E_t=0$ at every time, $Uz_{t,\alpha}$ is the exact coefficient of $a^\alpha$ in $h_t(a)$. The coefficient array uses $rN_{p,k}$ scalars, where $N_{p,k}=\binom{p+k}{k}-1$. At horizon $t$, degree $k\geq t-1$ gives exact recovery.
\end{proposition}
For a fixed finite history, the state in~\eqref{eq:family} is a polynomial in $a$. Since the initial state is zero and $b_t$ is parameter-independent, its degree is at most $t-1$. Write
\begin{equation}
h_t(a)=h_t^0+\sum_{\alpha\ne0}a^\alpha H_{t,\alpha},
\end{equation}
with nonexistent coefficients understood as zero. Matching powers in~\eqref{eq:family} gives the constant recurrence~\eqref{eq:anchor} and
\begin{equation}
H_{t,\alpha}=A_{0,t}H_{t-1,\alpha}
+\sum_{i:\alpha_i>0}
\begin{cases}
UD_{i,t}h_{t-1}^0,&\alpha=e_i,\\
UD_{i,t}H_{t-1,\alpha-e_i},&|\alpha|>1.
\end{cases}
\label{eq:full-coeff}
\end{equation}
All positive-degree coefficients start at zero. Assume inductively that $H_{t-1,\alpha}=Uz_{t-1,\alpha}$. If $E_t=0$, then $A_{0,t}U=UT_t$, and substituting into~\eqref{eq:full-coeff} gives exactly $U$ times~\eqref{eq:jets}. This proves the identity at time $t$. The induction applies independently of commutativity and does not depend on higher-degree coefficients. Consequently, truncating storage at degree $k$ leaves every retained coefficient exact.

The number of multi-indices of degree $j$ in $p$ variables is $\binom{p+j-1}{j}$. Summing from $j=1$ to $k$ gives $N_{p,k}=\binom{p+k}{k}-1$. Each reduced coefficient has $r$ coordinates, establishing the storage count. If $k\geq t-1$, no nonzero coefficient has been omitted, so~\eqref{eq:query} is exact. \hfill$\square$

\paragraph{Normalization and mixed terms.}
For $p=2$, the degree-two equations include
\begin{align}
z_{t,(2,0)}&=T_tz_{t-1,(2,0)}+G_{1,t}z_{t-1,(1,0)},\\
z_{t,(1,1)}&=T_tz_{t-1,(1,1)}+G_{1,t}z_{t-1,(0,1)}+G_{2,t}z_{t-1,(1,0)},\\
z_{t,(0,2)}&=T_tz_{t-1,(0,2)}+G_{2,t}z_{t-1,(0,1)}.
\end{align}
There is no factor of two in the repeated-coordinate equations because $z_{(2,0)}$ represents the coefficient, not the unnormalized second derivative. The mixed equation includes both orderings, without assuming commutativity.

\section{Uniform truncation and leakage bounds}\label{app:uniform}
\subsection{The projected response}
The projected coefficient recurrence remains defined when $E_t\ne0$. Define an untruncated reduced response by
\begin{equation}
z_t(a)=\left(T_t+\sum_i a_iG_{i,t}\right)z_{t-1}(a)
+\left(\sum_i a_iD_{i,t}\right)h_{t-1}^0,
\qquad z_0(a)=0.
\label{eq:full-projected}
\end{equation}
At any finite $t$ this is a finite polynomial, and its positive-degree coefficients are precisely~\eqref{eq:jets}. Define $\bar h_t(a)=h_t^0+Uz_t(a)$. This trajectory is used only in the analysis; the algorithm does not maintain it for unknown $a$.

From~\eqref{eq:assumptions} and orthonormality of $U$,
\begin{equation}
\norm{h_t^0}\leq\frac{B}{1-\rho},\qquad
\norm{T_t}\leq\rho,\qquad
\norm{G_{i,t}}\leq L.
\end{equation}
Equation~\eqref{eq:full-projected} then gives
\begin{equation}
\norm{z_t(a)}\leq\kappa\norm{z_{t-1}(a)}+\frac{LRB}{1-\rho}
\leq\frac{LRB}{(1-\rho)(1-\kappa)}.
\label{eq:z-bound}
\end{equation}
To compare with the true state, write $A_t(a)=A_{0,t}+\sum_i a_iUD_{i,t}$. Direct substitution shows
\begin{equation}
A_t(a)\bar h_{t-1}(a)+b_t-\bar h_t(a)=E_tz_{t-1}(a).
\end{equation}
Hence, for $e_t=h_t(a)-\bar h_t(a)$,
\begin{equation}
e_t=A_t(a)e_{t-1}+E_tz_{t-1}(a),\qquad e_0=0.
\end{equation}
Using~\eqref{eq:z-bound} and $\norm{A_t(a)}\leq\kappa$ gives
\begin{equation}
\norm{e_t}\leq\frac{\delta LRB}{(1-\rho)(1-\kappa)^2}.
\label{eq:leak-bound}
\end{equation}

\subsection{A multinomial coefficient bound}
\begin{lemma}\label{lem:coeff-bound}
For every nonzero multi-index $\alpha$, every time $t$, and $j=|\alpha|$,
\begin{equation}
\norm{z_{t,\alpha}}
\leq\binom{j}{\alpha}\frac{L^jB}{(1-\rho)^{j+1}}.
\label{eq:coeff-bound}
\end{equation}
\end{lemma}
\begin{proof}
Set $C_\alpha=\binom{j}{\alpha}L^jB/(1-\rho)^{j+1}$. All coefficients are zero at time zero. Suppose all previous-time coefficients satisfy the bound. When $j=1$, the forcing in~\eqref{eq:jets} has norm at most $LB/(1-\rho)=(1-\rho)C_\alpha$. The inherited term has norm at most $\rho C_\alpha$. Their sum is at most $C_\alpha$.

For $j>1$, use
\begin{equation}
\sum_{i:\alpha_i>0}\binom{j-1}{\alpha-e_i}
=\sum_{i:\alpha_i>0}\frac{\alpha_i}{j}\binom{j}{\alpha}
=\binom{j}{\alpha}.
\end{equation}
The total forcing norm is bounded by
\begin{equation}
L\sum_{i:\alpha_i>0}C_{\alpha-e_i}
=\binom{j}{\alpha}\frac{L^jB}{(1-\rho)^j}
=(1-\rho)C_\alpha.
\end{equation}
Adding $\rho C_\alpha$ from the propagated coefficient proves the induction. No cancellation or sign assumption is used.
\end{proof}

\subsection{Proof of Theorem~\ref{thm:uniform}}
For any $a$ in the declared ball, the multinomial theorem and Lemma~\ref{lem:coeff-bound} imply
\begin{align}
\sum_{|\alpha|=j}|a^\alpha|\norm{z_{t,\alpha}}
&\leq \frac{L^jB}{(1-\rho)^{j+1}}
\sum_{|\alpha|=j}\binom{j}{\alpha}\prod_i|a_i|^{\alpha_i}\\
&=\frac{B}{1-\rho}
\left(\frac{L\norm a_1}{1-\rho}\right)^j
\leq\frac{B}{1-\rho}q^j.
\end{align}
Since $U$ is an isometry on $\R^r$, the difference between the untruncated projected state and its degree-$k$ truncation satisfies
\begin{equation}
\norm{\bar h_t(a)-\widehat h_t^{(k)}(a)}
\leq\frac{B}{1-\rho}\sum_{j=k+1}^{\infty}q^j
=\frac{B}{1-\kappa}q^{k+1}.
\label{eq:tail-proof}
\end{equation}
For finite $t$, the actual tail is finite, and the infinite sum is only an upper bound. Combining~\eqref{eq:tail-proof} with~\eqref{eq:leak-bound} proves~\eqref{eq:uniform}. The condition $\kappa<1$ implies $q<1$, justifying the geometric series. \hfill$\square$

\paragraph{Degree selection.}
Let $\varepsilon$ exceed the leakage bound $E_{\mathrm{leak}}$ in~\eqref{eq:leak-bound}. For $0<q<1$, any integer $k\geq1$ satisfying
\begin{equation}
k+1\geq
\frac{\log\bigl((\varepsilon-E_{\mathrm{leak}})(1-\kappa)/B\bigr)}{\log q}
\end{equation}
is sufficient whenever $B>0$. If the logarithmic ratio is negative, the inequality is automatically satisfied by all sufficiently small positive degrees as well. When $L=0$ or $B=0$, the parameter-dependent response is zero and no positive-degree coefficients are necessary. If the requested tolerance is at or below the leakage bound, increasing degree alone cannot certify that tolerance using this theorem. Failure of this bound to certify the tolerance does not imply that the tolerance is unattainable on the given stream.

\section{Online residuals and successive updates}\label{app:certificate}
\subsection{Proof of Theorem~\ref{thm:certificate}}
For $j\geq1$, write
$z_{t}^{[j]}(a)=\sum_{|\alpha|=j}a^\alpha z_{t,\alpha}$ and
$w_t^{(k)}(a)=\sum_{j=1}^kz_t^{[j]}(a)$. By the definition of $M_{t,j}$ and the multinomial theorem,
\begin{equation}
\norm{z_t^{[j]}(a)}
\leq M_{t,j}\sum_{|\alpha|=j}\binom{j}{\alpha}|a^\alpha|
=M_{t,j}\norm a_1^j
\leq R^jM_{t,j}.
\label{eq:mass}
\end{equation}
Define $G_t(a)=\sum_i a_iG_{i,t}$. Summing the retained recurrences shows that the one-step residual of the decoded polynomial is exactly
\begin{align}
r_t(a)
&=A_t(a)\widehat h_{t-1}^{(k)}(a)+b_t-\widehat h_t^{(k)}(a)\\
&=E_tw_{t-1}^{(k)}(a)+UG_t(a)z_{t-1}^{[k]}(a).
\label{eq:residual-exact}
\end{align}
All terms of degrees one through $k$ cancel by~\eqref{eq:jets}. The second term is the omitted degree-$(k+1)$ contribution, while the first is off-subspace propagation of every retained degree.

Using $\norm{G_t(a)}\leq R\ell_t$ in~\eqref{eq:residual-exact} and applying~\eqref{eq:mass} gives
\begin{equation}
\norm{r_t(a)}
\leq\delta_t\sum_{j=1}^kR^jM_{t-1,j}
+\ell_tR^{k+1}M_{t-1,k}.
\end{equation}
The migration error $d_t(a)=h_t(a)-\widehat h_t^{(k)}(a)$ satisfies
$d_t(a)=A_t(a)d_{t-1}(a)+r_t(a)$ and $d_0(a)=0$. Taking the supremum over the ball, bounding the transition by $\kappa_t$, and inducting on $t$ proves~\eqref{eq:certificate}. The proof uses only finite sums and does not require contraction. \hfill$\square$

\paragraph{Numerical certification.}
Equation~\eqref{eq:certificate} is not, by itself, a validated floating-point certificate. If an implementation can supply a rigorous upper bound $\eta_t$ on the additional one-step defect caused by rounded arithmetic and rounded coefficient updates, the same induction gives the valid update $c_t=\kappa_tc_{t-1}+\text{residual bound}+\eta_t$. Obtaining $\eta_t$ and certified operator-norm bounds requires additional numerical analysis or interval arithmetic. The numerical experiments use ordinary float64 arithmetic. Agreement with independently replayed states at the stated tolerance does not control the rounding defect, so it does not justify setting $\eta_t=0$.

\subsection{Repeated queries do not re-center the approximation}\label{app:successive}
Let $\tau_1<\tau_2<\cdots$ be update times and let $a_j$ be any permitted replacement coordinate revealed at $\tau_j$. The coordinate can depend on observations up to that point. The radius-uniform guarantees already hold for every such coordinate. At the switch, set the active state to $\widehat h_{\tau_j}^{(k)}(a_j)$ while continuing to maintain the same anchor and coefficients. For $s$ subsequent inputs before the next switch, exact execution of the new recurrence gives
\begin{equation}
\norm{h_{\tau_j+s}^{\mathrm{active}}-h_{\tau_j+s}(a_j)}
\leq\left(\prod_{v=1}^s\norm{A_{\tau_j+v}(a_j)}\right)c_{\tau_j}
\leq\kappa^s c_{\tau_j}
\end{equation}
under the uniform assumptions. At time $\tau_{j+1}$ the next query uses the anchor memory, not this active approximation. Its error is therefore bounded by $c_{\tau_{j+1}}$, independently of the number of preceding switches.

This statement depends on retaining the original anchor family. Translating a finite Taylor polynomial to a new center does not recover the omitted coefficients and is not exact re-centering of the underlying historical response. Nor does the statement recover the trajectory generated by applying the old parameters to old epochs and the new parameters only to new epochs. The target is the retrospective constant-parameter trajectory defined in Section~\ref{sec:lower}.

\section{Observable equivalence and predictive fidelity}\label{app:outputs}
\subsection{Exact future-output equivalence}
Consider the target system $h_{s+1}=Ah_s+Bu_{s+1}$ with output $y_s=Ch_s$, and two initial states $h,h'$. Under any common future input sequence their state difference is $A^s(h-h')$, so their output difference is $CA^s(h-h')$. Equality of all current and future outputs implies equality for $s=0,\ldots,d-1$, hence $\mathcal O_{A,C}(h-h')=0$.

Conversely, if these first $d$ output differences vanish, the characteristic polynomial identity expresses $A^d$ as a linear combination of $I,A,\ldots,A^{d-1}$. Multiplying successively by $A$ shows that every later power is in the same linear span. Therefore $CA^s(h-h')=0$ for every $s\geq0$. This proves the equivalence. The input matrix and input sequence need not satisfy additional restrictions because their effects cancel in the state difference.

For a finite set of target models with history-to-state maps $R_j$, define
\begin{equation}
R_{\mathrm{obs}}=
\begin{bmatrix}
\mathcal O_{A_1,C_1}R_1\\
\vdots\\
\mathcal O_{A_K,C_K}R_K
\end{bmatrix}.
\end{equation}
The exact-information dimension is
$\rank([R_0^\top,R_{\mathrm{obs}}^\top]^\top)-\rank(R_0)$ by Theorem~\ref{thm:width}. The decoder may first recover the observable vector for target $j$, then choose any state realizing it, for example $\mathcal O_{A_j,C_j}^\dagger$ times that vector. Because the observable vector belongs to the range of the observability matrix, this produces an output-equivalent state. The approximate singular-value formula concerns this stacked observable-vector norm, not an unweighted infinite-horizon output norm.

\paragraph{Current output is not enough.}
Let
\begin{equation}
A=\begin{bmatrix}0&1\\0&0\end{bmatrix},
\qquad C=\begin{bmatrix}1&0\end{bmatrix},
\qquad e=\begin{bmatrix}0\\1\end{bmatrix}.
\end{equation}
Then $Ce=0$ but $CAe=1$. A presently invisible state difference changes the next output. In contrast, if $A$ is diagonal and $C$ reads only its first coordinate, all other coordinates are unobservable for this fixed model. A history direction discarded by the old model may then be unnecessary for output-equivalent migration even when exact hidden-state recovery would require retaining it. These claims concern a fixed target during subsequent execution. Requiring equivalence under arbitrary later model switches can demand a larger space involving products of different transitions.

\subsection{Predictive KL and its proof}
For a fixed $\kappa$-contractive target under common future inputs, $L_g$-Lipschitz logits, and an initial state-error bound $c_t$, we have
\begin{equation}
\sum_{s\geq0}\KL(p_{t+s}^{\rm replay}\|p_{t+s}^{\rm migrated})
\leq\frac{L_g^2c_t^2}{4(1-\kappa^2)}.
\label{eq:kl}
\end{equation}
Let $\psi(z)=\log\sum_i e^{z_i}$ and $p(z)=\nabla\psi(z)$. Its Hessian is
$\nabla^2\psi(z)=\diag(p)-pp^\top$. The absolute row sum at row $i$ is
\begin{equation}
p_i(1-p_i)+\sum_{j\ne i}p_ip_j=2p_i(1-p_i)\leq\frac12.
\end{equation}
The Hessian is symmetric positive semidefinite, so its operator norm is at most $1/2$. The integral Taylor formula therefore implies
\begin{equation}
\psi(z')-\psi(z)-\nabla\psi(z)^\top(z'-z)
\leq\frac14\norm{z'-z}^2.
\end{equation}
The left side is $\KL(p(z)\|p(z'))$. With the same subsequent inputs, contraction bounds the state discrepancy at offset $s$ by $\kappa^sc_t$. If the logit map is $L_g$-Lipschitz in the state, the logit discrepancy is at most $L_g\kappa^sc_t$. Thus
\begin{equation}
\KL(p_{t+s}^{\mathrm{replay}}\|p_{t+s}^{\mathrm{migrated}})
\leq\frac{L_g^2}{4}\kappa^{2s}c_t^2.
\end{equation}
Summing the geometric series proves~\eqref{eq:kl}. A direct input-to-logit term is harmless when both trajectories receive the same inputs, since it cancels in the comparison. For the experimental head, a valid state-Lipschitz constant is $\norm{W_o}\norm{W_h}$ because coordinatewise $\tanh$ is 1-Lipschitz.

The KL comparison is an expectation under the replay model's distribution. Ground-truth test labels need not follow that distribution exactly. Consequently the difference in empirical negative log-likelihood can have either sign, and lower KL need not imply better test loss. We report both quantities to distinguish state fidelity from task improvement.

\section{A first-order extension to smooth recurrent models}\label{app:nonlinear}
The higher-order algorithm in the main text relies on an affine hidden-state recurrence. The following first-order result extends the analysis to smooth recurrences, without asserting the same compact higher-order structure for general nonlinear models.

\begin{proposition}[Projected first-order migration]\label{prop:nonlinear}
Let $h_t(a)=F_t(h_{t-1}(a),a)$ with $h_0(a)=0$, for $a$ in a convex Euclidean ball around zero. Assume $F_t$ is twice continuously differentiable on a domain containing the trajectories for every parameter in that ball, with
\begin{equation}
\norm{D_hF_t}\leq\rho<1,\qquad \norm{D_aF_t}\leq L,
\end{equation}
uniformly there. Bound its joint second derivative by $M$ under the product norm $\norm{(v,w)}=\norm v+\norm w$. At the anchor trajectory, set $J_t=D_hF_t(h_{t-1}(0),0)$, $B_t=D_aF_t(h_{t-1}(0),0)$, and $P=UU^\top$. Suppose
\begin{equation}
\norm{(I-P)J_tU}\leq\delta,
\qquad \norm{(I-P)B_t}\leq\beta.
\end{equation}
Maintain $K_0=0$ and $K_t=U^\top J_tUK_{t-1}+U^\top B_t$. Then
\begin{align}
\norm{h_t(a)-h_t(0)-UK_ta}
\leq{}&\left(\frac{\delta L}{(1-\rho)^2}+\frac{\beta}{1-\rho}\right)\norm a\\
&+\frac{M(1+L/(1-\rho))^2}{2(1-\rho)}\norm a^2.
\label{eq:nonlinear-bound}
\end{align}
\end{proposition}
\begin{proof}
Let $S_t(a)=D_ah_t(a)$. Differentiating the recurrence gives
$S_t(a)=D_hF_t\,S_{t-1}(a)+D_aF_t$. Thus $\norm{S_t(a)}\leq L/(1-\rho)$ uniformly. The projected recurrence similarly gives $\norm{K_t}\leq L/(1-\rho)$. At $a=0$, subtract the projected sensitivity from the true one to obtain
\begin{equation}
S_t(0)-UK_t
=J_t\bigl(S_{t-1}(0)-UK_{t-1}\bigr)
+(I-P)J_tUK_{t-1}+(I-P)B_t.
\end{equation}
Its norm is bounded by a geometric recurrence with forcing $\delta L/(1-\rho)+\beta$, yielding
$\norm{S_t(0)-UK_t}\leq\delta L/(1-\rho)^2+\beta/(1-\rho)$.

For unit directions $v,w$, a second differentiation yields an inherited second-state-derivative term with norm at most $\rho\norm{D_a^2h_{t-1}}$ and a joint second-derivative term bounded by
$M(\norm{S_{t-1}v}+1)(\norm{S_{t-1}w}+1)$. Hence
\begin{equation}
\norm{D_a^2h_t(a)}\leq
\frac{M(1+L/(1-\rho))^2}{1-\rho}.
\end{equation}
Taylor's integral remainder along the segment from zero to $a$ bounds
$h_t(a)-h_t(0)-S_t(0)a$ by one half of this constant times $\norm a^2$. Adding the projected-sensitivity discrepancy proves~\eqref{eq:nonlinear-bound}.
\end{proof}

This is a projected sensitivity statement with explicit leakage, not an exact state-invariant nonlinear reduction. Higher derivatives of $F_t$ can generate directions outside $\im U$ even when first-order leakage vanishes. The experiments do not use this extension. It gives no theorem for arbitrary GRU, LSTM, or multilayer selective-model fine-tuning.

%% file: sections/new_proofs.tex
\section{Exact local memory as a causal realization problem}\label{app:causal-minimal}
This section gives a causal counterpart to the fixed-horizon information calculation. It concerns exact local approximation order, not a fixed nonzero error tolerance. Consider the time-invariant family
\begin{equation}
h_t(a)=\left(A_0+\sum_{i=1}^p a_iA_i\right)h_{t-1}(a)+Bx_t,\qquad h_0(a)=0,
\label{eq:general-local-family}
\end{equation}
where $x_t\in\R^q$ and the allowed parameter set contains a neighborhood of zero. Let $\mathcal J_k=\{\alpha\in\mathbb N_0^p:|\alpha|\leq k\}$ and $n_k=d|\mathcal J_k|$. Stack the coefficients through degree $k$ into $j_t\in\R^{n_k}$, including the constant coefficient. Their exact recurrence is
\begin{equation}
j_t=\mathsf T_k j_{t-1}+\mathsf B_kx_t,\quad
(\mathsf T_k)_{\alpha,\beta}=
\begin{cases}
A_0,&\alpha=\beta,\\
A_i,&\alpha=\beta+e_i,\\
0,&\text{otherwise},
\end{cases}
\label{eq:multi-lift}
\end{equation}
with $B$ in the constant block of $\mathsf B_k$ and zeros elsewhere. The cases in~\eqref{eq:multi-lift} are disjoint. Define
\begin{equation}
\mathcal K_k=\sum_{s=0}^{n_k-1}\im(\mathsf T_k^s\mathsf B_k),\qquad
\pi_0j=j_0,\qquad
\nu_k=\dim\mathcal K_k-\dim(\pi_0\mathcal K_k).
\label{eq:reachable-jet}
\end{equation}
The notation $j_0$ in the projection denotes the constant multi-index block, not time zero. Cayley--Hamilton makes $\mathcal K_k$ invariant under $\mathsf T_k$. All coefficient states reached from the zero initial condition lie in this space.

\subsection{Proof of Theorem~\ref{thm:causal-minimal}}
We first prove that local order is an information requirement on the entire coefficient vector. Suppose two fixed histories have the same old state and auxiliary summary, so the decoder returns the same function of $a$ for both histories. If each approximation has error $O(\norm a^{k+1})$ as $a\to0$, their target difference has this order as well. Let $\ell\leq k$ be the least degree with a nonzero coefficient in their difference, and let $P_\ell(a)$ be the corresponding vector-valued homogeneous polynomial. There is a vector $v$ with $P_\ell(v)\ne0$. Otherwise every scalar component of this polynomial would vanish identically, contrary to the choice of $\ell$. Along $a=sv$,
\begin{equation}
h_t(sv;x)-h_t(sv;x')=s^\ell P_\ell(v)+O(|s|^{\ell+1}).
\end{equation}
Dividing by $|s|^\ell$ contradicts the assumed $O(|s|^{k+1})$ difference. Thus all coefficients through degree $k$ agree. The constants in the local error statements may depend on the two histories. They are fixed once those histories are selected, which is all the argument requires.

For the lower bound, take the horizon $n_k$. Its map $F$ from the concatenated input vector in $\R^{qn_k}$ to $j_{n_k}$ has image $\mathcal K_k$, since its block columns are the controllability columns in reverse order. Its old-state map is $\pi_0F$. The restriction of $F$ to $\ker(\pi_0F)$ has rank
\begin{equation}
\rank F-\rank(\pi_0F)=\dim\mathcal K_k-\dim(\pi_0\mathcal K_k)=\nu_k.
\end{equation}
If a continuous auxiliary summary has $m<\nu_k$ coordinates, choose an $(m+1)$-dimensional subspace of $\ker(\pi_0F)$ on which $F$ is injective. Restrict the summary to its unit sphere. Borsuk--Ulam gives antipodal histories $x,-x$ with the same summary and old state zero. Their coefficient difference $2Fx$ is nonzero. The preceding local-order argument rules this out. It follows that $m\geq\nu_k$ for any continuous causal summary that works at every horizon. In fact, this single horizon is already enough for the lower bound.

For attainability, choose a fixed linear map $L:\mathcal K_k\to\R^{\nu_k}$ such that
\begin{equation}
H:j\longmapsto(\pi_0j,Lj)
\end{equation}
is injective on $\mathcal K_k$. Such a map exists by selecting coordinates on $\ker(\pi_0|_{\mathcal K_k})$ and extending them to a direct-sum complement. The inverse of $H$ on its image is linear. Maintain the ordinary anchor $h_t(0)$ and the $\nu_k$ additional coordinates $s_t=Lj_t$. At each new input, recover $j_{t-1}$ by this inverse and set
\begin{equation}
s_t=L(\mathsf T_kj_{t-1}+\mathsf B_kx_t),\qquad
h_t(0)=A_0h_{t-1}(0)+Bx_t.
\end{equation}
This update is causal and independent of any future parameter. Its reconstructed coefficient vector is exact by induction. Evaluate the degree-$k$ polynomial only when the query arrives. At every finite horizon, the omitted polynomial terms have order $O(\norm a^{k+1})$. This proves attainability and the theorem. \hfill$\square$

The characterization applies reachability and state realization to an unknown-query local-accuracy requirement; it introduces no new controllability criterion. A dense basis for $\mathcal K_k$ can be expensive to construct, numerically ill-conditioned, or costly to update. The theorem identifies the minimum continuous state dimension for this exact local objective. It does not optimize runtime or bits, or guarantee uniform accuracy at a fixed radius. If the inputs are confined to a finite alphabet rather than a Euclidean ball, the topological lower bound does not automatically apply.

For the rank-one family of Theorem~\ref{thm:rankone}, $\mathcal K_k=\R^{d(k+1)}$ and $\pi_0\mathcal K_k=\R^d$, so $\nu_k=kd$. For an update-closed family with parameter-independent injection, the nonconstant coefficients lie in $\im U$. Hence $\nu_k\leq rN_{p,k}$ whenever the ordinary anchor is fully reachable. More generally the same bound follows by choosing the $rN_{p,k}$ reduced coefficients as auxiliary coordinates. Equality need not hold. Reachability, vanishing monomials, and output equivalence can all reduce the required memory.

\section{Finite tolerance, finite bits, and query radius}\label{app:finite-tolerance}
\subsection{A normalized finite-query objective}
Fix $J$ permitted parameters $a^{(1)},\ldots,a^{(J)}$ and their linear history maps $R_1,\ldots,R_J$. Define
\begin{equation}
R_Q=J^{-1/2}[R_1^\top,\ldots,R_J^\top]^\top,\qquad
M_Q=R_QP_0,\qquad P_0=I-R_0^\dagger R_0.
\end{equation}
A stacked decoder returns all $J$ target estimates from the same old state and summary. Its Euclidean error is the root mean squared error over the queries. Theorem~\ref{thm:width} applies directly to $R_Q$. Consequently, the number of auxiliary continuous coordinates needed to achieve worst-history RMS error at most $\epsilon$ is exactly
\begin{equation}
\#\{i:\sigma_i(M_Q)>\epsilon\}.
\label{eq:finite-m-dimension}
\end{equation}
The inequality is strict because a singular value equal to the tolerance may be omitted.

A uniform guarantee of $\epsilon$ at each query implies the RMS guarantee, so~\eqref{eq:finite-m-dimension} is a necessary dimension for that stronger objective. It is not generally sufficient. Conversely, an RMS guarantee of $\epsilon/\sqrt J$ is sufficient for per-query error $\epsilon$ on this finite set. A guarantee on this finite grid is not equivalent to a guarantee over a continuum of allowed queries.

\subsection{Proof of the finite-bit statement in Theorem~\ref{thm:bits}}
Let the auxiliary encoder have at most $2^b$ possible messages. It can be discontinuous. Condition on old state zero and restrict histories to the unit ball of $\ker R_0$. The set of normalized targets on this fiber is the ellipsoid
\begin{equation}
\mathcal E=M_Q\B_2^n
\end{equation}
whose positive semiaxes are the singular values $\sigma_1\geq\cdots\geq\sigma_s>0$. Every history with a given message has the same decoded target. Thus $2^b$ balls of radius $\epsilon$ must cover $\mathcal E$. Project these balls and the ellipsoid onto the span of the first $j$ left singular vectors, for any $1\leq j\leq s$. The projected balls have volume at most $v_j\epsilon^j$, whereas the ellipsoid has volume $v_j\prod_{i=1}^j\sigma_i$, with $v_j$ the unit-ball volume in $\R^j$. Subadditivity of volume yields
\begin{equation}
2^b\epsilon^j\geq\prod_{i=1}^j\sigma_i.
\end{equation}
Taking logarithms and maximizing over $j$, also allowing $j=0$, gives
\begin{equation}
b\geq\sum_{i=1}^s\log_2^+\!\left(\frac{\sigma_i}{\epsilon}\right),
\qquad \log_2^+x=\max(0,\log_2x).
\label{eq:bit-lower-proof}
\end{equation}
Neither continuity nor measurability of the encoder is required, because the covering centers and balls are finite in number. The old state is still supplied exactly without charge. This is therefore an auxiliary-bit lower bound conditional on exact side information, not a bound on the total precision of a deployed machine.

A quantized upper bound makes the precision requirement explicit. Choose $m$ such that $\sigma_{m+1}\leq\epsilon/2$. If $m=0$, return the known side-information component and use no auxiliary bits. Otherwise retain the output singular coordinates
\begin{equation}
y_i=\sigma_i v_i^\top u\in[-\sigma_i,\sigma_i],\qquad i=1,\ldots,m.
\end{equation}
Cover each interval by a uniform grid with mesh at most $\epsilon/\sqrt m$ including its endpoints, then encode the nearest grid point. There are at most $1+\lceil2\sigma_i\sqrt m/\epsilon\rceil$ grid points. The Euclidean rounding error of the retained vector is at most $\epsilon/2$. The discarded singular coordinates have norm at most $\epsilon/2$. Reconstructing
\begin{equation}
R_QR_0^\dagger R_0u+\sum_{i=1}^m\widetilde y_i w_i
\end{equation}
therefore incurs RMS error at most $\epsilon$ using at most
\begin{equation}
\sum_{i=1}^m\left\lceil\log_2\left(1+\left\lceil\frac{2\sigma_i\sqrt m}{\epsilon}\right\rceil\right)\right\rceil
\label{eq:bit-upper-proof}
\end{equation}
bits. This is a fixed-horizon encoding. Updating quantized singular coordinates online would require a separate roundoff analysis. \hfill$\square$

\subsection{How an exact local obstruction becomes a finite-radius obstruction}\label{app:radius-bridge}
For a scalar parameter at horizon $t$, write $h_t(a;u)=\sum_{\ell=0}^{t-1}a^\ell C_\ell u$. The old map is $C_0$. For $J\geq k$ selected query coordinates, form
\begin{equation}
V_{j\ell}=J^{-1/2}(a^{(j)})^\ell,\qquad
1\leq j\leq J,\quad1\leq\ell\leq k,
\end{equation}
and $C_+=[C_1^\top,\ldots,C_k^\top]^\top$. Suppose $V$ has full column rank. The normalized query map restricted to the old-state kernel satisfies
\begin{equation}
M_Q=(V\otimes I_d)C_+P_0+F_kP_0,
\label{eq:radius-decompose}
\end{equation}
where $F_k$ stacks the terms of degree above $k$. For every index $i$, extending lists of singular values by zeros,
\begin{equation}
\sigma_i(M_Q)\geq
\left[\sigma_{\min}(V)\sigma_i(C_+P_0)-\norm{F_kP_0}\right]_+.
\label{eq:radius-lower}
\end{equation}
To prove this, the inequality $V^\top V\succeq\sigma_{\min}(V)^2I$ gives
$((V\otimes I)C)^\top((V\otimes I)C)\succeq\sigma_{\min}(V)^2 C^\top C$. The eigenvalue min--max principle bounds every singular value of the product from below. The singular-value perturbation inequality, which also follows from the norm min--max characterization, subtracts at most $\norm{F_kP_0}$ after adding the remainder. Nonnegativity gives~\eqref{eq:radius-lower}. If $|a^{(j)}|\leq R$,
\begin{equation}
\norm{F_kP_0}\leq\sum_{\ell=k+1}^{t-1}R^\ell\norm{C_\ell P_0}.
\end{equation}
This follows by applying the triangle inequality to each block and then using the normalized stack. Together with the width theorem,~\eqref{eq:radius-lower} is a finite-radius lower bound for any retained dimension. It can be zero. Narrowing the interval worsens the conditioning of $V$, and reachability alone says nothing about the conditioning of $C_+P_0$. Thus the exact $kd$ theorem does not establish that $kd$ coordinates are needed at an arbitrary practical tolerance.

\subsection{A causal upper bound from a retained suffix}\label{app:suffix}\label{app:decision}
For a target recurrence with $\norm{A_t(a)}\leq\kappa<1$ and $\norm{b_t}\leq B$, all states from zero satisfy $\norm{h_t(a)}\leq H=B/(1-\kappa)$. Replaying only the last $L$ inputs from zero differs from full replay by
\begin{equation}
A_t(a)\cdots A_{t-L+1}(a)h_{t-L}(a),
\end{equation}
whose norm is at most $H\kappa^L$. Hence
\begin{equation}
L\geq\left\lceil\frac{\log(\epsilon/H)}{\log\kappa}\right\rceil
\label{eq:suffix-budget}
\end{equation}
suffices when $0<\epsilon<H$. An old-model checkpoint has an additional initial error, so the zero-initialized bound does not directly apply to checkpoint replay. The experiments use the stronger checkpoint baseline and measure its actual error.

For coefficient memory, write the leakage term in~\eqref{eq:uniform} as $Hf$ with
\begin{equation}
f=\frac{\delta LR}{(1-\rho)(1-\kappa)}.
\end{equation}
For a normalized tolerance $\tau=\epsilon/H$, the global certificate can succeed only if $f<\tau$, except for degenerate zero-tail cases. If $0<q<1$ and $f<\tau<1+f$, a sufficient polynomial degree is
\begin{equation}
k\geq\max\left\{0,\left\lceil\frac{\log(\tau-f)}{\log q}\right\rceil-1\right\}.
\label{eq:degree-budget}
\end{equation}
The case $k=0$ returns the anchor, and the general coefficient formula extends with no nonconstant coefficients. Degree zero is included only in the sufficient-budget diagnostic, not in the residual implementation that assumes $k\geq1$. If $q=0$, the updates vanish and the anchor is exact. These are competing \emph{sufficient} causal budgets with different storage units and maintenance costs. Neither is a minimax characterization at finite tolerance.

\section{Causal memory for a common output quotient}\label{app:output-quotient}
\begin{corollary}[Output-oriented retained memory]\label{cor:output-memory}
Suppose $Q$ has orthonormal rows, $QA_i=\bar A_iQ$ for $i=0,\ldots,p$, and $C=\bar C Q$. Then $Qh_t$ is an exact recurrent state for all outputs, including under subsequent parameter switches. If the original family has exact update closure $U$, its reduced coefficient memory needs only $r_{\rm out}N_{p,k}$ scalars, where $r_{\rm out}=\rank(QU)\leq r$.
\end{corollary}

Consider time-invariant matrices $A_0,A_1,\ldots,A_p$, an input matrix $B$, and a fixed linear readout $C$. Let $Q\in\R^{q\times d}$ have orthonormal rows and suppose
\begin{equation}
QA_i=\bar A_iQ\quad(0\leq i\leq p),\qquad C=\bar C Q.
\label{eq:common-quotient}
\end{equation}
These conditions mean that $\ker Q$ is invariant under every transition component and invisible to the readout. This is a stronger condition than matching the present output or being unobservable for one fixed target parameter.

\subsection{Proof of Corollary~\ref{cor:output-memory}}
Multiplying the full recurrence by $Q$ gives the exact reduced recurrence
\begin{equation}
\bar h_t(a)=\left(\bar A_0+\sum_i a_i\bar A_i\right)\bar h_{t-1}(a)+QB x_t,
\qquad \bar h_t(a)=Qh_t(a).
\end{equation}
The output is $C h_t(a)=\bar C\bar h_t(a)$. This identity continues to hold when the parameter changes between subsequent inputs, because~\eqref{eq:common-quotient} holds separately for each transition component.

Suppose $A_i=UD_i$ for $i\geq1$ and $A_0U=UT$. Let $\bar U$ be an orthonormal basis for $\im(QU)$, with $r_{\mathrm{out}}=\rank(QU)$. Since $Q^\top$ is a right inverse of $Q$, the reduced matrices can be taken as $\bar A_i=QA_iQ^\top$. Define
\begin{equation}
\bar D_i=\bar U^\top QUD_iQ^\top.
\end{equation}
Then $\bar A_i=\bar U\bar D_i$, because $QA_iQ^\top$ has image contained in $\im(QU)$. Also
\begin{equation}
\bar A_0(QU)=QA_0U=QUT,
\end{equation}
so $\im\bar U$ is invariant under $\bar A_0$. Proposition~\ref{prop:coeff} therefore applies to the reduced family, retaining exactly its degree-$k$ coefficients in $r_{\mathrm{out}}N_{p,k}$ scalar locations. Differentiating the identity $\bar h_t(a)=Qh_t(a)$ shows that these coefficients are the projections of the full coefficients. The query reproduces the degree-$k$ output expansion without retaining state distinctions in $\ker Q$. This proves the corollary. \hfill$\square$

The reduction retains all input information needed for the outputs. The injection is explicitly $QB x_t$, and the intertwining conditions ensure that discarded state directions never re-enter observable ones. When these conditions fail, simply dropping components of $U$ according to the current readout is invalid.

\subsection{Constructing the common quotient}
Starting with the row space of $C$, repeatedly append the rows of $QA_i$ for every $i=0,\ldots,p$, taking a basis after each step. The row space increases until it is invariant under right multiplication by all the matrices. At most $d$ strict dimension increases are possible. At stabilization, an orthonormal row basis $Q$ satisfies~\eqref{eq:common-quotient}. Its kernel is the largest common invariant subspace contained in $\ker C$. Indeed, its rows are spanned by $C$ followed by finite words in the matrices, and every common invariant subspace invisible to $C$ is annihilated by all these words.

The construction uses simultaneous observability reduction. Its migration-specific consequence is the count of pre-query coefficients and the validity of subsequent outputs under parameter replacements. It is not a general claim that an optimal nonlinear or distribution-dependent output memory has this dimension. In floating point, numerical rank thresholds define an approximate quotient; an exact theorem requires exact intertwining, or additional error terms for its violation.

\section{A sharp scalar-input bound independent of update dimension}\label{app:universal}
This result separates persistent memory from the number of allowed update coordinates. The local-order statement is algebraic and does not provide a precision-independent stable numerical realization.

\subsection{Proof of Corollary~\ref{cor:universal}}
Assume the anchor pair $(A_0,b)$ is controllable and the input is scalar. Let $P$ be the characteristic polynomial of $A_0$, of degree $d$. Group the coefficient lift~\eqref{eq:multi-lift} by total degree. Its block diagonal consists of copies of $A_0$, and every off-diagonal block increases total degree. By Cayley--Hamilton, all diagonal blocks of $P(\mathsf T_k)$ vanish. Hence $P(\mathsf T_k)$ increases the degree filtration by at least one, and
\begin{equation}
P(\mathsf T_k)^{k+1}=0.
\label{eq:annihilating-lift}
\end{equation}
This is a monic annihilating polynomial of degree $d(k+1)$. The reachable Krylov space of a single input has dimension at most that degree, because every higher power of $\mathsf T_k$ applied to its input vector is a linear combination of the first $d(k+1)$ powers. Controllability of the anchor gives $\dim(\pi_0\mathcal K_k)=d$. Theorem~\ref{thm:causal-minimal} now yields
\begin{equation}
\nu_k\leq d(k+1)-d=kd.
\end{equation}
The rank-one family in Theorem~\ref{thm:rankone} attains $kd$, making this a sharp worst-case bound. No commutativity among $A_0,A_1,\ldots,A_p$ was used.

The stronger direction-independent construction below shows that the retained state can be fixed without specifying the update matrices. Let $z$ denote a formal delay variable, set
\begin{equation}
p(z)=\det(I-zA_0),\qquad J(z)=\operatorname{adj}(I-zA_0),
\end{equation}
and let $\Delta$ be the matrix replacement revealed at query time. The transfer kernel of the degree-$k$ Taylor expansion in $\Delta$ is
\begin{align}
H_k(z,\Delta)
&=\sum_{\ell=0}^k (I-zA_0)^{-1}
\bigl[z\Delta(I-zA_0)^{-1}\bigr]^\ell b\\
&=\frac{N_k(z,\Delta)}{p(z)^{k+1}},\label{eq:universal-rational}\\
N_k(z,\Delta)
&=\sum_{\ell=0}^k [J(z)z\Delta]^\ell J(z)b\,p(z)^{k-\ell}.
\label{eq:universal-numerator}
\end{align}
The order of matrix multiplication is fixed as written. Each numerator term has degree at most
$\ell d+(d-1)+(k-\ell)d=d(k+1)-1$.
Thus all possible $\Delta$ share the scalar denominator $p(z)^{k+1}$ and a numerator of degree less than $m=d(k+1)$.

Write $p(z)^{k+1}=1+q_1z+\cdots+q_mz^m$, padding with zeros if necessary. Maintain the scalar filter
\begin{equation}
w_t=x_t-\sum_{j=1}^m q_jw_{t-j},\qquad w_t=0\quad(t\leq0),
\label{eq:universal-filter}
\end{equation}
and its $m$ most recent values. This memory depends only on $A_0$ and $k$. At a query, evaluate the numerator coefficients in~\eqref{eq:universal-numerator} and take their linear combination of these filter values. Formal power-series multiplication proves exact agreement with the degree-$k$ truncated target for every finite history. All filter updates precede the query and require no discarded inputs.

To reduce the additional count from $m$ to $m-d$, let $y_t=(w_t,\ldots,w_{t-m+1})$. The constant-parameter transfer kernel $J(z)b/p(z)$ has a numerator of degree below $m$ when expressed with denominator $p(z)^{k+1}$, so $h_t(0)=Hy_t$ for a fixed matrix $H$. The scalar filter has a reachable companion realization, and anchor controllability implies $\rank H=d$. Choose $m-d$ rows $S$ such that the square matrix $[H^\top,S^\top]^\top$ is invertible. Store $(h_t(0),Sy_t)$ and update this coordinate transform of the companion realization. There are exactly $kd$ auxiliary coordinates. Its query reconstructs $y_t$ linearly and then evaluates~\eqref{eq:universal-numerator}. This completes the proof. \hfill$\square$

\paragraph{Accuracy and conditioning.}
At a fixed horizon, the target is a polynomial in the entries of $\Delta$, so the omitted terms are $O(\norm\Delta^{k+1})$ in a neighborhood of zero. If $\norm{A_0}\leq\rho<1$, $\norm\Delta\leq R<1-\rho$, and $\norm{b x_t}\leq B$, the same insertion-count argument as the coefficient-tail proof gives the horizon-uniform truncation bound
\begin{equation}
\norm{h_t(A_0+\Delta)-\widehat h_t^{(k)}(\Delta)}
\leq \frac{B}{1-\rho-R}\left(\frac{R}{1-\rho}\right)^{k+1}
\label{eq:universal-tail}
\end{equation}
in exact arithmetic. The input filter and coordinate transform can be severely ill-conditioned near repeated poles. Their roundoff is not controlled by~\eqref{eq:universal-tail}. The implementation is therefore tested on small well-conditioned instances, not substituted for the high-dimensional empirical method.

Forming the universal numerator after an update can also be expensive. A declared-family method instead evaluates the retained coefficients directly. The bound is about persistent real-coordinate memory, not about universal superiority in bits, maintenance cost, or query time. In a scalar-input update-closed family, both constructions give the informative upper bound $\nu_k\leq\min\{kd,rN_{p,k}\}$, but either can exceed the actual reachable dimension.

\subsection{Why noncommuting updates can still share a small history basis}\label{app:shared-moments}
For a scalar anchor transition $\lambda I_r$ and scalar injection $b x_t$, let $G(a)=\sum_i a_iG_i$ with arbitrary, possibly noncommuting $G_i$. Since $\lambda I_r$ commutes with $G(a)$,
\begin{equation}
h_t(a)=\sum_{\ell=0}^{t-1}G(a)^\ell b\,m_{t,\ell},\qquad
m_{t,\ell}=\sum_{s=0}^{t-1}\binom{s}{\ell}\lambda^{s-\ell}x_{t-s},
\end{equation}
where terms with $s<\ell$ vanish. The causal update is $m_{t,0}=\lambda m_{t-1,0}+x_t$ and $m_{t,\ell}=\lambda m_{t-1,\ell}+m_{t-1,\ell-1}$. A degree-$k$ response therefore uses $k+1$ scalar history moments, not one vector for every multivariate monomial. If $b\ne0$, the anchor determines $m_{t,0}$. Noncommutativity alone does not make the full coefficient count necessary. This shared-moment structure helps explain the efficiency of the history-basis alternative in the structured system-identification experiment.

%% file: sections/implementation.tex
\section{Algorithm and complete memory accounting}\label{app:algorithm}
\subsection{A causal implementation}
Algorithm~\ref{alg:memory} gives the general update without relying on a particular experimental architecture. The reference implementation uses a separate next-coefficient array for clarity. An implementation can instead update coefficients in decreasing total degree, because every nonconstant forcing term depends only on a lower degree from the previous time. The anchor must be updated last. Certificate statistics are computed from the previous coefficients before any overwrite.

\begin{algorithm}[t]
\caption{Pre-update polynomial memory}
\label{alg:memory}
\begin{algorithmic}[1]
\Require A declared family~\eqref{eq:family}, basis $U$, radius $R$, degree $k\geq1$
\State Enumerate $\calI_k$ and precompute each predecessor $\alpha-e_i$.
\State Initialize $h^0=0$, $z_\alpha=0$ for $\alpha\in\calI_k$, and $c=0$.
\For{each arriving input at time $t$}
  \State Obtain $A_{0,t},D_{i,t},b_t$ without using a future query parameter.
  \State Obtain $T_t,G_{i,t},\kappa_t,\delta_t,\ell_t$ from the declared family.
  \State Compute $M_{t-1,j}$ from~\eqref{eq:degree-mass} and update $c$ by~\eqref{eq:certificate}.
  \For{$\alpha\in\calI_k$ in decreasing total degree}
    \State $v\gets T_tz_\alpha$
    \For{$i$ with $\alpha_i>0$}
      \If{$\alpha=e_i$}
        \State $v\gets v+D_{i,t}h^0$
      \Else
        \State $v\gets v+G_{i,t}z_{\alpha-e_i}$
      \EndIf
    \EndFor
    \State $z_\alpha\gets v$
  \EndFor
  \State $h^0\gets A_{0,t}h^0+b_t$
  \If{a migration query $a$ is revealed}
    \State Check that $\norm a_1\leq R$.
    \State Return $h^0+U\sum_{\alpha\in\calI_k}a^\alpha z_\alpha$ and the bound $c$.
  \EndIf
\EndFor
\end{algorithmic}
\end{algorithm}

An active deployed state can be advanced in parallel with this memory. Under exact closure, represent it as $h^0+Uv$ for a reduced active response $v\in\R^r$. While the deployed coordinate is $a$, update
\begin{equation}
v_t=\left(T_t+\sum_i a_iG_{i,t}\right)v_{t-1}
+\sum_i a_iD_{i,t}h_{t-1}^0.
\label{eq:active-response}
\end{equation}
At a switch, replace $v_t$ by $\sum_\alpha a^\alpha z_{t,\alpha}$. This shares the anchor between the auxiliary memory and deployed state. With leakage,~\eqref{eq:active-response} no longer equals the full-model transition. A generic implementation must then retain a separate $d$-dimensional active state to execute the full target recurrence exactly between switches.

\subsection{Storage and arithmetic}
The generic auxiliary memory alone contains $d+rN_{p,k}+1$ real scalars, including the anchor and radius-uniform certificate. Keeping an additional full active state gives total storage $2d+rN_{p,k}+1$. Under exact closure, sharing the anchor as in~\eqref{eq:active-response} reduces this total to $d+rN_{p,k}+r+1$. Compared with the $d$ scalars of a conventional active state, the extra storage is then $rN_{p,k}+r+1$. Omitting the optional certificate subtracts one scalar. Sparse coefficients can reduce these counts, as in the language experiment.

The model description is shared across streams and is counted separately from per-stream memory. It includes $A_0$, $U$, the $D_i$, and any precomputed reduced matrices and norm bounds. For a dense time-invariant family, computing $A_0U$ and $U^\top A_0U$ costs $O(d^2r+dr^2)$, and computing the $G_i=D_iU$ costs $O(pr^2d)$. Obtaining exact operator norms may require further matrix factorizations. These are offline costs for fixed matrices, not free operations when a family changes with each input.

For a fixed dense family, the anchor transition costs $O(d^2)$. The products $D_i h^0$ cost $O(prd)$ and can be shared between coefficient and active-response updates. Propagating $N_{p,k}$ reduced coefficients and their at most $p$ predecessors costs $O(pr^2N_{p,k})$. Computing all coefficient norms and degree maxima costs $O(rN_{p,k})$. Polynomial evaluation costs $O(pN_{p,k}+rN_{p,k})$ and lifting to a full state costs $O(dr)$. None of these operations requires the discarded history. For time-varying matrices, the cost of producing the reductions must be added unless the model supplies them directly.

\begin{table}[t]
\caption{Number $N_{p,k}$ of nonconstant monomials before structural sparsity. Multiply each entry by the reduced dimension $r$ for coefficient storage.}
\centering
\begin{tabular}{rrrr}
\toprule
Update coordinates $p$ & Degree 1 & Degree 2 & Degree 3\\
\midrule
1&1&2&3\\
2&2&5&9\\
4&4&14&34\\
8&8&44&164\\
\bottomrule
\end{tabular}
\end{table}

Storage grows with $p$, so this construction is intended for a small declared coordinate family. Low-rank matrix parameterizations with freely trainable factors can still have many update coordinates, can change the update image, and need not preserve the chosen $U$. A fixed basis of update matrices is a stricter assumption than unconstrained low-rank fine-tuning.

The implementation is included in the supplement.

%% file: sections/experiments.tex
\section{Experimental protocols and complete results}\label{app:experiments}
We describe the geometry and language experiments, including data partitions, training, migration baselines, and resource accounting. The language experiment uses the Lee corpus and five independently trained models.

\subsection{Dense noncommuting state-recovery experiment}\label{app:geometry}
For each seed $s\in\{0,1,2,3,4\}$, we draw an orthogonal matrix $Q$ from a Gaussian QR factorization using NumPy's default generator with seed $s+7301$. In the $Q$ coordinates, begin with a $24\times24$ diagonal matrix whose diagonal ranges linearly from $0.72$ to $0.96$. Replace its leading $4\times4$ block by $0.94I+0.008Z$, where $Z$ has independent standard normal entries. Fill the top-right block with $0.004$ times independent standard normals. The lower-left block is $\lambda_{\mathrm{leak}}E$, where $E$ is an independent Gaussian matrix normalized to operator norm one and
\begin{equation}
\lambda_{\mathrm{leak}}\in\{0,0.001,0.003,0.01,0.03\}.
\end{equation}
Normalize the entire matrix $M$ to operator norm $0.97$ and set $A_0=QMQ^\top$ and $U=Q_{:,1:4}$. Independently draw two $4\times24$ Gaussian reduced update matrices, scale each to operator norm $0.02$, and rotate their input coordinates by $Q^\top$. The resulting update matrices $UD_i$ are dense and generally do not commute with each other or with $A_0$. The leakage scalar controls the nominal value before normalization. The actual $\delta=\norm{A_0U-UU^\top A_0U}$ is stored for every measurement.

The family radius is $R=0.8$, giving the uniform bounds $\rho=0.97$, $L=0.02$, $\kappa=0.986$, and $q=8/15$. Each of twelve streams receives a correlated injection. Starting from $s_0=0$, draw independent standard Gaussian vectors $\xi_t,\eta_t$ and set
\begin{equation}
s_t=0.95s_{t-1}+0.05\xi_t,
\qquad b_t=0.05\tanh(s_t+0.4\eta_t).
\end{equation}
The coordinatewise nonlinearity makes $\norm{b_t}\leq0.05\sqrt{24}$ deterministically. For a given seed, the same input draws and query direction are used across leakage levels. A two-dimensional Gaussian direction is normalized in $\ell_1$ and scaled to each of the six radii $0.025$, $0.05$, $0.1$, $0.2$, $0.4$, and $0.8$. This parameter is never passed to the memory-update method. The algorithm knows only the declared family and its radius.

The degree-four memory is advanced once per input. Lower-order queries use its retained lower-degree coefficients, which are unaffected by higher degrees. Separate certificate recurrences are advanced for degrees one through four. Reference states are computed by independent full recurrences at the selected query parameters. Errors are measured at prefixes $256$, $1024$, and $4096$. The Cartesian product of five seeds, five leakage levels, three prefixes, four degrees, six query radii, and twelve streams yields exactly $21{,}600$ measurements.

\begin{figure}[t]
\centering\includegraphics[width=\linewidth]{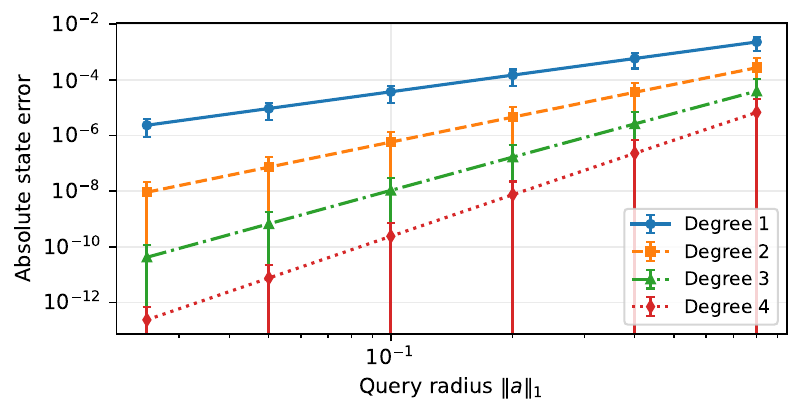}
\caption{Dense noncommuting structural test of approximation order. State error versus query radius at zero leakage. Error bars are the sample SD of five seed-level averages over twelve streams.}
\label{fig:geometry}
\end{figure}

For each plotted setting, the twelve stream errors are averaged within seed, and the five resulting averages determine the plotted mean and sample standard deviation. The slopes in Figure~\ref{fig:geometry} are least-squares slopes of log mean error on log radius using only the three smallest query radii at zero leakage and prefix $4096$. The fitted slopes are $1.9962$, $2.9926$, $3.9913$, and $4.9875$. The highest-order errors at small radii approach floating-point scales; the slopes are not arbitrary-precision estimates.

\begin{figure}[t]
\centering
\includegraphics[width=\linewidth]{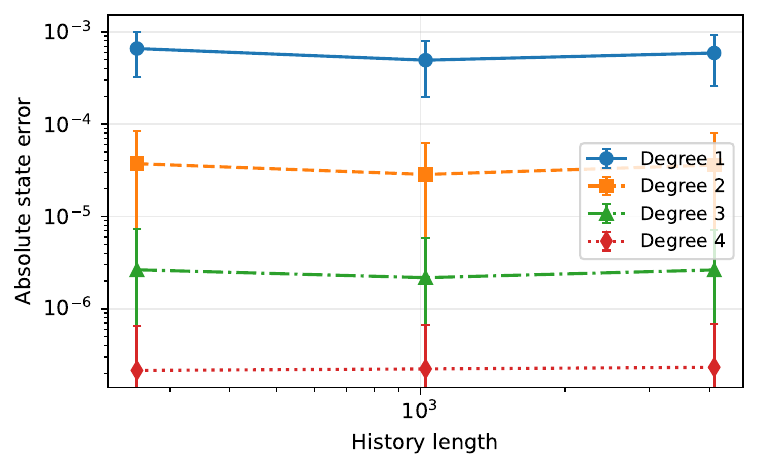}
\caption{State-recovery error versus history length at zero nominal leakage and query radius $0.4$. Error bars are the sample standard deviation of five per-seed stream averages. The guarantee is uniform in the horizon, not monotonic in the horizon. The fluctuations do not contradict a horizon-uniform bound.}
\label{fig:horizon}
\end{figure}

All measured errors lie below both~\eqref{eq:uniform} and~\eqref{eq:certificate}, allowing an absolute numerical tolerance of $10^{-10}$. The online bound uses the full radius $0.8$ even when the query radius is $0.025$, so it is not expected to be tight for every query. Absence of sampled violations does not imply probabilistic coverage. The bounds hold deterministically in exact arithmetic; these comparisons use finite-precision calculations.

\begin{figure}[t]
\centering\includegraphics[width=\linewidth]{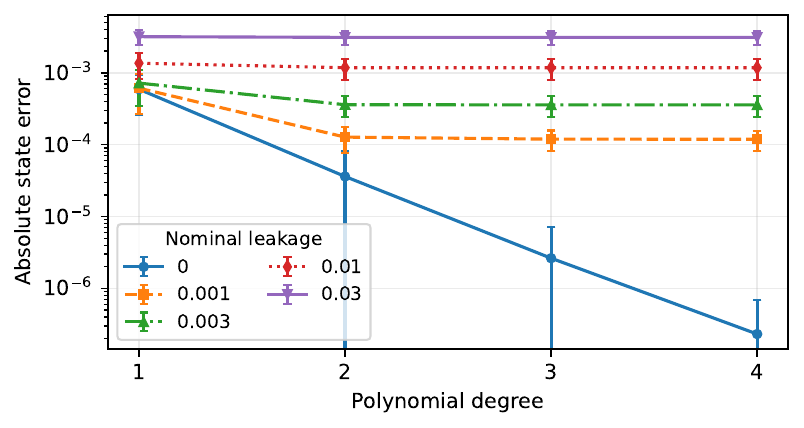}
\caption{Increasing degree under controlled off-subspace leakage. Error bars are the sample SD of five seed-level averages over twelve streams. Displayed leakage is nominal before normalization.}
\label{fig:geometry-leakage}
\end{figure}

\subsection{Language data and architecture}\label{app:language}
We use the Lee news corpus distributed with Gensim \citep{gensimlee}, containing $360{,}082$ bytes and $300$ newline-delimited records. We operate directly on bytes, with all 256 values assigned embeddings and logits. No text decoding, Unicode normalization, vocabulary fitting, or document-level sampling is applied. The four disjoint contiguous ranges, in zero-based half-open byte offsets, are
\begin{equation}
[0,216049),\quad[216049,252057),\quad
[252057,288065),\quad[288065,360082).
\end{equation}
They are used respectively for base training, adapter tuning, stitching calibration, and evaluation. The remaining evaluation bytes beyond the eight selected streams are unused. Corpus access details are provided in the supplement; the source news text is not redistributed.

The recurrence has $d=64$ real states. With a learned embedding $e(x)\in\R^{64}$ and fixed base decays $\lambda_j$, it is
\begin{equation}
h_{t,j}(a)=\bigl(\lambda_j+0.002a_{j\bmod2}\mathbf1\{j<8\}\bigr)h_{t-1,j}(a)
+(1-\lambda_j)e_j(x_t),\qquad j=0,\ldots,63.
\label{eq:lm-recurrence}
\end{equation}
The first eight decays are evenly spaced from $0.980$ to $0.996$, and the other 56 from $0.05$ to $0.95$. The injection always uses the \emph{base} decay, so it is independent of the adapter coordinate. The nonlinear logits are
\begin{equation}
g(h_t,x_t)=W_o\tanh(W_hh_t+b_h+W_se(x_t))+b_o,
\end{equation}
with hidden width 128 and output width 256. The skip matrix $W_s$ has no bias. Base training learns the embedding and the prediction head, not the decay or update-direction buffers. Embeddings are initialized from a zero-mean Gaussian with standard deviation $0.35$ and the linear layers use PyTorch's default initialization. The model is small, and its direct input-to-readout path limits the predictions' dependence on the migrated state.

Each base model is trained from an independent seed in $\{0,1,2,3,4\}$ for 600 AdamW steps, with learning rate $0.002$, weight decay $0.01$, batch size 12, and sequence length 256. The optimizer uses its library-default moment coefficients. Random contiguous training blocks are sampled with replacement. Every block starts from state zero, and cross-entropy excludes its first 64 positions. Gradients are clipped to norm one. Training computes the diagonal recurrence by FFT convolution, while evaluation uses a causal filtering implementation. The saved checkpoints contain the trained embedding and head, together with the fixed recurrence buffers.

After base training, these weights are frozen. Starting from $a=0$, Adam with learning rate $0.035$ performs eight stages of 15 steps, using only the tuning split. Each step samples 12 length-512 blocks, excludes the first 128 positions from its loss, and projects $a$ onto the $\ell_1$ ball of radius $0.75$. This gives eight successive learned coordinates per seed. The learned update schedules are included in the supplement. The projection and training weights use float32, so the stored coordinates can exceed the nominal radius by at most $4.5\times10^{-8}$ in these runs. This excess is due to numerical rounding and does not change the adaptation task. Exact-arithmetic statements about the experiment can use the enclosing radius $0.750001$; no machine-validated certificate is claimed.

\subsection{Migration protocol and baselines}
Each seed is evaluated on eight contiguous held-out streams, each with 8192 inputs and one additional byte for next-byte labels. The last label byte of one stream is the first input byte of the next, but their 8192-input histories do not overlap. Each history starts from state zero. The successive replacement times are
\begin{equation}
1024,\ 2048,\ 3072,\ 4096,\ 5120,\ 6144,\ 7168,\ 7936.
\end{equation}
Between replacements, every baseline advances its own active state with the deployed coordinate. At a replacement, the reference is a full replay of the complete current prefix under the new coordinate. Evaluation copies each migrated state forward for 31 additional inputs, producing the current next-byte prediction and 31 subsequent predictions. These scoring copies do not overwrite the active state or change the next epoch's starting time.

\paragraph{Full replay, reset, and stale reuse.}
Full replay is the exact counterfactual reference. Since only eight coordinates depend on the update, all methods preserve the unaffected 56 coordinates exactly. Reset zeros only the eight affected coordinates, making it a stronger baseline than resetting the entire state. Stale reuse retains its current affected state without transformation. After migration, both baselines continue with the new transition.

\paragraph{Checkpoint replay.}
For a suffix of length $L\in\{64,256\}$, the baseline retains the affected active state at time $t-L$ and the next $L$ raw bytes. When the replacement is revealed at $t$, it replays those bytes from the retained checkpoint under the new coordinate. The checkpoint itself was generated under the previous deployed model and can include error from earlier migrations. The protocol uses neither zero initialization nor an updated-model checkpoint that would require replaying earlier inputs. It assumes update times can be announced $L$ inputs ahead, but does not reveal the new coordinate early.

\paragraph{Linear and nonlinear stitching.}
For each adjacent pair of learned coordinates, exact old and new states are computed from the distinct calibration split. The full 64-dimensional old state is used as a feature, and the target is the correction to its first eight coordinates. States are sampled every four positions after a 512-position warmup, yielding 8874 candidate pairs. Only the first 80\% of these pairs are used for fitting, and the final 20\% are unused. No hyperparameters are selected on either those unused pairs or the test streams.

Features are centered and scaled by their fitting-set standard deviation plus $10^{-5}$. The linear stitch is ridge regression with an intercept and penalty $10^{-3}$ applied to all design coefficients. The nonlinear stitch has widths $64\to64\to8$ with a $\tanh$ hidden activation. Its targets are scaled by their fitting-set standard deviation plus $10^{-6}$. It is trained for 250 AdamW steps with batch size 256, learning rate $0.002$, and weight decay $0.001$. These maps are shared across evaluation streams. Each adjacent model pair has its own maps. Calibration replays are permitted on the separate calibration data, not on unavailable deployment histories.

\paragraph{Sensitivities and UCM.}
The diagonal structure makes each affected state coordinate depend on only one adapter coordinate. With $z_{t,0,j}=h^0_{t,j}$ and zero initial coefficients,
\begin{equation}
z_{t,\ell,j}=\lambda_jz_{t-1,\ell,j}+0.002z_{t-1,\ell-1,j},
\qquad \ell\geq1,\ j<8.
\end{equation}
The degree-$k$ query is $h^0_{t,j}+\sum_{\ell=1}^k a_{j\bmod2}^{\ell}z_{t,\ell,j}$. All other multivariate coefficients are identically zero. Exact first-order sensitivity is this formula with $k=1$, rather than a truncated-history approximation. UCM uses $k=2$ and $k=3$. None is re-centered after an update.

The UORO-style baseline approximates the $8\times2$ anchor sensitivity by an outer product $uv^\top$. It uses independent Rademacher vectors for each stream, propagates the first factor through the diagonal anchor Jacobian, and applies a norm-balanced rank-one update. A symmetric $10^{-30}$ stabilization is used in scale denominators. The random seed is $100s+j+666$ for model seed $s$ and stream $j$. The baseline uses UORO-style rank-one sensitivity updates, but does not reimplement the full recurrent-training procedure. Its migration query remains first-order in $a$, so it does not remove Taylor truncation error.

\paragraph{Quadratic interpolation bank.}
Before the stream, maintain the affected anchor and two auxiliary filters per affected coordinate, one at coordinate value $+0.75$ and one at $-0.75$. For a query scalar $a$ associated with that state coordinate, return
\begin{equation}
h^0+\frac{a}{2R_b}(h^+-h^-)
+\frac{a^2}{2R_b^2}(h^++h^--2h^0),\qquad R_b=0.75.
\end{equation}
This is ordinary three-node quadratic interpolation with exactly two extra filter states per affected coordinate, matching the coefficient budget of degree-two UCM. Because coordinates are independent, the two auxiliary global filters can assign both coordinates $+R_b$ or both $-R_b$. Those auxiliary vectors are outside the permitted replacement $\ell_1$ ball, but their scalar filters are stable and are defined before the stream. The method never uses a future replacement value to choose a filter. The comparison tests Taylor memory against an interpolation alternative at the same coefficient budget, beyond the first-order approximation baseline.

\subsection{Aggregation and predictive-loss results}
For a seed $s$, let $\mathcal E_s$ be its 64 stream/update pairs. With errors measured on the affected coordinates, define
\begin{equation}
\operatorname{NRMSE}_s=
\left(\frac{\sum_{e\in\mathcal E_s}\norm{\widehat h_e-h_e^{\mathrm{replay}}}^2}
{\sum_{e\in\mathcal E_s}\norm{h_e^{\mathrm{replay}}}^2}\right)^{1/2}.
\end{equation}
The manuscript reports the arithmetic mean and sample standard deviation of these five seed-level ratios. It does not average per-event relative errors, whose small denominators can distort the comparison. Predictive KL is averaged over each 32-position window, then over the 64 events, then across model seeds. All scoring uses float64 filtering and logits. Tiny negative KL values caused by rounding are clipped to zero. Differences in bits per byte are signed differences from the same seed and event's full-replay loss, not absolute discrepancies.

\begin{table}[t]
\caption{All-seed state NRMSE, in percent. Each entry aggregates the eight streams and eight update events belonging to one trained model.}
\centering\small
\input{tables/revision_language_seeds}
\label{tab:seed-results}
\end{table}

\begin{table}[t]
\caption{Ground-truth predictive loss relative to full replay. The middle column is in micro-bits per byte and reports mean $\pm$ sample standard deviation across five seeds. The right column also gives the seed-level minimum and maximum in micro-bits per byte. Negative values are lower empirical loss than replay, not necessarily better state fidelity.}
\centering\small
\input{tables/revision_language_loss}
\label{tab:loss-results}
\end{table}

\begin{figure}[t]
\centering
\includegraphics[width=\linewidth]{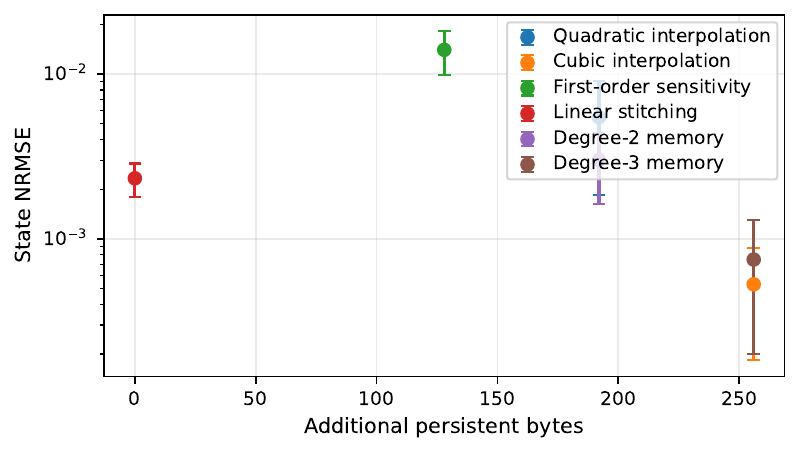}
\caption{State fidelity versus additional logical memory in the language diagnostic. Points and error bars are mean and sample SD across five model seeds. The equal-memory cubic bank is included. Shared fitting costs are not per-stream history bytes.}
\label{fig:state-prediction}
\end{figure}

Reset produces a predictive discrepancy, whereas several alternatives already yield small KL. Linear stitching gives lower state error than degree-two UCM, and the relative ordering of state error and predictive loss is not identical. The equal-memory cubic bank has smaller mean state error and KL than degree three, while both methods change true predictive loss negligibly. These results do not support claims of improved perplexity, adaptation quality, or general language-model performance.

\subsection{Timing, calibration, and resource accounting}\label{app:timing}
Timing uses the first trained model, its eight-update replacement schedule, and one 8192-byte held-out stream. The CPU is an AMD EPYC 9V74 80-Core Processor. The measured kernels use Numba and float64, with single-threaded execution throughout. The recorded software environment includes Python 3.13, NumPy 2.3.5, and PyTorch 2.10.0+cpu. A warmup compiles each kernel configuration. Twenty-one subsequent repetitions randomize the order of methods using a fixed trial-dependent permutation.

Every measured whole-stream call includes state initialization, embedding lookup, active-state updates, anchor and auxiliary-state maintenance when required, and all eight migrations. The state-only version does not evaluate logits. The second version evaluates the full nonlinear prediction head at every input for every method. All methods share the unaffected state coordinates, and methods that do not need the affected anchor do not update those anchor coordinates. Common-kernel scratch arrays are not a measured peak-memory claim. Logical retained-state counts in Table~\ref{tab:language} reflect the information needed by each method, not incidental allocations in the reference implementation.

\begin{table}[t]
\caption{Complete timing results. Entries are median [25th percentile, 75th percentile] across 21 repetitions. Times are for one complete stream, not one migration.}
\centering\small
\input{tables/revision_language_timing}
\label{tab:timing-full}
\end{table}

The certificate timing variant adds the degree-three radius-uniform residual update with enclosing radius $0.750001$, using operator-norm bounds computed from the stored decay and direction arrays and the highest-degree coefficient norm over each parameter group. It is an optional cost beyond the coefficient-only variant. Timing excludes training, compilation, calibration, and loading the data and models. Training and calibration costs are reported separately, without assuming a deployment count over which to amortize them. The calibration measurements include exact state-pair construction and fitting both the linear and nonlinear maps. They were collected while other experiment processes could be active, so they should not be used as isolated hardware performance claims.

For logical memory, float64 states use eight bytes per coordinate and retained raw inputs use one byte per token. A degree-$k$ method stores an additional eight-dimensional anchor and $8k$ coefficients beyond the active state, giving $64(k+1)$ bytes. Degree three therefore needs 256 extra bytes, or 264 with the optional scalar certificate. The quadratic bank needs 192 bytes and the cubic bank needs 256. Checkpoint replay needs $64+L$ bytes. The rank-one sensitivity baseline needs eight anchor values, eight values for one factor, and two for the other, or 144 bytes. Full replay needs $t$ retained input bytes at prefix $t$. Stitching has no additional per-stream historical state, but requires shared maps and calibration data. A linear map has $65\times8=520$ regression coefficients plus feature normalization parameters. The nonlinear map has 4680 network parameters plus feature and target normalization parameters. These are incurred per adjacent model pair, not per stream. The timing comparison includes both fitted stitchers and the cubic bank, with calibration recorded separately. UORO is included in the accuracy table but is not timed in this deterministic serving comparison. Model weights and the current global adapter coordinate are common deployment costs.

The work measurements do not support a general speed advantage. For this small model, checkpoint replay is a strong inexpensive alternative, and the prediction head largely masks state-maintenance differences. The timing table reports this trade-off across the whole stream, rather than only the polynomial query's latency.

\subsection{Evidence boundaries}
The controlled numerical streams test the mathematical mechanism, including dense noncommuting transitions and failure under leakage. They are not a substitute for forecasting, control, or anomaly-detection benchmarks. The language study tests five small learned models with a particularly favorable invariant structure and a modest update radius. It does not test pretrained models, unknown update subspaces, multi-layer input changes, or arbitrary continual fine-tuning. The corpus is small, and no claim about broad linguistic generalization is made. The direct prediction-head skip path is another reason not to infer task improvements from the large relative state-error reductions.

The experiments provide evidence about the implemented retained-memory mechanism within the stated assumptions. The theoretical lower bounds explain why those assumptions matter. Whether a practical application benefits from the mechanism depends on the closure dimension, the tolerated predictive discrepancy, the update frequency, the available replay buffer, and the maintenance cost paid before each update.

%% file: tables/revision_language_seeds.tex
\begin{tabular}{lrrrrr}
\toprule
Method & Seed 0 & Seed 1 & Seed 2 & Seed 3 & Seed 4 \\
\midrule
Quadratic interpolation & 0.97179 & 0.46294 & 0.85123 & 0.37411 & 0.08055 \\
Cubic interpolation & 0.09378 & 0.04386 & 0.08466 & 0.01676 & 0.02591 \\
First-order sensitivity & 1.73214 & 1.07336 & 1.91705 & 0.96552 & 1.32557 \\
Full replay & 0.00000 & 0.00000 & 0.00000 & 0.00000 & 0.00000 \\
Linear stitching & 0.30478 & 0.19829 & 0.26668 & 0.22509 & 0.17058 \\
Nonlinear stitching & 0.56802 & 0.32188 & 0.36767 & 0.33579 & 0.26696 \\
Affected-state reset & 100.00000 & 100.00000 & 100.00000 & 100.00000 & 100.00000 \\
Stale state & 4.15607 & 2.84449 & 3.09316 & 2.67017 & 1.97735 \\
Degree-2 memory & 0.33924 & 0.17474 & 0.42413 & 0.13603 & 0.43159 \\
Degree-3 memory & 0.07009 & 0.03118 & 0.09701 & 0.01989 & 0.15642 \\
UORO-style rank one & 24.89516 & 30.22215 & 22.82005 & 22.03831 & 33.06318 \\
Checkpoint replay, 256 & 1.57449 & 0.52350 & 0.65105 & 0.37370 & 0.25552 \\
Checkpoint replay, 64 & 3.17064 & 1.71402 & 1.93558 & 1.50801 & 1.01748 \\
\bottomrule
\end{tabular}

%% file: tables/revision_language_loss.tex
\begin{tabular}{lrr}
\toprule
Method & $\Delta$ bits/byte ($10^{-6}$) & Minimum / maximum seed \\
\midrule
Full replay & $0.00\pm0.00$ & $0.00\,/\,0.00$ \\
Affected-state reset & $10051.64\pm3279.34$ & $5582.65\,/\,12803.97$ \\
Stale state & $-47.57\pm36.51$ & $-96.32\,/\,-7.99$ \\
Checkpoint replay, 64 & $-28.72\pm28.60$ & $-71.67\,/\,-3.01$ \\
Checkpoint replay, 256 & $-10.48\pm14.84$ & $-35.89\,/\,-0.73$ \\
Linear stitching & $0.05\pm3.12$ & $-5.25\,/\,2.59$ \\
Nonlinear stitching & $-0.33\pm3.93$ & $-6.93\,/\,3.05$ \\
UORO-style rank one & $205.22\pm336.66$ & $-240.01\,/\,468.20$ \\
First-order sensitivity & $-0.63\pm21.89$ & $-34.97\,/\,24.47$ \\
Quadratic interpolation & $1.37\pm6.10$ & $-4.54\,/\,10.69$ \\
Degree-2 memory & $0.80\pm7.53$ & $-5.09\,/\,13.61$ \\
Degree-3 memory & $-0.94\pm2.28$ & $-4.90\,/\,0.59$ \\
Cubic interpolation & $0.48\pm1.13$ & $-0.65\,/\,2.06$ \\
\bottomrule
\end{tabular}

%% file: tables/revision_language_timing.tex
\begin{tabular}{lrr}
\toprule
Method & State-only ms [IQR] & With-head ms [IQR] \\
\midrule
Stale state & $0.956\;[0.938,0.986]$ & $57.921\;[57.436,60.541]$ \\
Full replay & $1.305\;[1.277,1.352]$ & $59.716\;[57.798,60.472]$ \\
Checkpoint replay, 64 & $0.928\;[0.916,0.950]$ & $58.912\;[58.057,60.988]$ \\
Checkpoint replay, 256 & $0.970\;[0.949,1.008]$ & $59.290\;[58.121,60.569]$ \\
Linear stitching & $0.948\;[0.933,0.965]$ & $58.741\;[58.461,59.124]$ \\
Nonlinear stitching & $0.965\;[0.963,1.004]$ & $59.279\;[58.002,60.087]$ \\
First-order sensitivity & $1.030\;[1.014,1.044]$ & $58.994\;[58.534,60.481]$ \\
Degree-2 memory & $1.081\;[1.060,1.159]$ & $58.921\;[57.827,61.254]$ \\
Degree-3 memory & $1.123\;[1.098,1.209]$ & $58.853\;[58.452,59.937]$ \\
Degree 3 + certificate & $1.425\;[1.345,1.502]$ & $59.870\;[58.619,60.401]$ \\
Quadratic interpolation & $1.063\;[1.045,1.076]$ & $59.077\;[57.900,60.713]$ \\
Cubic interpolation & $1.163\;[1.152,1.201]$ & $58.937\;[57.908,59.777]$ \\
\bottomrule
\end{tabular}

%% file: sections/revision_experiments.tex
\section{Additional protocols and evidence}\label{app:revision}
We give the finite-accuracy calculations, equal-memory comparisons, coupled identification task, and cost accounting used in the main text. The language comparisons use the same five trained models described in Appendix~\ref{app:language}, changing only the migration method. The models and numerical results are included in the supplement.

\subsection{Finite-query spectra and precision budgets}\label{app:revision-spectra}
The rank-one spectral diagnostic takes $d\in\{1,4,8\}$,
\begin{equation}
A=\diag\bigl(\operatorname{linspace}(0.3,0.9,d)\bigr),\qquad
v=\frac{(1,2,\ldots,d)^\top}{\norm{(1,2,\ldots,d)}},\quad
D=0.15vv^\top,\quad b=\frac{\mathbf1}{\sqrt d}.
\end{equation}
For $d=1$, the single diagonal entry is $0.3$. The horizon is $t\in\{3d,8d,64d\}$ and the query radius is $R\in\{0.01,0.05,0.2,0.5\}$. For each configuration, use 33 nodes
\begin{equation}
a_j=R\cos\!\left(\frac{(2j+1)\pi}{66}\right),\qquad j=0,\ldots,32.
\end{equation}
All queried recurrences are contractive, since $\norm A+0.15R\leq0.975$. The horizon map is formed explicitly as $[A(a)^{t-1}b,\ldots,b]$. An orthonormal basis $Z$ for $\ker R_0$ is obtained by singular-value decomposition, and the reported spectrum is that of $R_QZ$. It has the same nonzero singular values as $R_QP_0$ and avoids forming a large projector.

\begin{figure}[t]
\centering\includegraphics[width=\linewidth]{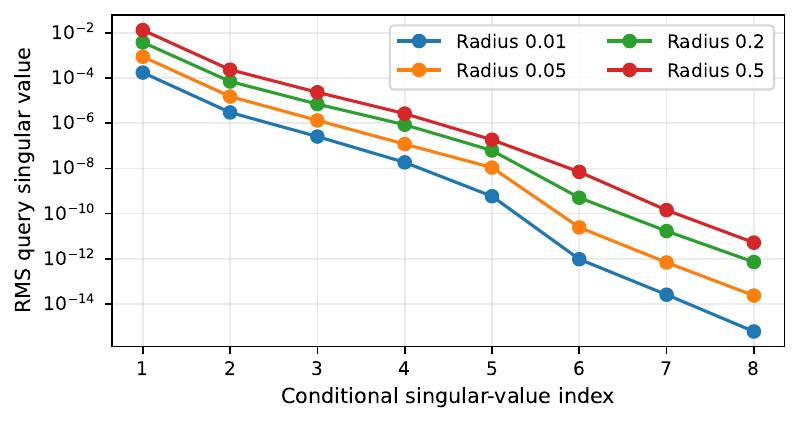}
\caption{RMS-normalized conditional singular values for the four-state rank-one family at horizon 12. The query radius changes finite-tolerance requirements even when exact rank is unchanged. The 33 nodes are stated above.}
\label{fig:spectrum}
\end{figure}

The $d=4,t=12,R=0.2$ example has necessary continuous dimensions $\SpectralDimensions$ at RMS tolerances $10^{-2},10^{-3},10^{-4},10^{-6},10^{-8}$. The auxiliary-bit lower bounds are $\SpectralBits$ before rounding to an integer. Table~\ref{tab:round2-spectrum} gives the tolerance budgets, and Figure~\ref{fig:spectrum} shows how the conditional spectrum varies with query radius. These are fixed-horizon calculations on the unit-energy history ball, not claims of a cheap causal encoder.

\begin{table}[t]
\caption{Conditional information for the stated four-state, horizon-12, radius-0.2 example. The bit column is a real-valued lower bound on an integer message length, with an exact old state supplied separately.}
\label{tab:round2-spectrum}
\centering\input{tables/round2_spectrum}
\end{table}

The leading five conditional singular values at radius $0.2$ are $3.74416862\times10^{-3}$, $6.95036610\times10^{-5}$, $6.93525064\times10^{-6}$, $8.47191936\times10^{-7}$, and $6.17494116\times10^{-8}$. One implementation uses explicit powers and an orthonormal null-space basis. A second builds the maps recursively and removes the anchor row space. Their maximum absolute discrepancy is $1.1\times10^{-16}$. The full numerical results are included in the supplement.

For the language recurrence, we also evaluate the continuous relaxation in which each of the eight affected coordinates has an independently varying injection. For each base decay $\lambda_j$, take the scalar injection gain $1-\lambda_j$, update slope $0.002$, 17 Chebyshev nodes, radii $0.1,0.4,0.75$, and horizons $256,1024,4096$. The direct-sum conditional spectrum is obtained by combining the eight scalar spectra. This history ball contains injection sequences that actual byte embeddings cannot generate. The spectrum therefore describes the recurrent operators, not an information lower bound for the discrete language data.

The memory-budget comparison uses the sufficient bounds of Appendix~\ref{app:decision}. It varies $\rho\in\{0.9,0.97,0.99,0.995,0.999\}$, $q\in\{0.05,0.2,0.4,0.7,0.9,0.99\}$, $p\in\{1,2,4\}$, $r\in\{2,4,8\}$, normalized tolerance $\tau\in\{0.1,0.01,0.001\}$, and relative leakage $\delta/(1-\rho)\in\{0,10^{-4},10^{-3},10^{-2}\}$. An entry marked infeasible means that the displayed sufficient bound cannot certify the tolerance, not that migration itself is impossible.

\subsection{Equal-memory cubic interpolation in the language model}\label{app:revision-language}
The cubic bank uses four fixed nodes
\begin{equation}
\xi_j=0.750001\cos\!\left(\frac{(2j+1)\pi}{8}\right),\qquad j=0,1,2,3.
\end{equation}
For each affected coordinate, it retains the four scalar filter states at those nodes and evaluates the ordinary Lagrange interpolant when the replacement is revealed. The nodes are neither selected from a test query nor calibrated on evaluation histories. A separable coordinate uses only its own parameter, so each scalar node is also a valid query on that coordinate axis. Bundling independent coordinate filters into arrays does not require treating a vector with both coordinates equal to a node as an allowed model replacement.

The bank stores four float64 values per affected coordinate, or 256 extra bytes in total, with no affected anchor needed by the bank. Degree-three coefficient memory stores an affected anchor and three coefficients per coordinate, also 256 bytes. Both methods keep an active state for ordinary execution between replacements. The unaffected 56 coordinates are shared. Neither method is given the replacement coordinate during historical maintenance. The bank and coefficient methods may receive multiple subsequent queries without changing their underlying retained memory.

\begin{table}[t]
\caption{Complete language comparison, including all replay, stitching, sensitivity, coefficient, and interpolation baselines. Entries are mean $\pm$ sample SD across the five trained models. State errors are measured only on affected coordinates. KL is in nats.}
\centering\small
\input{tables/revision_language_all}
\label{tab:language-complete-revised}
\end{table}

The cubic bank improves mean state NRMSE and mean KL relative to degree-three coefficients. Its state error is lower in three seeds and higher in two (Figure~\ref{fig:language-paired}). The ordering is specific to this experiment. Interpolation accuracy depends on node placement, parameter dimension, and query location. A comparison against first-order sensitivity alone cannot identify the best use of the extra memory. The signed empirical loss differences for both methods remain on the micro-bit scale.

Linear and nonlinear stitching maps use the independent calibration split, seeds, optimizer settings, and sample counts described above. Map application is timed separately from fitting. The timing comparison includes both stitchers and cubic interpolation. UORO remains in the accuracy comparison but has no measurement in the deterministic kernel timing table; its cost is not zero.

\subsection{Learning the coupled model and its update family}\label{app:revision-task}
The coupled study is a fully observed synthetic system-identification problem. Its observation model provides states during calibration, and its data-generating system deliberately contains an invariant slow block. These assumptions permit learning a useful restricted update family. They do not establish that comparable structure can be learned in an arbitrary pretrained neural model. The matrices are dense in the reported coordinates, and the perturbations need not commute. The scalar-input setting also provides an appropriate test of history features that can be shared across coefficient directions.

For seed $s\in\{0,1,2,3,4\}$, initialize NumPy's generator with $18001+s$. Obtain an orthogonal $128\times128$ matrix $Q_*$ from a Gaussian QR factorization, and let $U_*$ be its first eight columns. For a selected $\rho\in\{0.97,0.995,0.999\}$, set
\begin{equation}
A_*=Q_*\diag\bigl(\rho I_8,\operatorname{linspace}(0.5,0.9,120)\bigr)Q_*^\top.
\end{equation}
Draw independent $8\times8$ Gaussian matrices, normalize their operator norms to $0.6(1-\rho)$, and lift them through $U_*$ to obtain two true update matrices $D^*_1,D^*_2$. Their images are in the slow block. The scalar injection vector has independent Gaussian slow coordinates and Gaussian fast coordinates scaled by $0.2$, is normalized by the norm of its slow coordinates, and is then rotated by $Q_*$ and multiplied by $\sqrt{1-\rho^2}$. The scalar readout is a fixed known random unit vector in $\im U_*$ and is not estimated from outputs alone. There is no direct input-to-output path.

The learner first estimates the entire source transition $A$ from 4096 independent, randomly initialized one-step observations. The initial states and scalar inputs are standard Gaussian, and the next-state observations have independent Gaussian noise with standard deviation $0.6(1-\rho)\times10^{-3}$. The known injection $bx$ is subtracted before least squares. These are observed-state calibration data, not latent-state sequence supervision.

Six pilot domains use coefficients $(0.6,0),(-0.6,0),(0,0.6),(0,-0.6),(0.3,0.3),(-0.3,-0.3)$ in the true update basis. Each supplies 2048 independent observations and an unrestricted transition estimate. Subtract the fitted source transition, flatten the six differences, and retain the least number of right-singular directions explaining $99.9\%$ of squared singular-value mass. This selects $p=2$ for every reported fit. Concatenate the images of those directions and apply the same energy rule to obtain $r=8$. No evaluation history participates in either SVD. The retained image basis is the learned $U$, and its two reduced directions are normalized to operator norm $0.6(1-\rho)$.

Four further domains use the true coefficient pairs $(0.55,0.25)$, $(-0.45,0.35)$, $(0.25,-0.55)$, and $(-0.45,-0.35)$. Each supplies 2048 new observed transitions. The restricted model fits two coordinates in $A+U\sum_i a_iD_i$ by least squares. The declared rule scales a coordinate vector onto the unit $\ell_1$ ball if its norm exceeds one. No projection is triggered in the recorded fits. The unrestricted comparator estimates a full transition from the same observations. Both are evaluated against the true domain's conditional mean. This comparison isolates task loss from errors due to using a restricted adaptation family.

\begin{table}[t]
\caption{Measured structure of the fitted coupled families. The last column is the mixed coefficient norm at the end of the final held-out stream. Matrices are time invariant, so their leakage norm is the same on every stream. Nonzero mixed coefficients and noncommutators do not by themselves establish independent historical information.}
\centering\small
\input{tables/revision_task_structure}
\label{tab:task-structure}
\end{table}

The fitted bases are not exactly invariant. At the main setting $\rho=0.995$, measured leakage ranges from $1.80\times10^{-4}$ to $5.01\times10^{-4}$, or $3.59\%$ to $10.01\%$ of the fitted source stability margin. Relative leakage rises as the margin narrows. All fitted families satisfy the recorded operator-norm contraction bound for the full declared radius. Gaussian inputs are not uniformly bounded over all realizations, so the global bounded-injection theorem is not asserted uniformly over this data distribution. A finite run has a finite realized injection bound, and the residual theorem does not require a fixed global injection bound. The experiments report measured errors separately from the conservative leakage bound.

\subsection{Independent calibration and streaming evaluation}
The separate decoder-calibration history contains 8192 independent standard Gaussian inputs, cast to float32, from seed $26001+s$. It is processed under the fitted source and four target matrices. State-pair samples use positions $1024,1028,\ldots,8188$. The first $75\%$ of these samples fit each regression and the last $25\%$ select among ridge penalties $10^{-8},10^{-6},10^{-4},10^{-2}$. Feature means and standard deviations are estimated only on the fitting portion. Five stitching maps cover source-to-first-target and the four cyclic target transitions. The fitted history baseline uses four separate maps from its retained history coordinates to the target state. All calibration is independent of evaluation histories.

For held-out stream $j$ in seed $s$, the generator seed is $37001+100s+j$. Each stream has 4096 independent standard Gaussian float32 inputs and starts from zero. The main update period is 512 inputs. The seven replacement times are $512,1024,\ldots,3584$, and the four domain coordinates repeat cyclically. The memory is initialized before the stream and knows the fitted family and its radius, but no query coordinate is supplied to its update. At each event, the reference processes the whole prefix under that single target matrix. The prediction window comprises the current output and 31 subsequent common inputs. Scoring copies do not overwrite a method's active state or advance its next event.

The contract is retrospective re-estimation under a newly selected domain model. The protocol does not change physical dynamics retrospectively or assume that a filter should always discard the true historical parameter sequence. Each baseline advances its own active state between events. Checkpoint replay retains its actual state $L$ inputs before the announced event, even if that suffix crosses an earlier replacement. It then replays the suffix with the new target transition. Replacement times are announced for checkpoint capture, but replacement coordinates are not.

For every seed and method, sum squared prediction errors across all events, streams, and 32-position windows, and divide by the corresponding sum of squared clean target outputs. This defines the reported normalized MSE. The five seed ratios then determine the mean and sample SD. The target is a conditional mean, not a noisy test observation. A separate fidelity column compares with the fitted target's exact replay rather than the true generating system. Both are retained in the raw event table.

\begin{table}[t]
\caption{All methods on the coupled task at $\rho=0.995$ and update period 512. Prediction NMSE is relative to the true conditional mean. Stream times are measured only for the indicated serving kernels. The analytic history bank and full replay have nearly identical task error, which is then dominated by fitting the replacement model rather than migration.}
\centering\small
\input{tables/revision_task_all}
\label{tab:task-all}
\end{table}

\begin{table}[t]
\caption{Every seed's coupled-task NMSE. The individual values show the between-seed variation hidden by a pooled result. The paired adaptation-benefit calculation uses each seed's own no-adaptation and full-replay errors.}
\centering\small
\input{tables/revision_task_seeds}
\label{tab:task-seeds}
\end{table}

Let $L_{s,m}$ denote a seed's NMSE and define the recovered adaptation benefit as
\begin{equation}
1-\frac{L_{s,m}-L_{s,\mathrm{replay}}}{L_{s,\mathrm{anchor}}-L_{s,\mathrm{replay}}}.
\end{equation}
Every denominator is positive in the recorded main setting. Degree three has mean recovered benefit $\TaskRecoveredBenefitMean\%$ and minimum $\TaskRecoveredBenefitMinimum\%$. This is a paired descriptive result on five constructed systems, not a confidence bound for a population of deployed models. The analytic history bank also attains essentially the full replay benefit.

\subsection{Equal-memory multivariate interpolation}
The nonconstant monomials of total degree at most three in two variables give $N_{2,3}=9$. Form candidate points $(i/8,j/8)$ for integers $i,j\in[-8,8]$ satisfying $0<|i|+|j|\leq8$. Evaluate the nine nonconstant monomials at those points and use column-pivoted QR on the transposed evaluation matrix to select nine nodes. The anchor value at the origin is already known. The selected square evaluation matrix $V$ has condition number $26.68$ in the recorded construction. The selection uses neither target coordinates nor test inputs.

At each selected node $\xi$, maintain the full projected reduced response
\begin{equation}
r_t(\xi)=\left(T+\sum_i\xi_iG_i\right)r_{t-1}(\xi)
+\sum_i\xi_iD_i h_{t-1}^0.
\end{equation}
For a query, let $m(a)$ be its vector of nine monomials and return $h_t^0+U m(a)^\top V^{-1}r_t$. The bank uses the same projected dynamics and possible leakage as coefficient memory, not a larger full-state filter bank. It retains the full 128-state anchor and nine eight-state responses. Its 1600-byte auxiliary budget matches degree-three coefficient memory exactly. The original separable language bank can use scalar filters without this projection.

\subsection{A strong history-basis baseline and its analytic decoder}\label{app:history-bank}
A model-independent history basis is a direct competitor in the scalar-input task. Following the classical Laguerre-model construction \citep{wahlberg1991}, we use a discrete Laguerre bank with pole $\alpha=\rho$ and width $M=200$, fixed before the stream. Define formal delay-domain basis functions
\begin{equation}
\phi_j(z)=\frac{\sqrt{1-\alpha^2}}{1-\alpha z}
\left(\frac{z-\alpha}{1-\alpha z}\right)^j,\qquad j\geq0.
\label{eq:laguerre-transfer}
\end{equation}
They can be maintained without the past inputs. With all coordinates initially zero,
\begin{align}
y_{t,0}&=\alpha y_{t-1,0}+\sqrt{1-\alpha^2}x_t,\\
y_{t,j}&=\alpha y_{t-1,j}+y_{t-1,j-1}-\alpha y_{t,j-1},\qquad j\geq1.
\label{eq:laguerre-update}
\end{align}
The second equation uses the newly computed preceding coordinate. A shared-history baseline of this form follows the general approach of retaining a basis of the past rather than a basis of one model's hidden representation \citep{gu2020}. It is a discrete Laguerre construction, not an implementation claim about the particular HiPPO-LegS algorithm.

The fitted variant learns a state decoder from independent calibration histories. A stronger variant needs no historical calibration data. For a new stable target matrix $A$ and known input vector $b$, its analytic state-decoder columns are
\begin{equation}
l_0=\sqrt{1-\alpha^2}(I-\alpha A)^{-1}b,
\qquad
l_j=(I-\alpha A)^{-1}(A-\alpha I)l_{j-1}.
\label{eq:analytic-laguerre}
\end{equation}
At a query, return $\widehat h_t=\sum_{j=0}^{M-1}l_j y_{t,j}$. Only the decoder depends on the new model. The history coordinates were already maintained before it was known. Between queries, the active state follows the exact target recurrence, as for all migration methods.

The matrix $L_M=[l_0,\ldots,l_{M-1}]$ reconstructs the full state. It is constructed from $A$, $b$, and $\alpha$, without the readout $C$, prediction labels, or an output-fitting objective. Changing $C$ therefore requires no decoder refit, although it can change the width needed for accurate outputs. The construction is readout-agnostic, not replacement-model-independent.

To justify the decoder, write $\phi_j(z)=\sum_{s\geq0}\phi_j[s]z^s$. For $|\alpha|<1$, these impulse responses form an orthonormal basis of one-sided square-summable sequences. On the unit circle the ratio $(z-\alpha)/(1-\alpha z)$ is all-pass, and its change of angular variable has Jacobian $(1-\alpha^2)/|1-\alpha z|^2$. Substitution in the Fourier inner product gives orthogonality and unit norm. Completeness follows by the corresponding invertible weighted composition operator applied to the ordinary nonnegative Fourier basis. The target impulse response is $A^sb$, so its projection coefficient is
\begin{equation}
\sum_{s\geq0}\phi_j[s]A^sb=\phi_j(A)b=l_j.
\end{equation}
The matrix functions are rational functions of the same $A$ and therefore commute. This yields~\eqref{eq:analytic-laguerre}. Stability ensures convergence. Truncating the basis approximates the target impulse response, not the parameter Taylor expansion.

The same construction gives a finite-tolerance causal certificate for the chosen bank. If $W_c=AW_cA^\top+bb^\top$ and $L_M=[l_0,\ldots,l_{M-1}]$, then the residual impulse-response Gram matrix in exact arithmetic is $E_M=W_c-L_ML_M^\top\succeq0$. For a history of Euclidean input norm at most one, the worst state error of this bank is $\sqrt{\lambda_{\max}(E_M)}$. For root-sum-squared future output error with fixed target $A,C$, replace this by $\sqrt{\lambda_{\max}(E_M^{1/2}W_oE_M^{1/2})}$, where $W_o=A^\top W_oA+C^\top C$. The claim follows by taking the operator norm of the omitted-history map, followed by the observability operator. At a fixed finite prefix it remains an upper bound. This is a certificate for a chosen scalar-input basis and target, not a minimum over all causal encoders or a uniform guarantee over an unbounded input distribution. Its Gramian computation and numerical validation are not included in the timing comparison, and no machine-level certificate is claimed for the nearly zero residuals of the main task.

The 200 retained float64 history coordinates use 1600 bytes, equal to the generic leaky degree-three memory. Their analytic decoder is shared model data, not per-stream memory. Its storage exceeds that of the polynomial query coefficients, and constructing it incurs a separate cost. Both are reported below. The basis's success is consistent with the shared-moment calculation in Appendix~\ref{app:shared-moments}. A dense coordinate rotation and noncommuting perturbations do not eliminate historical redundancy in the underlying scalar-input slow-block model.

\subsection{Contraction, frequency, and leakage regimes}\label{app:revision-sweeps}
We fit the full pipeline for each of five seeds at $\rho=0.97,0.995,0.999$ with period 512. The same seed conventions and true update coordinates are used, while update norms scale with the stability margin. This changes contraction and history length, not just the absolute size of a fixed transition perturbation. At $\rho=0.995$ we additionally use periods 128 and 1024. Reusing the period-512 run gives 25 distinct fitted/evaluated configurations, without counting that run twice. Periods 128,512,1024 give 31,7,3 migration events per stream. There are 33,000 event-method measurements in the complete coupled study.

\begin{figure}[t]
\centering\includegraphics[width=\linewidth]{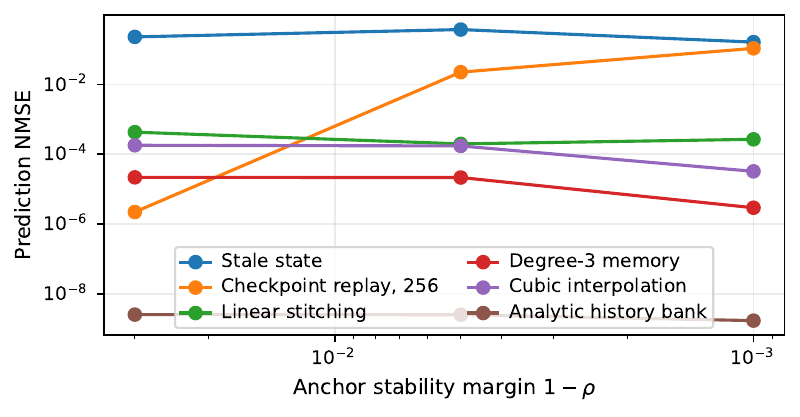}
\caption{Contraction-design sweep. Means across five fitted-model seeds. The update scale and relative learned leakage also vary with the nominal contraction target, so this is not an isolated causal effect of contraction.}
\label{fig:regime-sweeps}
\end{figure}

\begin{figure}[t]
\centering\includegraphics[width=\linewidth]{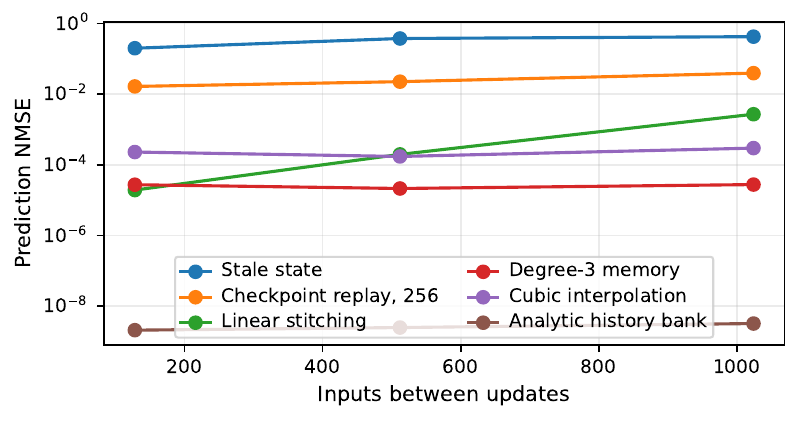}
\caption{Replacement-frequency sweep. The periods change historical exposure and the mixture of target domains as well as the number of queries. All methods use the same schedule within a setting.}
\label{fig:frequency-appendix}
\end{figure}

The tables below give the full variability across fitted models. Figures~\ref{fig:regime-sweeps} and~\ref{fig:frequency-appendix} show the corresponding means.

\begin{table}[t]
\caption{Contraction sweep. Mean $\pm$ sample SD across five fitted models per setting. The relative leakage of the learned subspace also changes with the stability margin, as shown in Table~\ref{tab:task-structure}.}
\centering\small
\input{tables/revision_contraction}
\label{tab:contraction-sweep}
\end{table}
\begin{table}[t]
\caption{Update-frequency sweep at $\rho=0.995$. Windows contain 32 predictions after each replacement. Query schedules are matched across methods within a setting.}
\centering\small
\input{tables/revision_frequency}
\label{tab:frequency-sweep}
\end{table}

The original noncommuting geometry study varies leakage independently of the other model parameters and sweeps six query radii. It therefore complements rather than duplicates the learned study. Neither study supports extrapolating a monotone error trend to arbitrary neural models. In particular, different frequencies expose different histories and different mixtures of the four replacement domains.


\subsection{Certificate tightness and its use as a decision rule}\label{app:revision-certificates}
At horizon 4096 and query radius $0.4$, Table~\ref{tab:certificate-ratios} reports the ratio of the radius-$0.8$ certificate to the actual error, pooled over the finite set of five seeds and twelve streams at each setting. These ratios describe conservatism, not probabilistic coverage. The certificate is uniform over a larger radius than the evaluated query, so it need not be close to one.
\begin{table}[t]
\caption{Conservatism of the online certificate at prefix 4096 and query radius $0.4$. Degree and nominal leakage identify the setting. The certificate was maintained for the full radius $0.8$.}
\centering\small
\input{tables/revision_certificate}
\label{tab:certificate-ratios}
\end{table}

To assess whether the certificate can select a degree, each run maintains certificates for degrees one through four before querying. At prefix 4096, choose the smallest already-stored degree whose radius-uniform certificate is below the desired absolute state tolerance. If none qualifies, record failure to certify rather than using unavailable replay or claiming a successful migration. The three tolerances are $0.001$, $0.01$, and $0.1$.
\begin{table}[t]
\caption{Certificate-based degree selection among degrees already maintained before the query. Each fraction is over 60 runs. A missing mean degree means no run was certified. The memory and historical maintenance cost remain those of degree four.}
\centering\small
\input{tables/revision_certificate_policy}
\label{tab:certificate-policy}
\end{table}

At zero leakage, every run certifies tolerance $0.01$ with degree two. At nominal leakage $0.003$, only $55\%$ can certify that tolerance with the available array. At leakage $0.01$ or $0.03$, none can. The failures show where the exact-arithmetic sufficient condition is too conservative to certify the requested tolerance. The ordinary floating-point experiments check inequalities with tolerance $10^{-10}$.

\subsection{Output-oriented retention diagnostic}\label{app:revision-output}
The common-output quotient experiment uses five seeds, $d=12$, an eight-dimensional update image, two noncommuting directions, and a four-dimensional linear readout. Four source coordinates are visible, only two of which lie in the update image. Six further update-image coordinates are invisible and are dynamically separated from the visible block. A Gaussian orthogonal rotation makes the reported matrices dense. Iterated row-space closure recovers a four-dimensional quotient with $r_{\rm out}=2$.

For degrees one through three, four independent input streams of length 1024 compare full and quotient coefficient propagation under the same query $(0.35,-0.4)$. The reduced version retains one quarter as many coefficient scalars. Its outputs agree with the full coefficient method within $5.66\times10^{-15}$ in the recorded runs, although lifting the quotient state by setting invisible coordinates to zero does not reconstruct the full hidden state. This numerical agreement illustrates the distinction between preserving outputs and preserving every coordinate under the common-invariance assumptions. It is not a compression result for nonlinear readouts.

\subsection{Complete serving, setup, and shared-memory costs}\label{app:revision-costs}
The coupled serving benchmark uses seed zero at $\rho=0.995$, a single 4096-input stream, and seven replacements. Every timed call includes all dense active-state transitions, all required anchor and auxiliary updates, migration work, initialization, and a scalar readout at each input. All methods receive the same float32 inputs and perform state arithmetic in float64. Each kernel is compiled and warmed before 21 trials, whose method order is randomly permuted. BLAS and OpenMP are restricted to one thread. The benchmark records timings, checksums, and the machine/software metadata. The benchmark times the serving computation, excluding the evaluation routine's stored oracle trajectories.
\begin{table}[t]
\caption{Complete coupled serving times. Median and interquartile range over 21 repetitions on one fitted model and one stream. These are timing repetitions, not independent model seeds.}
\centering\small
\input{tables/revision_task_timing}
\label{tab:task-serving}
\end{table}

Table~\ref{tab:shared-setup} measures model-side construction separately on the same fitted model, without amortizing it over an unspecified number of client streams. The analytic construction is summed over all four target decoders. The fitted history and stitching entries include processing the independent calibration history, generating exact calibration target states, fitting, and selecting the ridge penalty. The model-reduction entries are once per source family. Because the coefficient setup uses the same generic node-preparation routine as interpolation in the reference code, its reported time conservatively includes that routine rather than subtracting a hypothetical optimized cost. All setup measurements use 21 repetitions and exclude compilation and file I/O. The original base and domain model fitting is common to migration strategies and is recorded separately in each setting's metadata.
\begin{table}[t]
\caption{Shared setup measured independently of per-stream serving. These costs must be added for a single fresh deployment or amortized over an explicitly stated number of streams.}
\centering\small
\input{tables/revision_setup}
\label{tab:shared-setup}
\end{table}

Adding the median analytic setup for all four targets to its median serving time gives $13.61$ ms, versus $25.24$ ms for the coefficient setup plus degree-three serving on this instance. This sum of separately measured medians is a descriptive cost decomposition, not a new jointly measured latency quantile or a general hardware performance claim. Linear stitching serves quickly once fitted but has a larger measured calibration cost.

Shared model data are separate from stream-specific retained history. Table~\ref{tab:shared-storage} counts actual saved float64 arrays. A single analytic decoder contains $128\times200$ coefficients, or 204,800 bytes, and four cached decoders use 819,200 bytes. A deployment can construct and cache fewer model decoders at a time, trading storage against recomputation. The raw calibration history and temporary state-pair arrays can be discarded after fitting. Their work is included in setup, but they are not claimed to be part of a 1600-byte streaming summary.
\begin{table}[t]
\caption{Shared auxiliary model storage for the coupled benchmark. These are not per-stream history bytes. Base and target weights are common model data and are excluded from every method's auxiliary column.}
\centering\small
\input{tables/revision_shared_memory}
\label{tab:shared-storage}
\end{table}

The main per-stream counts exclude a common 128-coordinate active state, or 1024 bytes. Learned leakage prevents exact sharing of its complement with the anchor, so the coefficient and projected-node methods retain a full extra 128-state anchor. Degrees one, two, and three then require 1152,1344,1600 extra bytes. The generic cubic bank and both width-200 history banks also require 1600. A checkpoint stores a full 1024-byte state plus $4L$ raw input bytes. Full replay stores $4t$ raw bytes at prefix $t$. Stitching has no additional history state but uses the shared map data. Logical counts exclude temporary work arrays and are not measurements of peak Python-process memory.

\par
\paragraph{Amortization over streams and queries.}
For method $j$, let $S_j$ be shared setup work, $H_j$ shared auxiliary bytes, and $c_j,q_j,m_j$ per-input work, per-query work, and per-stream auxiliary bytes. For $N$ streams with $T$ inputs and $Q$ queries each, the corresponding totals are
\[
\mathcal W_j=S_j+N(Tc_j+Qq_j),\qquad
\mathcal M_j=H_j+Nm_j.
\]
In the work expression, $N$ counts processed streams sharing the prepared targets. In the memory expression it counts histories retained simultaneously. All methods must first meet the same accuracy requirement. When replay cost grows with prefix length, $Qq_j$ denotes the sum of its query-specific costs. Each newly encountered target also adds its decoder-construction or calibration cost to $S_j$. These formulas separate query frequency from the per-input maintenance charged even when no replacement occurs.

For the measured four-target, 4096-input, seven-query schedule, the retained medians give the following estimates, in milliseconds:
\[
\begin{aligned}
\mathcal W_{\rm bank}&=4.833+8.776N,&
\mathcal W_{\rm replay}&=30.495N,\\
\mathcal W_{\rm checkpoint}&=9.777N,&
\mathcal W_{\rm stitch}&=136.456+7.065N.
\end{aligned}
\]
The bank is cheaper than replay from one stream, cheaper than checkpoint-256 from five streams, and cheaper than stitching through 76 streams. Beyond that, stitching amortizes its calibration cost. These estimates add separately measured medians, rather than reporting medians of jointly measured deployment times. Concurrent execution was not measured. Checkpoint replay and stitching have larger errors in this experiment, so their lower cost is useful only when those errors are acceptable.

At prefix 4096, the corresponding auxiliary-memory counts are $819200+1600N$ bytes for the bank, $16384N$ for full replay, and $2048N$ for checkpoint-256. The bank becomes smaller at 56 and 1829 simultaneously retained streams, respectively. Stitching has no additional stream history but retains the shared maps in Table~\ref{tab:shared-storage}. These comparisons use the stated cached target set and exclude common model weights, common active states, and temporary construction arrays.
\par

\subsection{Conditioning of small universal realizations}\label{app:numerical-precision}
The direction-independent filter is deliberately tested only at small dimensions. For a two-state example with source eigenvalues $0.2,0.55$, scalar input vector $(1,0.7)^\top$, and degrees one through three, the required auxiliary counts are two, four, and six. The respective coordinate-transform condition numbers are approximately $15.6$, $27.5$, and $38.4$. They do not justify a numerical-stability claim at the 128-state experimental dimension. The high-dimensional experiment instead uses the explicitly described projected coefficient and Laguerre realizations.

\par
\paragraph{A sufficient precision condition for a chosen realization.}
Let $z_t=Fz_{t-1}+gx_t$, $z_0=0$, be an exact retained-state realization, with degree-$k$ query response $h_t^{(k)}(\Delta)=D_\Delta z_t$. For a computed trajectory initialized at zero, define its one-step defect against these exact matrices by
\[
r_t=\widehat z_t-F\widehat z_{t-1}-gx_t.
\]
Suppose $\norm{r_t}\leq\eta$ and decoder evaluation contributes error at most $\eta_D$. Subtracting the recurrences and iterating gives
\[
\widehat z_t-z_t=\sum_{s=1}^tF^{t-s}r_s,
\qquad
\norm{\widehat h_t-h_t^{(k)}(\Delta)}
\leq\eta\sum_{j=0}^{t-1}\norm{D_\Delta F^j}+\eta_D.
\]
If $\norm{F^j}\leq K_F\rho_F^j$ for $0\leq\rho_F<1$, this numerical contribution is at most $\eta\norm{D_\Delta}K_F/(1-\rho_F)+\eta_D$. Adding a valid truncation bound yields a sufficient absolute-error budget for that realization. Uniform accuracy over replacement queries requires uniform bounds on these quantities. A numerical projection onto an approximate reachable subspace adds its own representation error.

The defect allowance must include stored-coefficient rounding, arithmetic, and the input and state ranges. Stable eigenvalues alone do not make $K_F$ or the decoder norm small. Non-normal transients and poorly scaled coordinates can make a minimum-dimensional realization less accurate than a larger one. Computing the reachable basis also requires numerical separation of its retained subspace. The condition above explains which quantities must be controlled, but it does not estimate them for a large universal realization or provide validated floating-point certification. When their amplification is excessive, a larger scaled coefficient or orthogonal-history representation may be preferable.
\par

\section{Relationship to earlier methods and the scope of the contribution}\label{app:positioning}
\paragraph{Widths and conditional information.}
\citet{devore1989} develop continuous nonlinear widths, and \citet{siegel2024} discusses sharp manifold-width lower bounds using topological arguments. The fixed-horizon result here uses the same kind of singular-value upper bound and antipodal lower bound, conditioned on the old state. It does not introduce Borsuk--Ulam, a new general nonlinear width, or a lower bound for discontinuous real encodings. The finite-bit covering argument likewise uses an elementary ellipsoid-volume bound. The migration-specific distinction is between information already retained, information a future replacement can expose, and whether its attaining summary can be updated causally.

\paragraph{Realization and parametric model reduction.}
The coefficient lift, its reachable subspace, invariant projection, and common-denominator construction use classical linear-system tools. Projection-based parametric reduction and offline/online error estimation are established subjects \citep{benner2015,haasdonk2011}. The causal minimum follows by combining reachability with the requirement that histories sharing a summary agree in every coefficient demanded by local order. The sharp scalar-input comparison shows that an unknown arbitrary matrix direction and one suitably chosen rank-one direction have the same worst-case local information cost. The contribution lies in these information statements, not in sensitivity propagation or residual estimation.

\paragraph{Recurrent differentiation and restricted adaptation.}
RTRL, UORO, and more recent structured recurrent learning maintain or approximate derivatives for online learning \citep{williams1989,tallec2018,farr2026}. The present query asks how an unavailable historical prefix would be represented after a subsequently selected parameter replacement. First-order sensitivities can answer it approximately, but they are one retained representation rather than a universal minimum. Parameter-efficient adaptation motivates small families \citep{hu2022}, yet a freely trained low-rank factorization need not preserve a fixed image or a small number of scalar coordinates. A fixed image and a small scalar-coordinate family are additional assumptions, not consequences of the term ``adapter.''

\paragraph{Stitching and history bases.}
Stitching exploits compatibility of existing representations, with its behavior depending on training objectives, feature layers, and calibration \citep{bansal2021,mai2026}. It is not ruled out by worst-case collisions when real data occupy a restricted set of histories. Both linear and nonlinear versions are therefore tested. HiPPO provides a principled family of causal history-basis representations \citep{gu2020}. The Laguerre baseline retains model-independent history coordinates and illustrates how a compact shared history basis can replace parameter-specific coefficients. It is not claimed as a new family of orthogonal filters.

\paragraph{Context composition and state compression.}
PICASO composes context states within state-space dynamics \citep{liu2025}, whereas CompreSSM uses control-theoretic state reduction during training \citep{chahine2026}. The target here is a counterfactual state under changed parameters, and the memory must be chosen before those parameters are revealed. A smaller state under one model need not contain the distinctions exposed by another. Structured recurrent models and selective SSMs provide architectural motivation \citep{gu2022,orvieto2023,gu2024}, not automatic instances of the theorem. In particular, parameter-dependent input gates or upstream activations can invalidate the assumption that injections and base matrices are shared across replacements.

%% file: tables/round2_spectrum.tex
\begin{tabular}{rrr}
\toprule
RMS tolerance $\epsilon$ & Extra coordinates $m_\epsilon$ & Bit lower bound \\
\midrule
$10^{-2}$ & 0 & 0.00000 \\
$10^{-3}$ & 1 & 1.90465 \\
$10^{-4}$ & 1 & 5.22657 \\
$10^{-6}$ & 3 & 20.78339 \\
$10^{-8}$ & 5 & 49.74601 \\
\bottomrule
\end{tabular}

%% file: tables/revision_language_all.tex
\begin{tabular}{lrrr}
\toprule
Method & Extra bytes & State NRMSE & Predictive KL \\
\midrule
Full replay & $t$ & $0$ & $0$ \\
Affected-state reset & 0 & $(1.00\pm0.00)\times10^{0}$ & $(5.54\pm1.73)\times10^{-3}$ \\
Stale state & 0 & $(2.95\pm0.79)\times10^{-2}$ & $(2.93\pm0.98)\times10^{-6}$ \\
Checkpoint replay, 64 & 128 & $(1.87\pm0.80)\times10^{-2}$ & $(8.30\pm2.69)\times10^{-7}$ \\
Checkpoint replay, 256 & 320 & $(6.76\pm5.24)\times10^{-3}$ & $(6.72\pm7.12)\times10^{-8}$ \\
Linear stitching & 0 & $(2.33\pm0.54)\times10^{-3}$ & $(4.91\pm0.76)\times10^{-9}$ \\
Nonlinear stitching & 0 & $(3.72\pm1.15)\times10^{-3}$ & $(1.82\pm0.88)\times10^{-8}$ \\
UORO-style rank one & 144 & $(2.66\pm0.48)\times10^{-1}$ & $(1.01\pm0.69)\times10^{-4}$ \\
First-order sensitivity & 128 & $(1.40\pm0.41)\times10^{-2}$ & $(3.64\pm0.74)\times10^{-7}$ \\
Quadratic interpolation & 192 & $(5.48\pm3.63)\times10^{-3}$ & $(3.62\pm2.32)\times10^{-8}$ \\
Degree-2 memory & 192 & $(3.01\pm1.39)\times10^{-3}$ & $(1.32\pm1.29)\times10^{-8}$ \\
Degree-3 memory & 256 & $(7.49\pm5.50)\times10^{-4}$ & $(1.10\pm1.83)\times10^{-9}$ \\
Cubic interpolation & 256 & $(5.30\pm3.46)\times10^{-4}$ & $(2.78\pm2.27)\times10^{-10}$ \\
\bottomrule
\end{tabular}

%% file: tables/revision_task_structure.tex
\begin{tabular}{rrrrrr}
\toprule
Seed & $\rho$ & $p,r$ & Leakage $\delta$ & $\kappa$ & $\|z_{(1,1)}\|$ \\
\midrule
0 & 0.970 & 2, 8 & $4.66e-04$ & 0.988001 & $6.01e-02$ \\
1 & 0.970 & 2, 8 & $2.41e-04$ & 0.988001 & $9.88e-02$ \\
2 & 0.970 & 2, 8 & $2.51e-04$ & 0.988001 & $5.15e-03$ \\
3 & 0.970 & 2, 8 & $1.67e-04$ & 0.988001 & $1.94e-01$ \\
4 & 0.970 & 2, 8 & $2.67e-04$ & 0.988001 & $8.96e-03$ \\
0 & 0.995 & 2, 8 & $5.01e-04$ & 0.998000 & $1.34e-01$ \\
1 & 0.995 & 2, 8 & $2.60e-04$ & 0.998000 & $2.07e-01$ \\
2 & 0.995 & 2, 8 & $2.70e-04$ & 0.998000 & $1.71e-02$ \\
3 & 0.995 & 2, 8 & $1.80e-04$ & 0.998000 & $3.74e-02$ \\
4 & 0.995 & 2, 8 & $2.91e-04$ & 0.998000 & $3.30e-02$ \\
0 & 0.999 & 2, 8 & $5.06e-04$ & 0.999600 & $1.88e-02$ \\
1 & 0.999 & 2, 8 & $2.63e-04$ & 0.999600 & $3.03e-02$ \\
2 & 0.999 & 2, 8 & $2.73e-04$ & 0.999600 & $2.14e-02$ \\
3 & 0.999 & 2, 8 & $1.82e-04$ & 0.999600 & $6.27e-02$ \\
4 & 0.999 & 2, 8 & $2.94e-04$ & 0.999600 & $8.96e-02$ \\
\bottomrule
\end{tabular}

%% file: tables/revision_task_all.tex
\begin{tabular}{lrrr}
\toprule
Method & Extra bytes & Prediction NMSE & Stream ms \\
\midrule
No adaptation & 0 & $(1.34\pm2.34)\times10^{-1}$ & \textemdash \\
Unrestricted fit + replay & $4t$ & $(2.01\pm1.62)\times10^{-9}$ & \textemdash \\
Full replay & $4t$ & $(2.50\pm2.95)\times10^{-9}$ & $30.50$ \\
Affected-state reset & 0 & $(8.40\pm0.42)\times10^{-1}$ & \textemdash \\
Stale state & 0 & $(3.75\pm6.65)\times10^{-1}$ & $6.95$ \\
Checkpoint replay, 16 & 1088 & $(3.18\pm5.59)\times10^{-1}$ & $7.11$ \\
Checkpoint replay, 64 & 1280 & $(1.88\pm3.21)\times10^{-1}$ & $7.62$ \\
Checkpoint replay, 256 & 2048 & $(2.26\pm3.47)\times10^{-2}$ & $9.78$ \\
Linear stitching & 0 & $(1.98\pm4.29)\times10^{-4}$ & $7.06$ \\
First-order sensitivity & 1152 & $(8.38\pm17.50)\times10^{-3}$ & $18.18$ \\
Degree-2 memory & 1344 & $(4.63\pm9.07)\times10^{-4}$ & $21.04$ \\
Degree-3 memory & 1600 & $(2.15\pm4.54)\times10^{-5}$ & $24.67$ \\
Cubic interpolation & 1600 & $(1.73\pm3.33)\times10^{-4}$ & $30.41$ \\
Fitted history bank & 1600 & $(2.08\pm3.82)\times10^{-3}$ & $8.80$ \\
Analytic history bank & 1600 & $(2.50\pm2.95)\times10^{-9}$ & $8.78$ \\
\bottomrule
\end{tabular}

%% file: tables/revision_task_seeds.tex
\begin{tabular}{lrrrrr}
\toprule
Seed & No adaptation & Stale & Degree 3 & Cubic bank & Analytic history \\
\midrule
0 & $6.76\times10^{-2}$ & $1.86\times10^{-1}$ & $2.75\times10^{-7}$ & $1.57\times10^{-5}$ & $5.30\times10^{-10}$ \\
1 & $4.36\times10^{-2}$ & $1.22\times10^{-1}$ & $3.41\times10^{-6}$ & $8.23\times10^{-5}$ & $7.13\times10^{-9}$ \\
2 & $4.70\times10^{-3}$ & $6.25\times10^{-3}$ & $8.52\times10^{-7}$ & $3.70\times10^{-7}$ & $4.90\times10^{-10}$ \\
3 & $2.87\times10^{-3}$ & $6.70\times10^{-3}$ & $8.05\times10^{-8}$ & $5.73\times10^{-7}$ & $5.64\times10^{-10}$ \\
4 & $5.50\times10^{-1}$ & $1.56\times10^{0}$ & $1.03\times10^{-4}$ & $7.65\times10^{-4}$ & $3.77\times10^{-9}$ \\
\bottomrule
\end{tabular}

%% file: tables/revision_contraction.tex
\begin{tabular}{llr}
\toprule
Anchor norm target & Method & Prediction NMSE \\
\midrule
0.97 & Stale state & $(2.31\pm4.28)\times10^{-1}$ \\
0.97 & Checkpoint replay, 256 & $(2.20\pm4.72)\times10^{-6}$ \\
0.97 & Linear stitching & $(4.29\pm9.47)\times10^{-4}$ \\
0.97 & Degree-3 memory & $(2.17\pm4.72)\times10^{-5}$ \\
0.97 & Cubic interpolation & $(1.80\pm3.47)\times10^{-4}$ \\
0.97 & Analytic history bank & $(2.52\pm2.66)\times10^{-9}$ \\
0.995 & Stale state & $(3.75\pm6.65)\times10^{-1}$ \\
0.995 & Checkpoint replay, 256 & $(2.26\pm3.47)\times10^{-2}$ \\
0.995 & Linear stitching & $(1.98\pm4.29)\times10^{-4}$ \\
0.995 & Degree-3 memory & $(2.15\pm4.54)\times10^{-5}$ \\
0.995 & Cubic interpolation & $(1.73\pm3.33)\times10^{-4}$ \\
0.995 & Analytic history bank & $(2.50\pm2.95)\times10^{-9}$ \\
0.999 & Stale state & $(1.64\pm2.88)\times10^{-1}$ \\
0.999 & Checkpoint replay, 256 & $(1.08\pm1.99)\times10^{-1}$ \\
0.999 & Linear stitching & $(2.67\pm5.48)\times10^{-4}$ \\
0.999 & Degree-3 memory & $(2.92\pm6.35)\times10^{-6}$ \\
0.999 & Cubic interpolation & $(3.22\pm6.57)\times10^{-5}$ \\
0.999 & Analytic history bank & $(1.69\pm1.88)\times10^{-9}$ \\
\bottomrule
\end{tabular}

%% file: tables/revision_frequency.tex
\begin{tabular}{llr}
\toprule
Update period & Method & Prediction NMSE \\
\midrule
128 & Stale state & $(1.99\pm3.65)\times10^{-1}$ \\
128 & Checkpoint replay, 256 & $(1.65\pm3.60)\times10^{-2}$ \\
128 & Linear stitching & $(1.93\pm4.07)\times10^{-5}$ \\
128 & Degree-3 memory & $(2.74\pm5.95)\times10^{-5}$ \\
128 & Cubic interpolation & $(2.30\pm4.49)\times10^{-4}$ \\
128 & Analytic history bank & $(2.11\pm2.27)\times10^{-9}$ \\
512 & Stale state & $(3.75\pm6.65)\times10^{-1}$ \\
512 & Checkpoint replay, 256 & $(2.26\pm3.47)\times10^{-2}$ \\
512 & Linear stitching & $(1.98\pm4.29)\times10^{-4}$ \\
512 & Degree-3 memory & $(2.15\pm4.54)\times10^{-5}$ \\
512 & Cubic interpolation & $(1.73\pm3.33)\times10^{-4}$ \\
512 & Analytic history bank & $(2.50\pm2.95)\times10^{-9}$ \\
1024 & Stale state & $(4.23\pm7.18)\times10^{-1}$ \\
1024 & Checkpoint replay, 256 & $(3.93\pm6.80)\times10^{-2}$ \\
1024 & Linear stitching & $(2.71\pm5.99)\times10^{-3}$ \\
1024 & Degree-3 memory & $(2.77\pm6.01)\times10^{-5}$ \\
1024 & Cubic interpolation & $(2.98\pm5.55)\times10^{-4}$ \\
1024 & Analytic history bank & $(3.24\pm3.08)\times10^{-9}$ \\
\bottomrule
\end{tabular}

%% file: tables/revision_certificate.tex
\begin{tabular}{rrrr}
\toprule
Nominal leakage & Degree & Median $c/e$ & 90th percentile \\
\midrule
0 & 1 & $50$ & $1.1e+02$ \\
0 & 2 & $1.4e+02$ & $4.3e+02$ \\
0 & 3 & $3.8e+02$ & $1.5e+03$ \\
0 & 4 & $1.2e+03$ & $7.3e+03$ \\
0.001 & 1 & $54$ & $1.2e+02$ \\
0.001 & 2 & $47$ & $84$ \\
0.001 & 3 & $31$ & $61$ \\
0.001 & 4 & $29$ & $57$ \\
0.003 & 1 & $56$ & $1e+02$ \\
0.003 & 2 & $35$ & $68$ \\
0.003 & 3 & $29$ & $58$ \\
0.003 & 4 & $29$ & $57$ \\
0.01 & 1 & $43$ & $74$ \\
0.01 & 2 & $30$ & $62$ \\
0.01 & 3 & $29$ & $58$ \\
0.01 & 4 & $29$ & $58$ \\
0.03 & 1 & $34$ & $62$ \\
0.03 & 2 & $29$ & $58$ \\
0.03 & 3 & $28$ & $57$ \\
0.03 & 4 & $28$ & $57$ \\
\bottomrule
\end{tabular}

%% file: tables/revision_certificate_policy.tex
\begin{tabular}{rrrr}
\toprule
Tolerance & Nominal leakage & Certified fraction & Mean selected degree \\
\midrule
0.001 & 0 & 1.00 & 3.12 \\
0.001 & 0.001 & 0.00 & \textemdash \\
0.001 & 0.003 & 0.00 & \textemdash \\
0.001 & 0.01 & 0.00 & \textemdash \\
0.001 & 0.03 & 0.00 & \textemdash \\
0.01 & 0 & 1.00 & 2.00 \\
0.01 & 0.001 & 1.00 & 2.07 \\
0.01 & 0.003 & 0.55 & 2.58 \\
0.01 & 0.01 & 0.00 & \textemdash \\
0.01 & 0.03 & 0.00 & \textemdash \\
0.1 & 0 & 1.00 & 1.00 \\
0.1 & 0.001 & 1.00 & 1.00 \\
0.1 & 0.003 & 1.00 & 1.00 \\
0.1 & 0.01 & 1.00 & 1.00 \\
0.1 & 0.03 & 0.70 & 1.31 \\
\bottomrule
\end{tabular}

%% file: tables/revision_task_timing.tex
\begin{tabular}{lrr}
\toprule
Method & Stream ms & Interquartile range, ms \\
\midrule
Cubic interpolation & $30.408$ & $[30.077,30.659]$ \\
First-order sensitivity & $18.176$ & $[17.982,18.377]$ \\
Full replay & $30.495$ & $[30.142,30.950]$ \\
Analytic history bank & $8.776$ & $[8.638,8.952]$ \\
Fitted history bank & $8.796$ & $[8.634,8.998]$ \\
Linear stitching & $7.065$ & $[6.950,7.159]$ \\
Stale state & $6.953$ & $[6.891,7.073]$ \\
Degree-2 memory & $21.036$ & $[20.849,21.222]$ \\
Degree-3 memory & $24.673$ & $[24.443,24.870]$ \\
Checkpoint replay, 16 & $7.106$ & $[7.020,7.270]$ \\
Checkpoint replay, 256 & $9.777$ & $[9.709,9.972]$ \\
Checkpoint replay, 64 & $7.615$ & $[7.584,7.738]$ \\
\bottomrule
\end{tabular}

%% file: tables/revision_setup.tex
\begin{tabular}{lrr}
\toprule
Shared operation & Median ms & IQR, ms \\
\midrule
Analytic history, four decoders & $4.83$ & $[4.51,5.03]$ \\
Coefficient model reductions & $0.57$ & $[0.39,1.18]$ \\
Cubic bank reductions and nodes & $0.56$ & $[0.43,1.11]$ \\
Fitted history, four maps & $141.73$ & $[137.48,145.36]$ \\
Linear stitching, five maps & $136.46$ & $[133.25,141.61]$ \\
\bottomrule
\end{tabular}

%% file: tables/revision_shared_memory.tex
\begin{tabular}{lr}
\toprule
Shared auxiliary model data & Bytes \\
\midrule
Linear stitching, five maps & 670,720 \\
Fitted history, four maps & 836,096 \\
Analytic history, four decoders & 819,200 \\
Cubic nodes and interpolation inverse & 792 \\
Shared $U,T,G_1,G_2$ & 9,728 \\
\bottomrule
\end{tabular}

%% file: sections/round2_experiments.tex
\section{History width, structural variation, and numerical realization}\label{app:round2}
This section studies two distinct questions: the width needed by a prescribed history representation at a fixed tolerance, and a small numerical realization of the causal information minimum. A finite-width Laguerre bank is an approximate representation with an independently chosen pole. A reachable-space realization represents a specified truncated polynomial response without approximation in real arithmetic. Neither experiment characterizes the optimal finite-bit causal encoder over all representations.

\subsection{Information timing and evaluation objects}\label{app:round2-timeline}
Table~\ref{tab:timeline} states when information becomes available. ``Calibration'' refers to independent transition observations, not to recovery of the evaluation stream. A decoder may be constructed after a target is fitted, but it may not use the past evaluation inputs unless those inputs were explicitly retained and charged as replay memory.
\begin{table}[ht]
\caption{Information available at each stage of the coupled identification experiment.}
\label{tab:timeline}
\centering\small
\begin{tabular}{p{.18\linewidth}p{.40\linewidth}p{.32\linewidth}}
\toprule
Stage & Available information & Operation \\
\midrule
Source and pilot fitting & Independent source transitions and six pilot domains, observed states, known injection and readout & Fit the anchor, update family, and image \\
Memory provision & Anchor, permitted family, chosen degree or bank width, prescribed pole & Fix the causal recurrence before evaluation inputs or replacement queries \\
Streaming & Current input, previous retained memory, anchor and active model & Maintain the summary and active prediction state \\
Target fitting & Independent transitions from a new domain & Fit replacement coordinates and construct shared target decoder \\
Migration query & Retained summary and fitted target, but not discarded evaluation inputs & Reconstruct a target state and continue under common future inputs \\
\bottomrule
\end{tabular}
\end{table}
Independent target fits are computed before evaluating the held-out streams. Neither family selection nor the retained-memory recurrence receives those fitted target coordinates. Moving the target fits after historical maintenance would therefore leave the summaries unchanged. The retrospective reference at each query runs the fitted target over the entire input prefix. The true conditional-mean target instead runs the generating target model over that prefix. They coincide only to the extent that system identification is accurate. Comparing against these two targets separates migration fidelity from task error.

\subsection{Prespecified bank-width comparison}\label{app:round2-width}
The repeated-pole arm uses the five frozen fitted systems from the main $\rho=0.995$, period-512 experiment. All fitted systems and seeds are retained without refitting or selection. The input seed for fitted seed $s$ and stream $j$ is $37001+100s+j$, as in the original experiment. Each of eight streams contains 4096 independent standard Gaussian inputs stored in float32. State calculations and retained bank coordinates use float64. Queries follow 512, 1024, \ldots, 3584 inputs, and cycle through the four fitted target domains. The evaluation window includes the query output and 31 subsequent outputs.

Before running this width comparison, the configuration fixes
\[
\mathcal M=\{1,2,4,8,16,32,64,128,200,400,800\},\qquad
\mathcal E=\{10^{-2},10^{-3},10^{-4}\}.
\]
Every bank uses pole $\alpha=0.995$, taken from the anchor design, without tuning to a replacement, a test history, or a target tolerance. Width is an experimental resource variable. The smallest successful tested width describes an empirical frontier. It is not a deployed selection rule with a population guarantee.

The discrete Laguerre transfer functions are
\begin{equation}
\phi_m(z)=\frac{\sqrt{1-\alpha^2}}{1-\alpha z}
\left(\frac{z-\alpha}{1-\alpha z}\right)^m,\qquad m=0,1,\ldots.
\label{eq:round2-laguerre}
\end{equation}
This is a classical system-identification basis \citep{wahlberg1991}. The causal recursion and analytic decoder are given in Appendix~\ref{app:revision-task}. All bank states start at zero. For each target matrix $A$, decoder columns satisfy
\begin{equation}
\ell_0=\sqrt{1-\alpha^2}(I-\alpha A)^{-1}b,\qquad
\ell_m=(I-\alpha A)^{-1}(A-\alpha I)\ell_{m-1}.
\label{eq:round2-decoder}
\end{equation}
A width-$M$ reconstruction is $\widehat h_t^{(M)}=\sum_{m=0}^{M-1}\ell_m s_{t,m}$, where $s_{t,m}$ is the bank's response to the evaluation prefix. The decoder uses only the target model and the prescribed basis. It is not fitted on evaluation histories. A width-800 bank is maintained once for efficient evaluation, and each smaller width uses its prefix. The first $M$ coordinates have a closed causal recursion that never uses later coordinates, so this evaluation gives the same states as an independently maintained width-$M$ bank. The width-$M$ memory count excludes the other coordinates in the 800-coordinate evaluation array.

\paragraph{Three error measures.}
For one fitted-model seed, sum over its 56 events, denoted by $e$. Let $h_e$ be the fitted target's exact replay state at the event, and $\widehat h_e$ a migrated state. Let $y^*_{e,s}$, $y^{\rm ref}_{e,s}$, and $\widehat y_{e,s}$ be the generating conditional mean, fitted-target replay output, and evaluated method's output at offset $s\in\{0,\ldots,31\}$. We report
\begin{align}
\operatorname{NMSE}_{\rm task}
&=\frac{\sum_{e,s}(\widehat y_{e,s}-y^*_{e,s})^2}
{\sum_{e,s}(y^*_{e,s})^2},\label{eq:round2-task-error}\\
\operatorname{NRMSE}_{\rm out}
&=\left[\frac{\sum_{e,s}(\widehat y_{e,s}-y^{\rm ref}_{e,s})^2}
{\sum_{e,s}(y^{\rm ref}_{e,s})^2}\right]^{1/2},\label{eq:round2-output-error}\\
\operatorname{NRMSE}_{\rm state}
&=\left[\frac{\sum_e\norm{\widehat h_e-h_e}^2}
{\sum_e\norm{h_e}^2}\right]^{1/2}.\label{eq:round2-state-error}
\end{align}
Thus state fidelity is measured at migration and output fidelity over a subsequent common-input window. The two normalized quantities use different denominators and measure different losses. All task NMSE means and sample SDs use five seed-level ratios. Tolerance tables use the \emph{maximum of the five seed-level NRMSEs}. They do not bound the worst event, an unseen seed, an arbitrary input sequence, or an infinite future horizon. The nonzero readout in every tested stream gives positive denominators.

\paragraph{Memory accounting.}
A history bank needs $8M$ per-stream bytes in float64, beyond the common active 128-state model. It needs no retained anchor. Its four target-specific $128\times M$ decoder matrices require $4\cdot128\cdot M\cdot8$ shared bytes. At $M=8,64,200$, these are 32,768, 262,144, and 819,200 shared bytes, respectively. Direct decoder multiplication costs $O(dM)$ per query, bank maintenance costs $O(M)$ per input, and analytic decoder construction is shared work after the target is known. The degree-three coefficient and multivariate interpolation methods retain 200 scalars, including their anchor, or 1600 extra bytes. Full replay retains the input prefix. The width sweep reports accuracy and logical storage, not newly measured timings or a claim that model-dependent decoder setup is free.

\begin{figure}[t]
\centering\includegraphics[width=\linewidth]{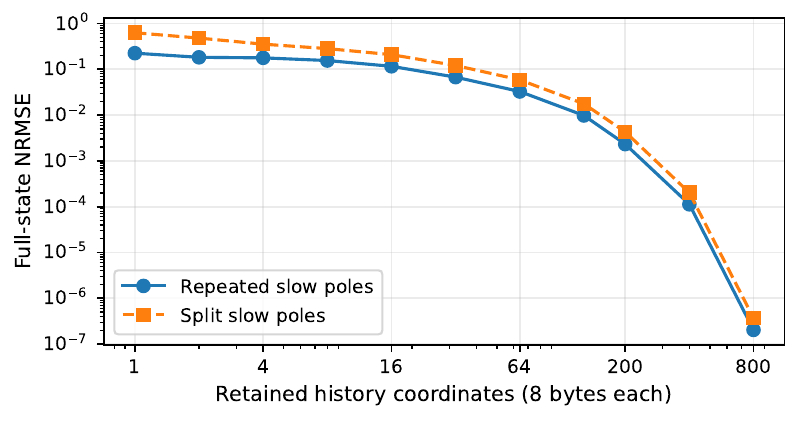}
\caption{Full-state fidelity for the same bank widths and queries as Figure~\ref{fig:width-output}. The common readout can be accurate while many state coordinates remain inaccurate. These results concern a finite empirical design; the common-invariant quotient theorem is a separate exact statement.}
\label{fig:width-state}
\end{figure}
\begin{table}[t]
\caption{Complete width comparison. Columns show maximum seed-level NRMSE over five independently fitted systems. ``Rep.'' denotes repeated slow poles. Byte counts are additional per-stream bank storage.}
\label{tab:round2-widths}
\centering\small\input{tables/round2_widths}
\end{table}

Eight coordinates have maximum seed output NRMSE $\HistoryEightOutputMaximum$ in the repeated-pole arm, while their mean full-state NRMSE is $\HistoryEightStatePercent\%$. Their task NMSE is $\RepeatedEightTaskNmse$, approximately the identification error achieved by replay. The 200-coordinate budget therefore exceeds the requirement of this tested prediction criterion. It is not a claim that eight is the optimal continuous dimension or that the retained history preserves every state coordinate. At output tolerance $10^{-2}$, the smallest successful tested width is four; at $10^{-3}$ and $10^{-4}$ it is eight. Table~\ref{tab:width} contrasts those counts with full-state tolerances.

\subsection{A paired split-pole structural variant}\label{app:round2-split}
The baseline generator uses a repeated slow pole $0.995$ on an eight-dimensional block. With scalar input, many update coefficients then share the same weighted historical moments. To weaken that property while keeping a useful adaptation problem, replace only the true slow diagonal by
\begin{equation}
\lambda_j=0.995-0.04\frac{j}{7},\qquad j=0,\ldots,7.
\label{eq:round2-split-poles}
\end{equation}
The 120 fast poles, Gaussian rotation, injection and readout construction, update-direction scale, calibration observation noise, number of calibration observations, and independent seed rules are unchanged. We fit all five source models, pilot families, and target domains anew. No method receives the true split-pole invariant basis. The same fixed singular-value energy rule chooses the fitted image and update directions. Each fitted model and its metadata are retained.

The bank pole remains $0.995$ in this arm. Its relative mismatch to some of the true poles therefore increases. Update perturbations are still confined by construction to the eight-dimensional slow block, and calibration still exposes the full state. The variant tests sensitivity to a repeated-pole assumption, not removal of all favorable structure. Different pole locations also change trajectory energy and the importance of old inputs. Consequently a between-arm error ratio is not an isolated causal estimate of ``number of distinct poles'' in an otherwise invariant prediction distribution.

Useful adaptation survives. Mean task NMSE is $\SplitAnchorNmse$ without adaptation and $\SplitStaleNmse$ with stale reuse, versus $\SplitReplayNmse$ for restricted replay and $\SplitUnrestrictedNmse$ for unrestricted replay. Degree-three coefficients attain $\SplitDegreeThreeNmse$. The analytic bank's eight-coordinate maximum seed output NRMSE is $\SplitEightOutputMaximum$, and 64 coordinates reduce it to $\SplitSixtyFourOutputMaximum$. Among the prescribed widths, 16, 32, and 64 meet output tolerances $10^{-2},10^{-3},10^{-4}$ in every seed. The split-pole setting weakens the small history bank without establishing a general advantage for another representation.

\begin{table}[t]
\caption{Prediction and fidelity in the repeated-pole arm. Task NMSE is mean $\pm$ sample SD across five fitted seeds. Output and state NRMSE are maxima of the five seed-level ratios. Baselines are evaluated on the identical histories and target schedule.}
\label{tab:round2-repeated-task}
\centering\small\input{tables/round2_repeated_task}
\end{table}
\begin{table}[t]
\caption{The corresponding split-pole arm. A method can preserve output fidelity more accurately than full-state fidelity. The unrestricted replay row evaluates its own fitted transition, not a migration error under the restricted family. Stitching and first-order memory are evaluated on the frozen fitted systems as described below.}
\label{tab:round2-split-task}
\centering\small\input{tables/round2_split_task}
\end{table}

The study contains two structural arms, five fitted seeds per arm, eight streams, seven events per stream, and 17 methods per event. Its 9,520 event-method records retain both squared-error numerators and target-energy denominators. The records include every stream and width, including those that fail the tolerances. The repeated-pole arm holds the fitted models fixed, while the split-pole arm fits a separate model family for each seed.

\par
\paragraph{Stitching and first-order memory.}
A follow-up holds the five fitted split-pole systems fixed and evaluates these two methods on the same eight streams and seven queries. Stitching uses the original independent calibration inputs: 8192 Gaussian values with seed $26001+s$ for fitted seed $s$, retaining every fourth state after the first 1024 inputs. The first 75\% of retained rows fit the map, and the remaining 25\% select the ridge penalty from $10^{-8},10^{-6},10^{-4},10^{-2}$. Model-pair maps cover the anchor-to-first-target transition and successive target transitions, including the return from target four to target one. Only these maps are fitted. The recurrent models, held-out histories, and 32-position evaluation windows are unchanged. Repeating the comparison in the repeated-pole arm gives 1120 event-method records across both arms.

Table~\ref{tab:round2-split-task} reports the split-pole results. Stitching has mean task NMSE $\SplitStitchNmse$, compared with $\SplitFirstNmse$ for first-order memory. Its maximum seed output NRMSE is $\SplitStitchOutputMax$, so it fails the $10^{-3}$ criterion across all five seeds. First-order memory truncates the response about the anchor. Stitching instead fits the finite replacement from full calibration states and can exploit correlations among histories in that distribution. The local-order lower bound does not rank those two approximations on this distribution.

\paragraph{Comparison at the same tolerance.}
At maximum seed output NRMSE $10^{-3}$, degree three and the width-32 bank both qualify, with errors $1.19\times10^{-4}$ and $2.36\times10^{-4}$. They use 1600 and 256 extra per-stream bytes. At $10^{-4}$, degree three fails, while width 64 qualifies with error $1.86\times10^{-7}$ and 512 bytes. Small task NMSE alone does not imply this output-fidelity requirement.

Including shared storage changes the $10^{-3}$ comparison. Four $128\times32$ float64 decoders use 131072 bytes, whereas the coefficient reductions use 9728 bytes under the auxiliary-storage convention of Appendix~\ref{app:revision-costs}. Thus the totals are $131072+256N$ and $9728+1600N$ bytes for $N$ simultaneously retained streams. The bank becomes smaller at 91 streams. Its recurrence costs $O(M)$ work per input and full-state decoding costs $O(dM)$ per query. The coefficient method also maintains the anchor and reduced responses. Width-specific runtimes were not measured, so these counts do not determine a runtime crossover.

For full-state NRMSE $10^{-4}$ instead, degree three qualifies at $5.04\times10^{-5}$ with 1600 extra bytes. The smallest successful tested bank has width 800 and uses 6400 bytes. This is a different reconstruction objective, not a predictive advantage or an optimum over all history bases.
\par

\subsection{An orthogonal realization of the causal minimum}\label{app:round2-reachable}
For a tractable fixed-family example, use
\begin{equation}
A_0=\begin{bmatrix}1/5&0\\0&2/5\end{bmatrix},\quad
b=\begin{bmatrix}1\\2\end{bmatrix},\quad
A_1=\frac1{50}\begin{bmatrix}1&2\\3&4\end{bmatrix},\quad
A_2=\frac1{50}\begin{bmatrix}3&-1\\2&1\end{bmatrix}.
\label{eq:round2-small-model}
\end{equation}
The two directions do not commute. Their raw degree-one, degree-two, and degree-three lifts have dimensions 6, 12, and 20. Rational arithmetic gives reachable dimensions 4, 6, and 8, respectively. Since the anchor is controllable, Theorem~\ref{thm:causal-minimal} gives exactly 2, 4, and 6 auxiliary coordinates. Replacing $A_0$ by $(1/5)I_2$ leaves anchor rank one and gives reachable dimensions 2, 3, and 4, with only 1, 2, and 3 auxiliary coordinates. Table~\ref{tab:round2-reachable} reports both cases. These are exact rank calculations on stated rational matrices, not floating-point rank estimates.

Here is an anchor-preserving realization for the distinct-pole case. Let $W$ have orthonormal columns spanning $\mathcal K_k$ in the lifted coordinates, and put $H=\pi_0W$. Then $H$ has row rank $d$. Let $N$ have orthonormal columns spanning $\ker H$, and define
\begin{equation}
J=\begin{bmatrix}H\\N^\top\end{bmatrix},\qquad
q_t=JW^\top j_t.
\label{eq:round2-coordinates}
\end{equation}
If $Jv=0$, then $Hv=0$ and $N^\top v=0$. The first identity puts $v$ in the column space of $N$, and the second then forces $v=0$. Thus $J$ is invertible. The first $d$ coordinates of $q_t$ are exactly the anchor; the remaining $\dim\mathcal K_k-d$ are auxiliary. In real arithmetic the causal dynamics are
\begin{equation}
q_t=JW^\top\mathsf T_kWJ^{-1}q_{t-1}
      +JW^\top\mathsf B_kx_t,\qquad
j_t=WJ^{-1}q_t.
\label{eq:round2-reduced-recurrence}
\end{equation}
Reachability ensures $j_t\in\mathcal K_k$, so $WW^\top j_t=j_t$; invariance of that subspace proves the recurrence. This is the linear-complement construction in the causal theorem expressed in orthogonal coordinates. It is not a new general model-reduction algorithm.

\par
The conditioning of this coordinate choice has an explicit form. With nonzero auxiliary dimension, $J^\top J=H^\top H+NN^\top$ acts as the identity on $\ker H$ and as $H^\top H$ on its orthogonal complement. Its eigenvalues are therefore one on the auxiliary subspace and the squared nonzero singular values of $H$ on the complement. Hence
\[
\kappa_2(J)=
\frac{\max\{1,\sigma_{\max}(H)\}}
     {\min\{1,\sigma_{\min}(H)\}}.
\]
Here $\sigma_{\min}(H)$ is the smallest of the $d$ positive singular values. Full row rank does not prevent this value from being small. Appendix~\ref{app:numerical-precision} gives the separate bound for accumulated step and decoder errors. Neither bound supplies a uniform conditioning guarantee for the direction-independent filter.
\par

For numerical implementation, $W$ is obtained by SVD of the finite Krylov matrix with relative singular-value threshold $10^{-12}$. Its selected dimension agrees with the rational calculation in every tested case. We report $\|W^\top W-I\|_2$, the closure residual $\|(I-WW^\top)\mathsf T_kW\|_2$, and $\operatorname{cond}_2(J)$ rather than equating a chosen threshold with exact rank. The coordinate condition numbers are approximately 1.124, 1.152, and 1.166 for degrees one through three. Closure residuals range from $2.9\times10^{-16}$ to $2.5\times10^{-13}$. Setup uses float64; the serving recurrence is evaluated separately in float32 and float64. The anchor block is updated directly under $A_0$ at each step, removing an unnecessary numerical discrepancy in the coordinates designated as side information.

Five Gaussian input sequences of length 4096 are generated with seeds $62001,\ldots,62005$. At every 128th input, a two-coordinate query is drawn uniformly from $[-0.5,0.5]^2$ using the same generator. The reduced state is decoded to its polynomial response and compared with an independently maintained full coefficient array. NRMSE pools squared response discrepancies and squared full-array response norms within a seed. This comparison excludes the common Taylor truncation error against the nonlinear-in-parameter target response. The maximum discrepancies across degrees and seeds are $\ReachableDoubleError$ in float64 and $\ReachableFloatError$ in float32.

\begin{table}[t]
\caption{Redundant lifts and an anchor-preserving numerical realization. ``Rep.'' replaces the distinct anchor poles by a repeated pole. Auxiliary dimensions are computed in rational arithmetic. NRMSE columns are maximum seed discrepancies from the full polynomial response in the distinct-pole case.}
\label{tab:round2-reachable}
\centering\small\input{tables/round2_reachable}
\end{table}
The computation realizes the local minimum with small numerical error in this fixed family. It does not prove that the same construction is well conditioned for large systems, near-defective dynamics, or arbitrary subsequently specified directions. The universal filter in Corollary~\ref{cor:universal} has a stronger direction-independent information contract and a different conditioning problem. The numerical check does not extend to that stronger contract.

\subsection{Paired seed comparisons}\label{app:round2-paired}
\begin{figure}[!ht]
\centering\includegraphics[width=\linewidth]{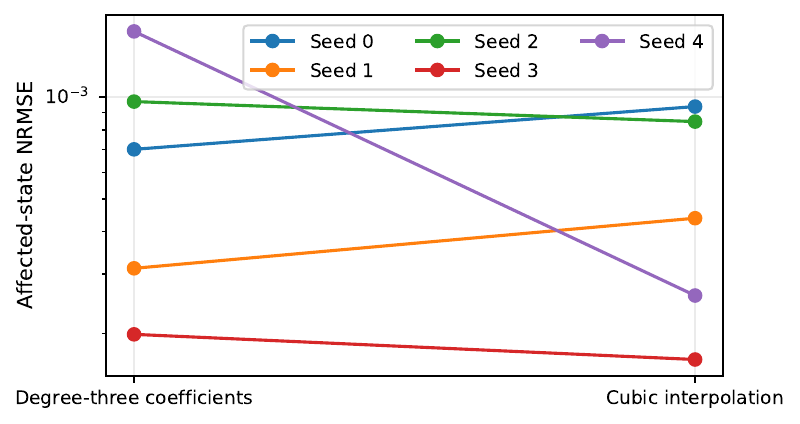}
\caption{The same five language models evaluated with degree-three coefficient memory and equal-memory cubic interpolation. Lines join methods within a fitted-model seed. The figure reports migration-state fidelity, not a meaningful improvement in predictive loss.}
\label{fig:language-paired}
\end{figure}

\begin{figure}[!ht]
\centering\includegraphics[width=\linewidth]{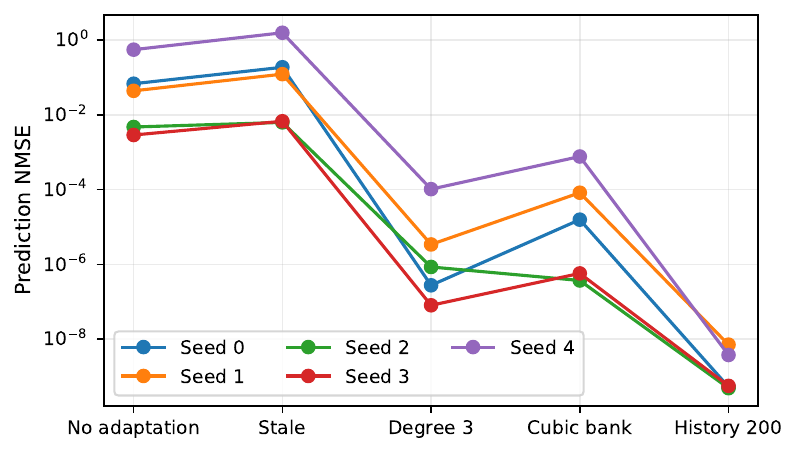}
\caption{Seed-paired task NMSE in the original coupled identification study. The analytic bank here has 200 coordinates. The individual points show the between-seed variation in coefficient and interpolation errors.}
\label{fig:task-paired}
\end{figure}

Seed-level averages can conceal variation across fitted models. Figure~\ref{fig:language-paired} retains all five language-model seeds. Cubic interpolation has lower state error in three and higher error in two. Its lower mean does not imply lower error for each model. Figure~\ref{fig:task-paired} shows the same paired structure for the original coupled study. Seed 4 contributes $\DominantSeedShare\%$ of the summed degree-three seed NMSE. The task's adaptation benefit is consequently computed within each seed before summarization, as in the original analysis. These plots are descriptive paired observations, not simultaneous confidence bands or evidence from newly independent training samples.
\par
For the language comparison, let $d_s$ be cubic-interpolation NRMSE minus degree-three NRMSE for fitted seed $s$. The five differences have mean $\PairedLanguageDifference$ and sample SD $\PairedLanguageSd$. A descriptive paired Student-$t$ interval, using four degrees of freedom, is $\PairedLanguageCI$ at 95\%, with two-sided $p=\PairedLanguageP$. Enumerating all $2^5$ sign patterns gives two-sided $p=\PairedLanguageFlipP$ under the paired-difference sign-symmetry assumption. Neither test establishes superiority. Migration events within a fitted seed are not independent trained models, and no multiplicity-adjusted claim is made.
\par


%% file: tables/round2_widths.tex
\begin{tabular}{rrrrrr}
\toprule
$M$ & Bytes & \multicolumn{2}{c}{Output NRMSE} & \multicolumn{2}{c}{State NRMSE} \\
 & & Repeated & Split & Repeated & Split \\
\midrule
1 & 8 & $4.64\times10^{-1}$ & $5.99\times10^{-1}$ & $2.25\times10^{-1}$ & $6.27\times10^{-1}$ \\
2 & 16 & $1.10\times10^{-1}$ & $3.91\times10^{-1}$ & $1.82\times10^{-1}$ & $4.79\times10^{-1}$ \\
4 & 32 & $2.16\times10^{-3}$ & $2.10\times10^{-1}$ & $1.78\times10^{-1}$ & $3.53\times10^{-1}$ \\
8 & 64 & $3.07\times10^{-7}$ & $6.56\times10^{-2}$ & $1.55\times10^{-1}$ & $2.83\times10^{-1}$ \\
16 & 128 & $8.16\times10^{-8}$ & $8.96\times10^{-3}$ & $1.16\times10^{-1}$ & $2.09\times10^{-1}$ \\
32 & 256 & $2.42\times10^{-8}$ & $2.36\times10^{-4}$ & $6.73\times10^{-2}$ & $1.21\times10^{-1}$ \\
64 & 512 & $5.11\times10^{-9}$ & $1.86\times10^{-7}$ & $3.27\times10^{-2}$ & $5.87\times10^{-2}$ \\
128 & 1024 & $1.13\times10^{-9}$ & $2.93\times10^{-9}$ & $9.80\times10^{-3}$ & $1.76\times10^{-2}$ \\
200 & 1600 & $2.98\times10^{-10}$ & $6.62\times10^{-10}$ & $2.32\times10^{-3}$ & $4.30\times10^{-3}$ \\
400 & 3200 & $1.12\times10^{-11}$ & $2.25\times10^{-11}$ & $1.13\times10^{-4}$ & $2.02\times10^{-4}$ \\
800 & 6400 & $1.79\times10^{-14}$ & $4.44\times10^{-14}$ & $2.04\times10^{-7}$ & $3.71\times10^{-7}$ \\
\bottomrule
\end{tabular}

%% file: tables/round2_repeated_task.tex
\begin{tabular}{lrrrr}
\toprule
Method & Task NMSE & Sample SD & Output NRMSE & State NRMSE \\
\midrule
No adaptation & $1.34\times10^{-1}$ & $2.34\times10^{-1}$ & $7.42\times10^{-1}$ & $1.90\times10^{-1}$ \\
Stale & $3.75\times10^{-1}$ & $6.65\times10^{-1}$ & $1.25\times10^{0}$ & $3.36\times10^{-1}$ \\
Restricted replay & $2.50\times10^{-9}$ & $2.95\times10^{-9}$ & $7.98\times10^{-17}$ & $0$ \\
Unrestricted replay & $2.01\times10^{-9}$ & $1.62\times10^{-9}$ & $1.17\times10^{-4}$ & $5.34\times10^{-5}$ \\
Degree 3 & $2.15\times10^{-5}$ & $4.54\times10^{-5}$ & $1.01\times10^{-2}$ & $2.42\times10^{-3}$ \\
Cubic bank & $1.73\times10^{-4}$ & $3.33\times10^{-4}$ & $2.76\times10^{-2}$ & $8.00\times10^{-3}$ \\
History, 4 & $9.60\times10^{-7}$ & $2.08\times10^{-6}$ & $2.16\times10^{-3}$ & $1.78\times10^{-1}$ \\
History, 8 & $2.50\times10^{-9}$ & $2.95\times10^{-9}$ & $3.07\times10^{-7}$ & $1.55\times10^{-1}$ \\
History, 16 & $2.50\times10^{-9}$ & $2.95\times10^{-9}$ & $8.16\times10^{-8}$ & $1.16\times10^{-1}$ \\
History, 32 & $2.50\times10^{-9}$ & $2.95\times10^{-9}$ & $2.42\times10^{-8}$ & $6.73\times10^{-2}$ \\
History, 64 & $2.50\times10^{-9}$ & $2.95\times10^{-9}$ & $5.11\times10^{-9}$ & $3.27\times10^{-2}$ \\
History, 200 & $2.50\times10^{-9}$ & $2.95\times10^{-9}$ & $2.98\times10^{-10}$ & $2.32\times10^{-3}$ \\
\bottomrule
\end{tabular}

%% file: tables/round2_split_task.tex
\begin{tabular}{lrrrr}
\toprule
Method & Task NMSE & Sample SD & Output NRMSE & State NRMSE \\
\midrule
No adaptation & $1.40\times10^{-2}$ & $2.26\times10^{-2}$ & $2.33\times10^{-1}$ & $6.84\times10^{-2}$ \\
Stale & $2.90\times10^{-2}$ & $4.99\times10^{-2}$ & $3.43\times10^{-1}$ & $1.23\times10^{-1}$ \\
Restricted replay & $3.08\times10^{-10}$ & $4.30\times10^{-10}$ & $1.15\times10^{-16}$ & $0$ \\
Unrestricted replay & $2.66\times10^{-10}$ & $2.48\times10^{-10}$ & $2.05\times10^{-5}$ & $2.08\times10^{-5}$ \\
\input{tables/discussion_split_controls}
Degree 3 & $3.54\times10^{-9}$ & $5.15\times10^{-9}$ & $1.19\times10^{-4}$ & $5.04\times10^{-5}$ \\
Cubic bank & $1.87\times10^{-8}$ & $1.91\times10^{-8}$ & $2.10\times10^{-4}$ & $3.24\times10^{-4}$ \\
History, 4 & $1.33\times10^{-2}$ & $1.76\times10^{-2}$ & $2.10\times10^{-1}$ & $3.53\times10^{-1}$ \\
History, 8 & $1.69\times10^{-3}$ & $1.77\times10^{-3}$ & $6.56\times10^{-2}$ & $2.83\times10^{-1}$ \\
History, 16 & $3.93\times10^{-5}$ & $3.66\times10^{-5}$ & $8.96\times10^{-3}$ & $2.09\times10^{-1}$ \\
History, 32 & $2.38\times10^{-8}$ & $2.59\times10^{-8}$ & $2.36\times10^{-4}$ & $1.21\times10^{-1}$ \\
History, 64 & $3.08\times10^{-10}$ & $4.31\times10^{-10}$ & $1.86\times10^{-7}$ & $5.87\times10^{-2}$ \\
History, 200 & $3.08\times10^{-10}$ & $4.30\times10^{-10}$ & $6.62\times10^{-10}$ & $4.30\times10^{-3}$ \\
\bottomrule
\end{tabular}

%% file: tables/discussion_split_controls.tex
Linear stitching & $1.59\times10^{-6}$ & $1.59\times10^{-6}$ & $2.04\times10^{-3}$ & $3.81\times10^{-3}$ \\
First-order sensitivity & $7.22\times10^{-5}$ & $1.32\times10^{-4}$ & $1.76\times10^{-2}$ & $5.14\times10^{-3}$ \\

%% file: tables/round2_reachable.tex
\begin{tabular}{rrrrrrr}
\toprule
Degree & Full lift & Extra & Extra, rep. & $\operatorname{cond}(J)$ & Float64 & Float32 \\
\midrule
1 & 6 & 2 & 1 & 1.124 & $1.50\times10^{-16}$ & $4.79\times10^{-8}$ \\
2 & 12 & 4 & 2 & 1.152 & $2.49\times10^{-16}$ & $4.72\times10^{-8}$ \\
3 & 20 & 6 & 3 & 1.166 & $1.47\times10^{-16}$ & $4.77\times10^{-8}$ \\
\bottomrule
\end{tabular}